\documentclass[runningheads]{llncs}

\usepackage{eccv}

\usepackage{eccvabbrv}
\usepackage{wrapfig}
\usepackage[export]{adjustbox}
\usepackage{graphicx}
\usepackage{booktabs}
\usepackage{listings}
\usepackage{makecell}
\usepackage{amsmath}

\usepackage{amsthm}
\usepackage{multirow}
\usepackage{colortbl}

\usepackage{bm}
\usepackage{subcaption}
\usepackage{float}
\usepackage{nicematrix, booktabs}
\usepackage{stackengine}
\usepackage{adjustbox}
\usepackage{cancel}

\newcommand{\ifcommentsenabled}[1]{}

\definecolor{tab_color}{rgb}{0.15,0.15,0.65}

\definecolor{felix_color}{rgb}{.6,.4,.05}
\definecolor{michael_color}{rgb}{0,0.35,0}
\definecolor{lukas_color}{rgb}{0.35,0.6,0.6}
\definecolor{thomas_color}{rgb}{0,0,0.85}
\definecolor{thomas_color2}{rgb}{0.1,0.2,0.55}
\definecolor{markus_color}{rgb}{0,0.35,0.35}
\definecolor{bernhard_color}{rgb}{0.35,0.35,0}

\newcommand{\todo}[1]{}

\newcommand{\milo}{MILo\xspace}

\definecolor{edited_color}{rgb}{.8,.15,.15}

\definecolor{revised_color}{rgb}{.15,.15,.8}

\newcommand{\para}[1]{\paragraph{\textbf{#1}.}}
\newcommand{\msub}[2]{{#1}_{\text{#2}}}

\newcommand{\conf}{\gamma}
\newcommand{\confa}{\tilde{\gamma}}
\newcommand{\Rbb}{\mathbb{R}}

\newcommand{\igt}{\mathbf{I}}
\newcommand{\irender}{\hat{\mathbf{I}}}
\newcommand{\iapp}{\hat{\mathbf{I}}^{\text{app}}}

\usepackage[accsupp]{axessibility}  

\usepackage[pagebackref,breaklinks,colorlinks,citecolor=eccvblue]{hyperref}

\usepackage{orcidlink}

\begin{document}

\title{Confidence-Based Mesh Extraction from 3D Gaussians} 


\author{
Lukas Radl\inst{1}\thanks{denotes equal contribution.}\and
Felix Windisch\inst{1}\protect\footnotemark[1]\and
Andreas Kurz\inst{1}\protect\footnotemark[1]\and \\
Thomas Köhler\inst{1}\and
Michael Steiner\inst{1}\and
Markus Steinberger\inst{1,2}
}

\authorrunning{L.~Radl et al.}

\institute{Graz University of Technology, Austria
\and
Huawei Technologies, Austria\\
\url{https://r4dl.github.io/CoMe/}
}

\maketitle

\begin{center}
\includegraphics[width=0.93\textwidth]{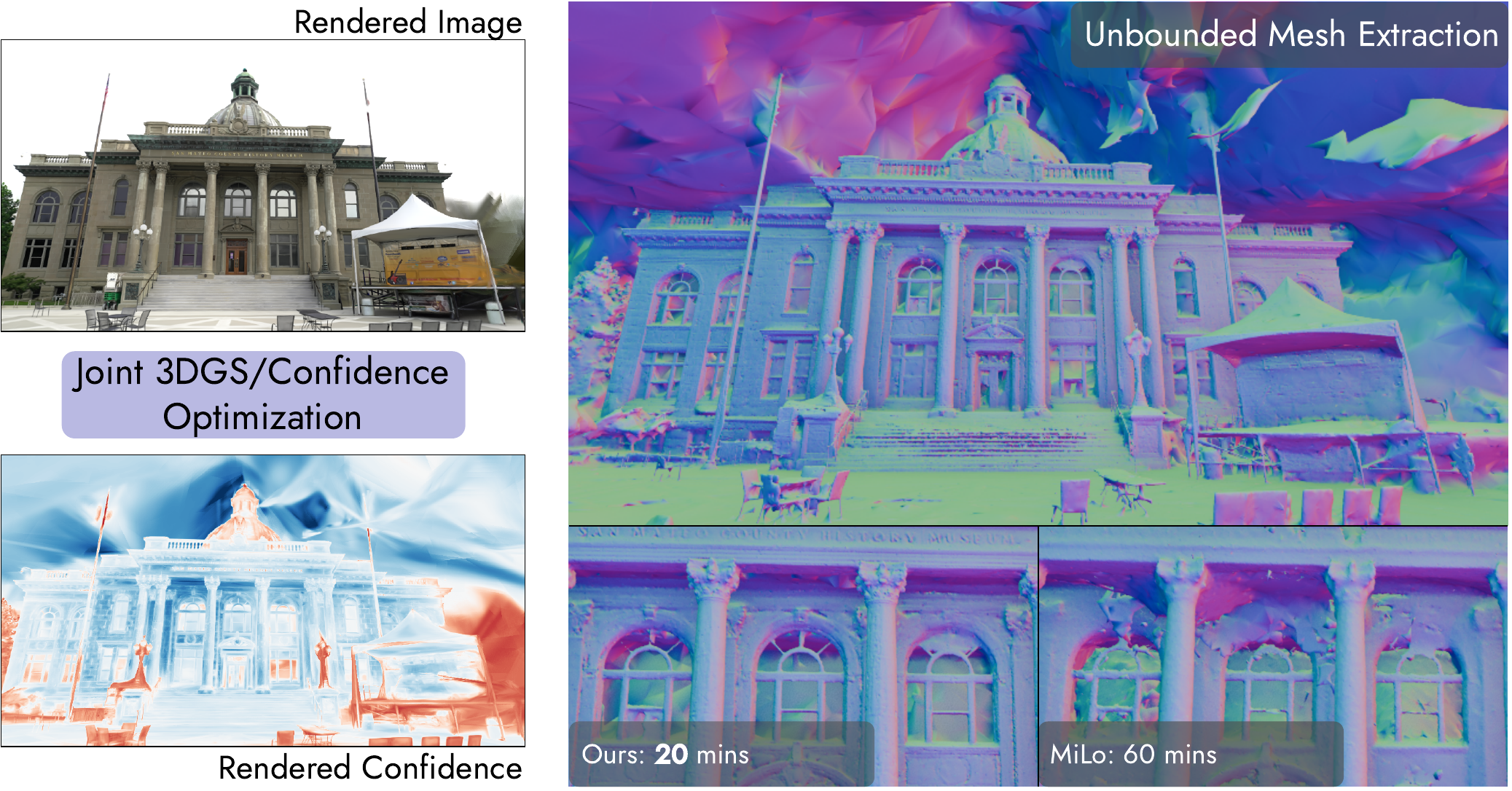}
    \captionof{figure}{\textbf{Teaser:}
We propose a novel, confidence-based method to extract meshes from 3D Gaussians.
Each Gaussian is equipped with additional confidence values that balance photometric and geometric losses in a self-supervised manner.
Compared to related work, our final meshes exhibit finer details and fewer artifacts.
} 
    \label{fig:teaser}
\end{center}%

\begin{abstract}
Recently, 3D Gaussian Splatting (3DGS) greatly accelerated mesh extraction from posed images due to its explicit representation and fast software rasterization. 
While the addition of geometric losses and other priors has improved the accuracy of extracted surfaces, mesh extraction remains difficult in scenes with abundant view-dependent effects.
To resolve the resulting ambiguities, prior works rely on multi-view techniques, iterative mesh extraction, or large pre-trained models, sacrificing the inherent efficiency of 3DGS.
In this work, we present a simple and efficient alternative by introducing a self-supervised confidence framework to 3DGS:
within this framework, learnable confidence values dynamically balance photometric and geometric supervision.
Extending our confidence-driven formulation, we introduce losses which penalize per-primitive color and normal variance and demonstrate their benefits to surface extraction. 
Finally, we complement the above with an improved appearance model, by decoupling the individual terms of the D-SSIM loss.
Our final approach delivers state-of-the-art results for unbounded meshes while remaining highly efficient.
\end{abstract}
\section{Introduction}
\label{sec:intro}
Accurate reconstruction of high-quality surfaces from multi-view captures is a long-standing problem in Computer Vision.
Recent progress in this field has mostly been driven by Novel View Synthesis methods such as Neural Radiance Fields (NeRF)~\cite{Mildenhall2020NeRF} and, more recently, 3D Gaussian Splatting (3DGS)~\cite{kerbl20233dgs}.
Most prior art optimizes a scene representation driven by photometric losses, with additional geometric priors during optimization; 
subsequently, a tailored mesh extraction algorithm is applied post-hoc to extract the final surface.

The improvements in speed and accuracy delivered by 3DGS have led to a recent surge in surface reconstruction methods~\cite{chen2024pgsr, guedon2023sugar, huang20242dgs, yu2024gof}.
The main difficulty to overcome in this specific setting is the inherent \emph{coupling} of geometry and appearance within 3DGS (\ie, geometric primitives are endowed with appearance features):
This implies that the appearance may be modified directly or indirectly by updates to the underlying geometry. 
In challenging scenarios with high-frequency view-dependent appearance, updating the geometry may, in fact, represent the only viable way to truly minimize the photometric error;
particularly since Spherical Harmonics (SH), as currently employed, are limited to low-frequency components~\cite{disario2025sphericalvoronoi}.
For instance, certain high-frequency view-dependent effects are often represented as highly opaque Gaussians behind a semi-transparent surface, which are only visible from certain viewpoints.


Recent work has already identified these challenges and proposed various solutions.
Some methods~\cite{chen2024pgsr, li2025vags, zhang2025qgs} rely on multi-view geometric and photometric constraints, which effectively constrain the geometry of Gaussians.
Other works~\cite{chen2024vcr, li2025geosvr, wang2025UAGS} rely on large, pre-trained monocular models for either normals or depth, enabling stronger supervision compared to the commonly used geometric constraints.
Another, more recent line of work~\cite{guedon2025milo} proposes mesh-aware optimization strategies, enforcing bidirectional consistency between Gaussians and the extracted mesh during training.
However, all of these approaches add inherent overhead that negatively impacts computational efficiency.

In this work, we pose the question:
How far can we push surface extraction without relying on any of the previously introduced techniques?
The answer comes in the form of a self-supervised confidence framework inspired by recent work on Bayesian Networks~\cite{kendall2017uncertainties} and feed-forward models~\cite{wang2024duster}, where learnable per-primitive confidence values dynamically balance photometric and geometric losses.
Further, we introduce confidence-steered densification, which mitigates the over-densification problem, where small Gaussians in difficult-to-reconstruct regions are cloned repeatedly.
Within this self-supervised framework, we identify blending variance as a primary indicator of geometric uncertainty. 
To address this, we introduce color and normal variance losses, further boosting the resulting mesh quality.
To support the optimization of a consistent 3D scene representation, we investigate commonly used appearance embeddings that compensate for varying illumination and propose a novel D-SSIM decoupling mechanism that improves mesh quality driven by a more consistent 3D scene representation.

Our \emph{contributions} can be summarized as follows:
\begin{itemize}
    \item We introduce a self-supervised confidence framework to 3D Gaussian Splatting, along with densification adaptations, improving surface reconstruction
    \item We demonstrate how reducing variance in color- and normal blending resolves visual and geometric ambiguities, thereby improving the accuracy of surfaces
    \item Motivated by an analysis of photometric losses in 3DGS, we propose a novel decoupled appearance embedding, improving reconstruction in challenging real-world datasets
\end{itemize}
Together, these purely self-supervised contributions allow us to achieve state-of-the-art performance for unbounded mesh extraction, while strictly avoiding the computational overhead of prior-dependent methods to maintain a highly competitive runtime.
\section{Related Work}
\label{sec:related}

\para{Radiance Field Methods}
Kerbl \emph{et al.} revolutionized novel view synthesis through 3D Gaussian Splatting (3DGS)~\cite{kerbl20233dgs}, which usurped previous implicit and hybrid methods~\cite{barron2023zip, Mildenhall2020NeRF, mueller2022instant}.
Recent work has extended this approach by mitigating various rendering artifacts~\cite{radl2024stopthepop, steiner2025aaagaussians, hahlbohm2025perspective, Tu2025VRSplat, Yu2024MipSplatting}, improved densification strategies~\cite{bulo2024revising, kheradmand2024mcmc, mallick2024taming, ye2024AbsGau}, and enhanced view-dependent appearance~\cite{liu2025beta, yang2024spec, disario2025sphericalvoronoi}.
Other methods focused on obtaining more lightweight representations~\cite{mallick2024taming, fang2024minisplatting,papantonakis2024reducing}, or extending 3DGS to support other camera models~\cite{loccoz20243dgrt, wu20253dgut}.
3DGS has also inspired a wave of follow-up works, focused on improving the underlying 3D representation itself~\cite{Sun2024SVRaster, govindarajan2025radfoarm, von2025linprim}.

\para{Uncertainty in Radiance Fields}
The optimization of radiance fields given imperfect, real-world data is inherently underconstrained.
To quantify the resulting uncertainty, existing works leverage Bayesian frameworks:
Variational inference has been used to learn probability distributions over radiance fields~\cite{shen2022conditional, shen2021stochastic, li2024variational}.
Alternatively, Bayes' Rays~\cite{goli2023bayes} estimates the uncertainty after optimization via Laplace approximation, while other works explore density-aware or self-ensembling techniques~\cite{sunderhauf2022density, zhao2025GSEnsemble}.
While these methods have proven effective for active reconstruction --- where agents are guided towards high-uncertainty regions~\cite{jin2025activegs, Jiang2024FisherRF, Xue2024NVF} --- applications to surface reconstruction remain underexplored:
Both UA-GS~\cite{wang2025UAGS} and VCR-GauS~\cite{chen2024vcr} use confidence to balance the influence of inconsistent, independently predicted pseudonormals.
In contrast, our self-supervised confidence loss balances the influence of photometric and geometric losses during optimization, without relying on large, pre-trained networks.

\para{Compensating for Varying Appearance in Radiance Fields}
Compensating for camera-specific image sensor processing (ISP) is central to all 3D reconstruction tasks from images.
NeRF in the Wild~\cite{martin2021nerfw} repurposes generative latent optimization~\cite{bojanowski2017optimizing} to compensate for variations in capture time;
this approach was adapted by follow-up works extending 3DGS to in-the-wild data~\cite{dahmani2024swag, kulhanek2024wildgaussian, zhang2024gaussianwild}.
For less extreme variations, lighter models may also be employed.
Both VastGaussian~\cite{lin2024vastgaussian} and Hierarchical 3DGS~\cite{kerbl2024hierarchical} compensate for exposure variations for large-scale data;
interestingly, the appearance model from VastGaussian has been widely adopted in surface reconstruction methods~\cite{yu2024gof, Radl2025SOF, guedon2025milo}.
Wang \emph{et al.} extend learnable affine mappings with bilateral grids~\cite{wang2024bilateral} to compensate for varying illumination.
ADOP~\cite{rueckert2022adop} models a camera compensation module as a learnable post-process, whereas NExF~\cite{niemeyer2025nexf} elevates exposure prediction to 3D.
Concurrently, PPISP~\cite{deutsch2026ppisp} proposed a physically-motivated camera compensation module with a custom controller module, allowing for inference of ISP properties during inference.
In contrast, we analyze and improve the commonly used appearance embeddings for mesh extraction, which we demonstrate to vastly improve surface reconstruction metrics.

\para{Surface Extraction from 3D Gaussians}
Extracting meshes from 3DGS is usually achieved by incorporating geometric priors during optimization and extracting a mesh as a post-hoc process.
Such geometric priors may come in the form of surface-alignment~\cite{guedon2023sugar}, multi-view constraints~\cite{chen2024pgsr, li2025vags}, monocular priors~\cite{chen2024vcr, li2025geosvr, li2025vags}, or the use of Gaussian surfels directly~\cite{huang20242dgs, Dai2024GaussianSurfels}.
These baseline approaches have been extended with rasterized depth~\cite{zhang2024rade} or quadric surfaces~\cite{zhang2025qgs}.
Some works establish an opacity field in 3D to use marching tetrahedra for unbounded meshes~\cite{Radl2025SOF, yu2024gof}.
Another recent line of work establishes bidirectional consistency between meshes and Gaussians~\cite{guedon2025milo}.
Alternatively, some works jointly optimize a neural SDF for mesh extraction~\cite{yu20243dgssdf, zhang2024GSPull, lyu20243dgsr, li2025udf}, while others leverage sparse voxel rasterization~\cite{Sun2024SVRaster, li2025geosvr}.
Another class of techniques explores stochastic frameworks for surface reconstruction~\cite{chen2024dipole, jiang2025geometryfieldsplatting, miller2024objectsvolumesstochasticgeometry}.
Building on these prior works, we demonstrate state-of-the-art unbounded mesh extraction.


\section{Method}

\subsection{Preliminaries}
3DGS~\cite{kerbl20233dgs} represents scenes using $N$ anisotropic 3D Gaussians. 
Each primitive is characterized by a position $\bm{\mu}_i \in \Rbb^{3}$, orientation $\bm{R}_i \in SO(3)$, diagonal scaling matrix $\bm{S}_i \in \text{diag}(\Rbb_+^{3})$, opacity $o_i \in [0,1]$, and spherical harmonics (SH) coefficients $\bm{\theta}_i$. 
The covariance matrix is given by $\bm{\Sigma}_i = \bm{R}_i \bm{S}_i \bm{S}_i^T \bm{R}_i^T$.

To render a pixel, Gaussians are sorted front-to-back along the corresponding ray $\bm{r}(t) = \bm{o} + t\bm{d}$ and accumulated via alpha-blending:
\begin{equation}
    \irender(\bm{r}) =\sum_{i=0}^{N-1}   w_i(\bm{r})\ \text{sh}(\bm{\theta}_i, \bm{d}),
     \label{eq:alphablending}
\end{equation}
where $w_i(\bm{r}) = \alpha_i(\bm{r}) \prod_{j=0}^{i-1} (1 - \alpha_j(\bm{r}))$.
Following recent works~\cite{hahlbohm2025perspective, steiner2025aaagaussians, yu2024gof}, we evaluate the opacity $\alpha_i(\bm{r})$ directly in 3D at the point $\bm{x}^*_i$ of maximum contribution along the ray:
\begin{equation}
    \alpha_i(\bm{r}) = o_i \exp\left(-\frac{1}{2}
    \left(\bm{x}^*_i - \bm{\mu}_i\right)^T
    \bm{\Sigma}_i^{-1}
    \left(\bm{x}^*_i - \bm{\mu}_i\right)
    \right),
\end{equation}
where $\bm{x}^*_i$ is analytically found by minimizing the Mahalanobis distance between the ray and the Gaussian center:
\begin{equation}
    \bm{x}^*_i = \bm{o} + 
    \frac{\bm{d}^T\bm{\Sigma}_i^{-1}\left(\bm{\mu}_i - \bm{o}\right)} 
    {\bm{d}^T \bm{\Sigma}_i^{-1} \bm{d}}
    \bm{d}.\label{eq:max_point_eval}
\end{equation}
%
The parameters are optimized to minimize a photometric reconstruction loss,
which is a combination of $\mathcal{L}_1$ and D-SSIM~\cite{Wang2004Structural} terms:
\begin{equation}
    \msub{\mathcal{L}}{rgb} = (1 - \msub{\lambda}{rgb}) \mathcal{L}_1(\igt, \irender) + \msub{\lambda}{rgb} \msub{\mathcal{L}}{D-SSIM}(\igt, \irender),\label{eq:photometric_loss}
\end{equation}
with $\msub{\lambda}{rgb} = 0.2$.
Here, $\msub{\mathcal{L}}{D-SSIM}$ is the dissimilarity, which can be re-written as
\begin{equation}
    \msub{\mathcal{L}}{D-SSIM}(\igt, \irender) = 1 - l(\igt, \irender) \cdot c(\igt, \irender) \cdot s(\igt, \irender),\label{eq:dssim}
\end{equation}
with $l,c,s$ the luminance, contrast, and structure terms, respectively. 
We refer the interested reader to~\cite{nilsson2020understandingssim} for an in-depth analysis of SSIM. 

\para{Appearance Modeling}
The appearance modeling approach of VastGaussian~\cite{lin2024vastgaussian} consists of per-image latents $\bm{\rho}_i \in \Rbb^{64}$, which are concatenated to the downsampled version of $\irender$.
A convolutional neural network $F_\Theta$ then predicts a full-resolution, per-channel corrective mapping:
\begin{equation}
    \mathbf{M}_i = F_\Theta\left([\text{ds}_{32}(\irender),\ \bm{\rho}_i]\right),\label{eq:vastgaussian_model}
\end{equation}
with $\mathbf{M}_i \in \Rbb^{H \times W \times 3}$, $\text{ds}_{32}(\cdot)$ denoting downsampling by a factor of $32$ and $[\cdot, \cdot]$ denoting channel-wise concatenation, where $\bm{\rho}_i$ is first spatially broadcast.
The rendered image is then transformed according to
\begin{equation}
    \iapp = \irender \odot \sigma\left(\mathbf{M}_i\right),\label{eq:vastgaussian_mapping}
\end{equation}
with $\sigma(\cdot)$ denoting the sigmoid activation.\footnote{The use of the sigmoid activation is following the re-implementation from GOF~\cite{yu2024gof}; note that the source code for VastGaussian~\cite{lin2024vastgaussian} was never released.}
Importantly, the modified image $\iapp$ is only used for the $\mathcal{L}_1$-loss term in \cref{eq:photometric_loss}. 

\subsection{Improved Decoupled Appearance}
Compensating for the camera-specific image sensor processing is central to obtaining a coherent 3D scene representation.
Previous mesh extraction pipelines~\cite{Radl2025SOF, guedon2025milo, yu2024gof} have addressed this by adopting the decoupled appearance model from VastGaussian~\cite{lin2024vastgaussian}, or by directly learning per-image transformation during training~\cite{chen2024pgsr}.
However, both of the aforementioned approaches only consider the modified image for the $\mathcal{L}_1$-loss and use the original rendered image for $\msub{\mathcal{L}}{D-SSIM}$, arguing that SSIM is inherently more structural.

\newcommand{\overlaylabel}[2]{%
    \begin{tikzpicture}
        \node[anchor=south west, inner sep=0] (image) at (0,0) {\includegraphics[width=0.19\linewidth]{#1}};
        \begin{scope}[x={(image.south east)},y={(image.north west)}]
            \node[anchor=south west, 
                  fill=black, fill opacity=0.5, 
                  text opacity=1, 
                  inner sep=2pt, 
                  rounded corners=1pt] 
            at (0.01, 0.01) { 
                \sffamily\scriptsize{\textcolor{white}{#2}} 
            };
        \end{scope}
    \end{tikzpicture}%
}

\newcommand{\errorbox}[2]{%
    \stackinset{l}{1pt}{b}{1pt}{
        \begin{tikzpicture}
            \node[fill=black, fill opacity=0.4, text opacity=1, 
                  inner sep=1.5pt, rounded corners=0.5pt] 
            {\sffamily\fontsize{4}{5}\selectfont{\textcolor{white}{#2}}};
        \end{tikzpicture}%
    }{#1}
}

\begin{figure*}[ht!]
\scriptsize\sffamily
\setlength{\tabcolsep}{1pt}%
\setlength{\fboxsep}{0pt}%
\setlength{\fboxrule}{0.5pt}%
\renewcommand{\arraystretch}{1.1}%
\resizebox{.99\linewidth}{!}{
\begin{tabular}{ccccc}
Render & $\mathcal{L}_1$-loss & Luminance & Contrast & Structure 
\\[-0.75pt]
\overlaylabel{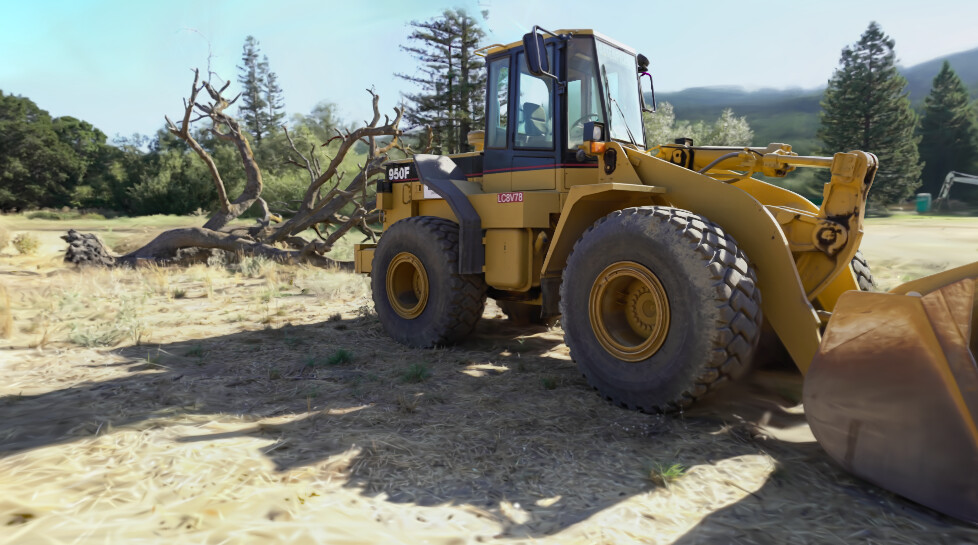}{$\irender$} &
\errorbox{\includegraphics[width=0.19\linewidth]{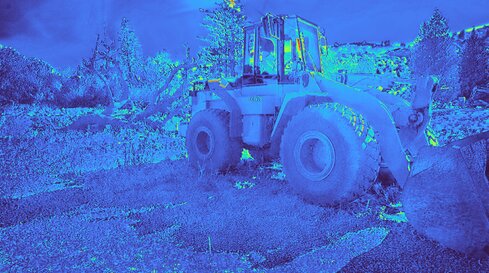}}{0.090} &
\errorbox{\includegraphics[width=0.19\linewidth]{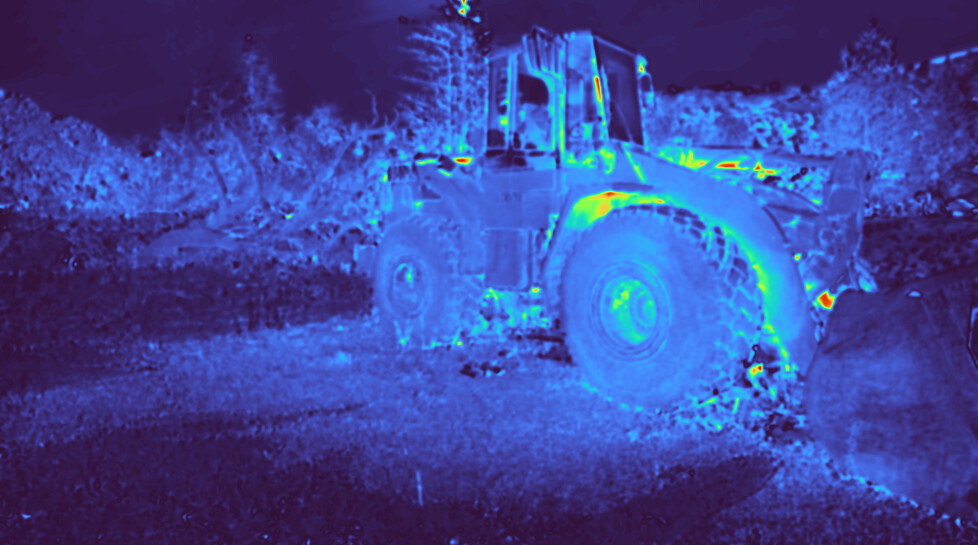}}{0.027} &
\errorbox{\includegraphics[width=0.19\linewidth]{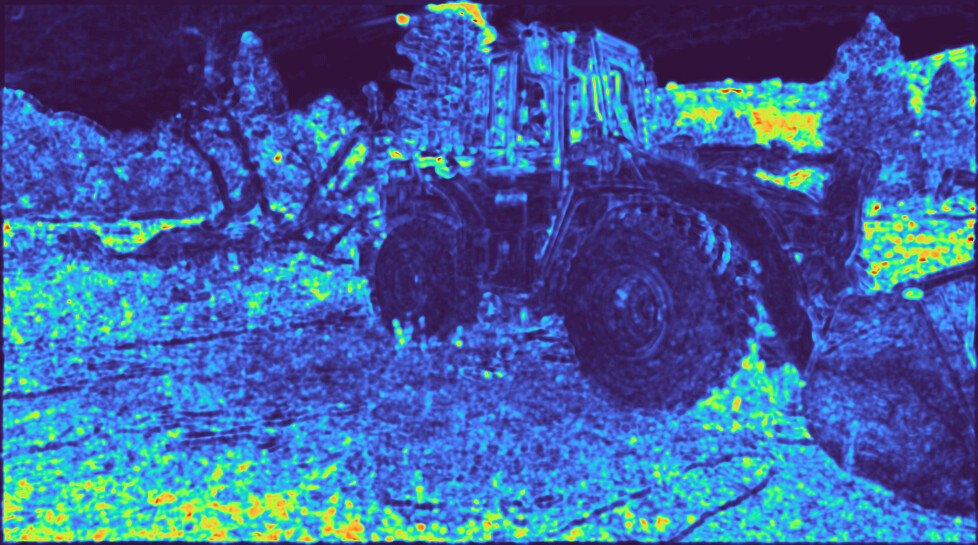}}{0.167} &
\errorbox{\includegraphics[width=0.19\linewidth]{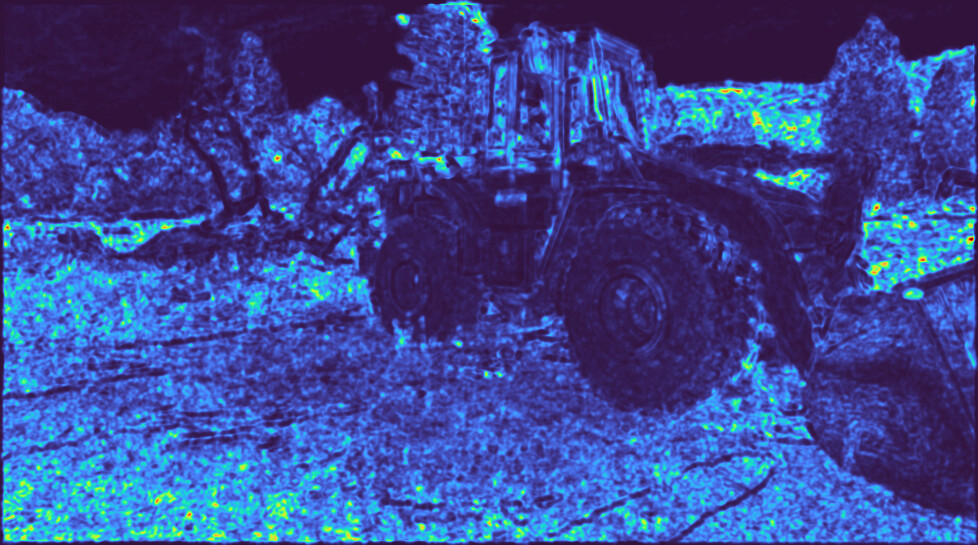}}{0.141} \\ \addlinespace[1pt]

\overlaylabel{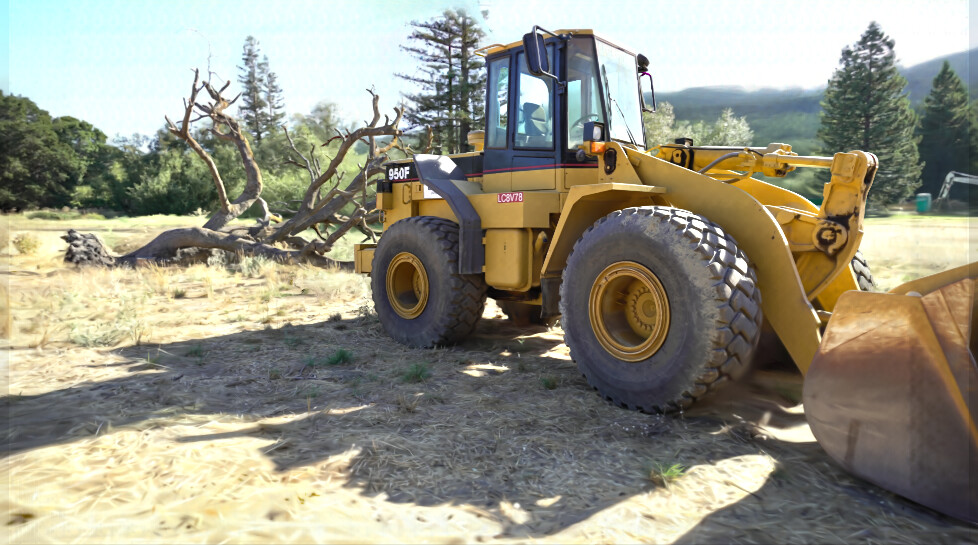}{$\iapp$} &
\errorbox{\includegraphics[width=0.19\linewidth]{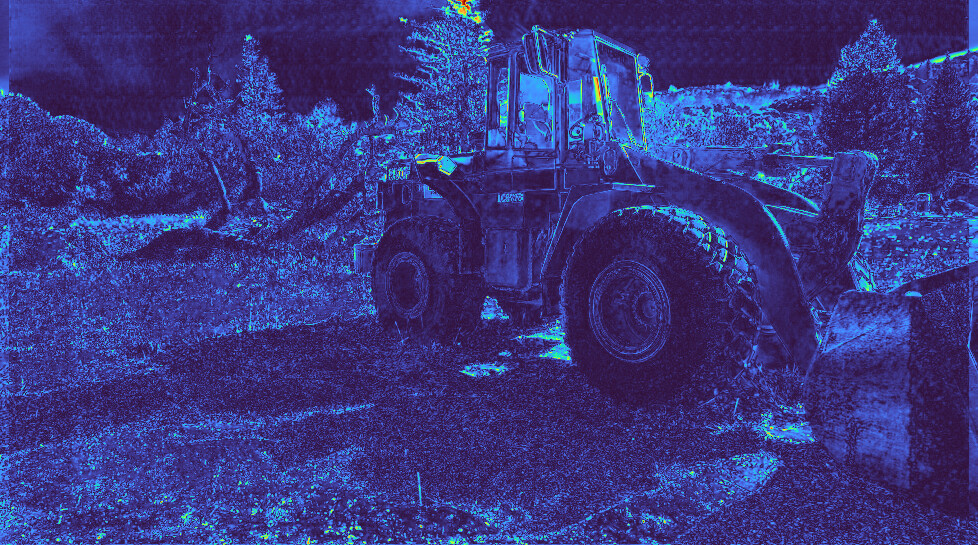}}{0.036} &
\errorbox{\includegraphics[width=0.19\linewidth]{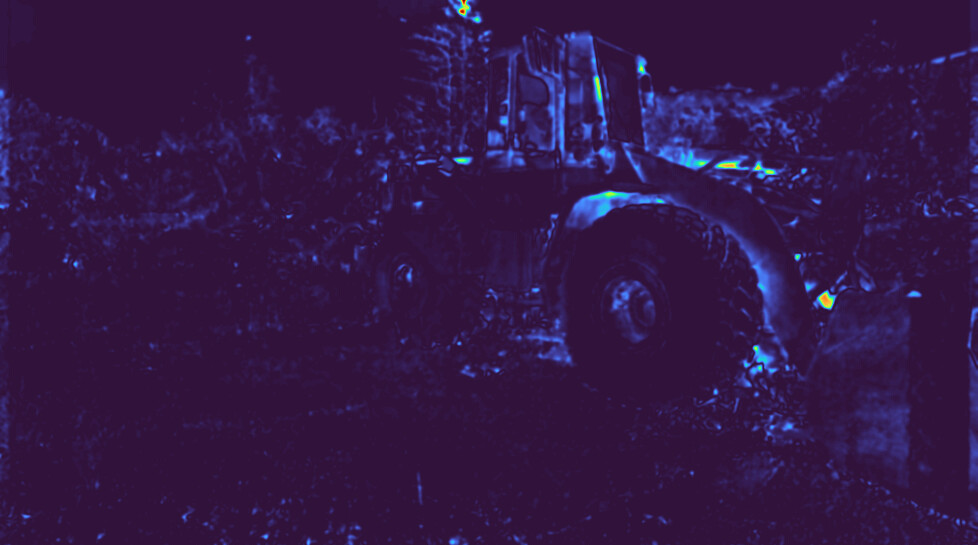}}{0.003} &
\errorbox{\includegraphics[width=0.19\linewidth]{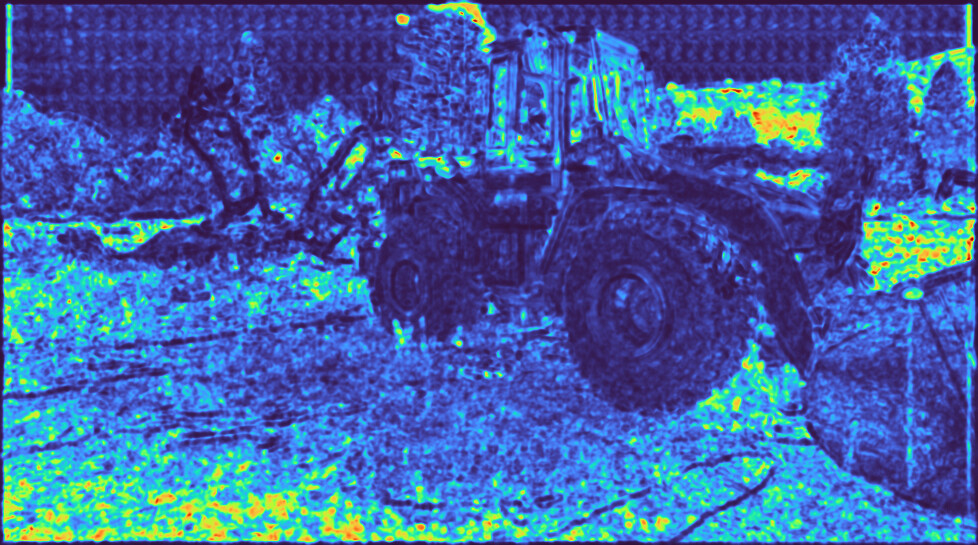}}{0.172} &
\errorbox{\includegraphics[width=0.19\linewidth]{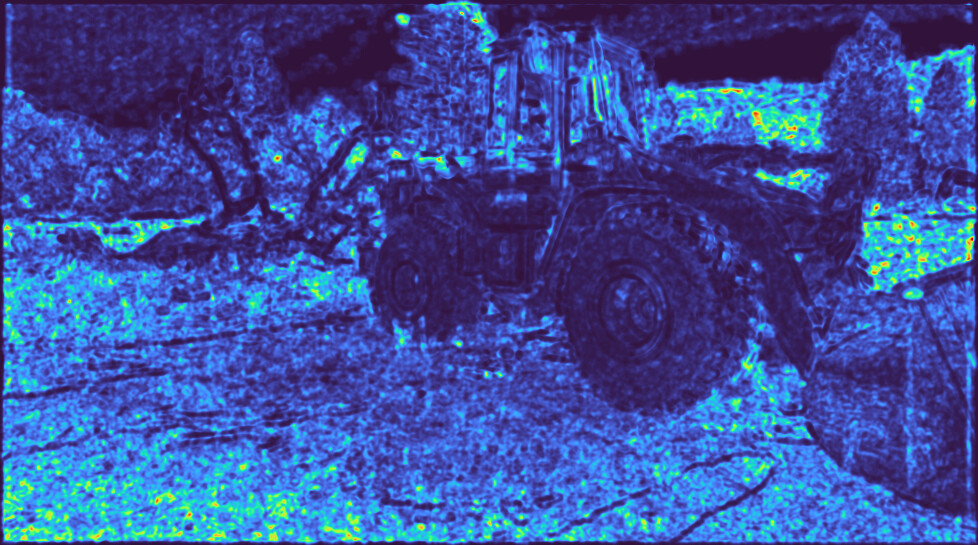}}{0.137} \\
\end{tabular}
}
  \caption{\label{fig:d-ssim}%
    \textbf{Analysis of the Photometric Loss Components}:
The luminance term of D-SSIM is highly dependent on the illumination; in this example, the luminance error is reduced by a factor of $9\times$ by considering the appearance embedding.
  }
\end{figure*}



Upon revisiting \cref{eq:dssim}, we find that the luminance term in particular is not structural, and dominated by variations in illumination; 
see \cref{fig:d-ssim} for a visualization.
Therefore, we use $\iapp$ for the luminance term, effectively modifying \cref{eq:dssim} to
\begin{equation}
    \msub{\mathcal{L}}{D-SSIM}^{\text{dec}} = 1 - l(\igt, \iapp) \cdot c(\igt, \irender) \cdot s(\igt, \irender).\label{eq:dssim_mod}
\end{equation}
As we will demonstrate, using the modified image \emph{only in the luminance-term} delivers superior results; 
intuitively, as the contrast and structure terms are illumination-invariant, using $\iapp$ for these terms may lead to the appearance model compensating for incorrect renderings.
Moreover, when considering the $\iapp$ for these terms, artifacts due to CNN upsampling (as exhibited in \cref{fig:d-ssim}) may pollute gradient flow.

We additionally improved the original implementation from VastGaussian~\cite{lin2024vastgaussian}, unlocking a minor performance boost in itself. 
We refer to \cref{app:decoupled} for details.

\subsection{Confidence-Aware Gaussian Splatting}
Previous approaches for surface extraction struggle with balancing photometric and geometric losses; 
in regions with high-frequency view-dependent appearance, photometric losses dominate, and large gradient magnitudes lead to over-densification, where small Gaussians are repeatedly cloned.
In such regions of high-primitive density, 3DGS is particularly prone to faking view-dependent appearance with geometry.
\begin{figure}[ht!]
    \centering
\begin{minipage}{0.9\linewidth}
        \centering
        \stackon[1.5pt]{\includegraphics[width=0.32\linewidth]{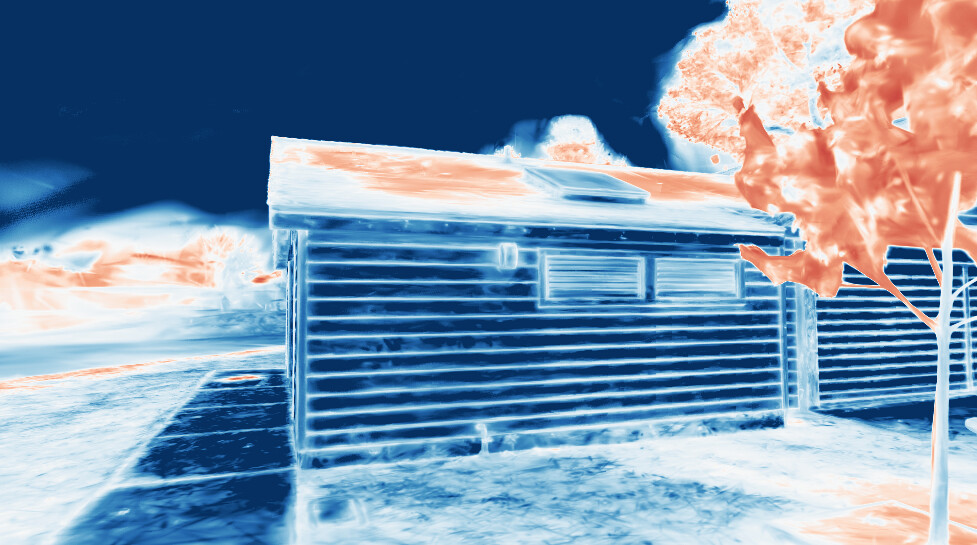}}{\scriptsize {Barn}}
        \stackon[1.5pt]{\includegraphics[width=0.32\linewidth]{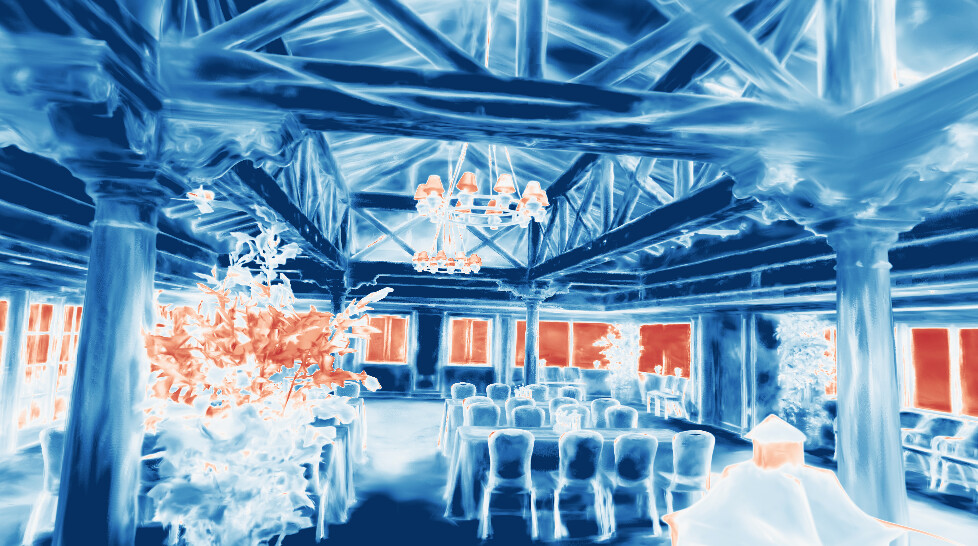}}{\scriptsize {Meetingroom}}
        \stackon[1.5pt]{\includegraphics[width=0.32\linewidth]{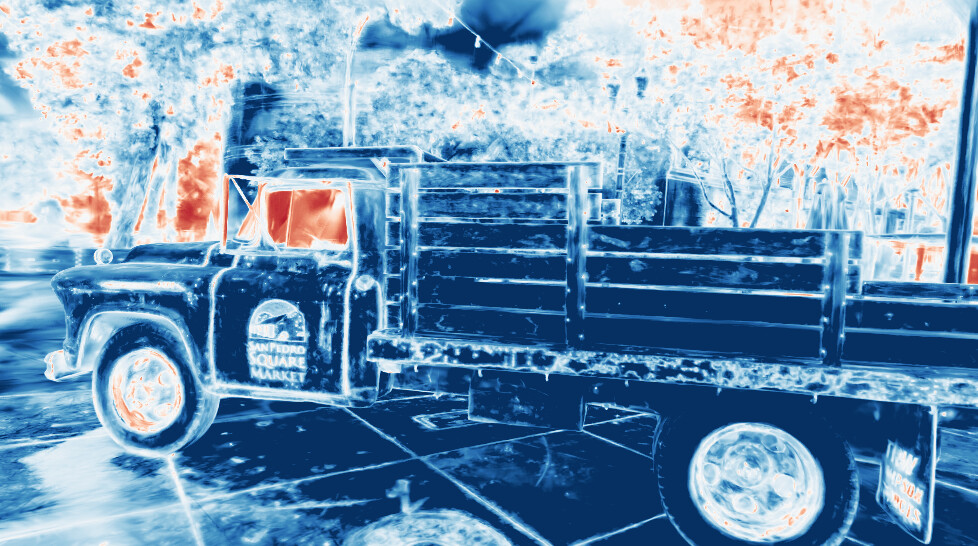}}{\scriptsize {Truck}} \\
        \vspace{1mm} 
        \includegraphics[width=0.32\linewidth]{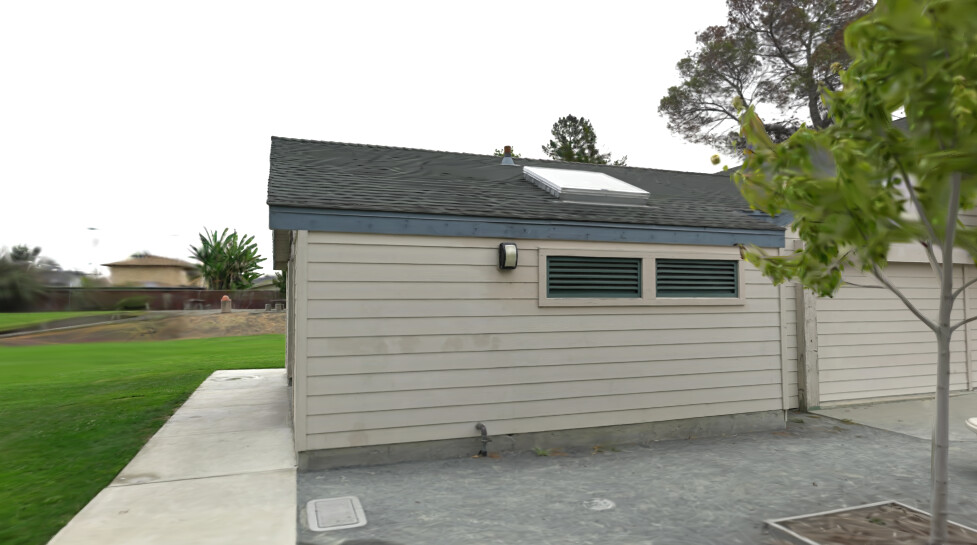}
        \includegraphics[width=0.32\linewidth]{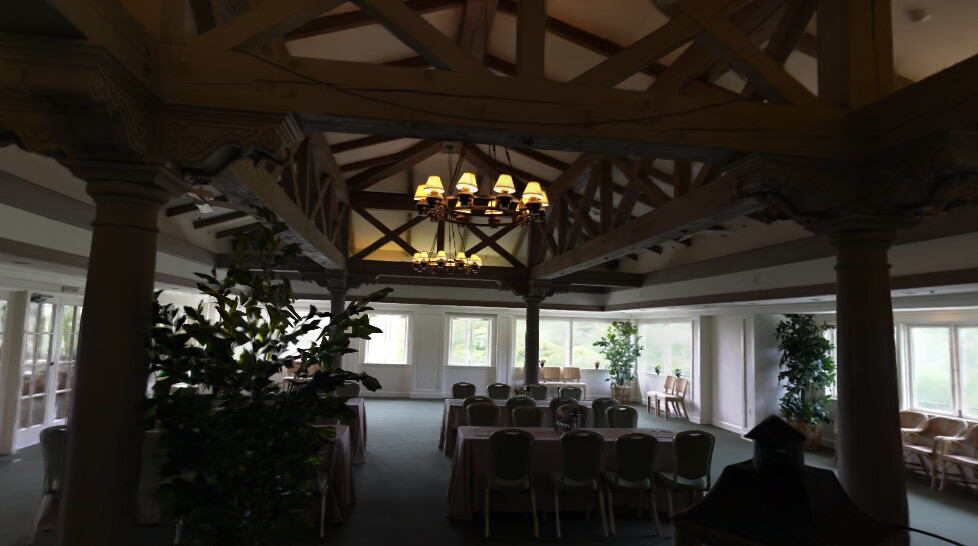}
        \includegraphics[width=0.32\linewidth]{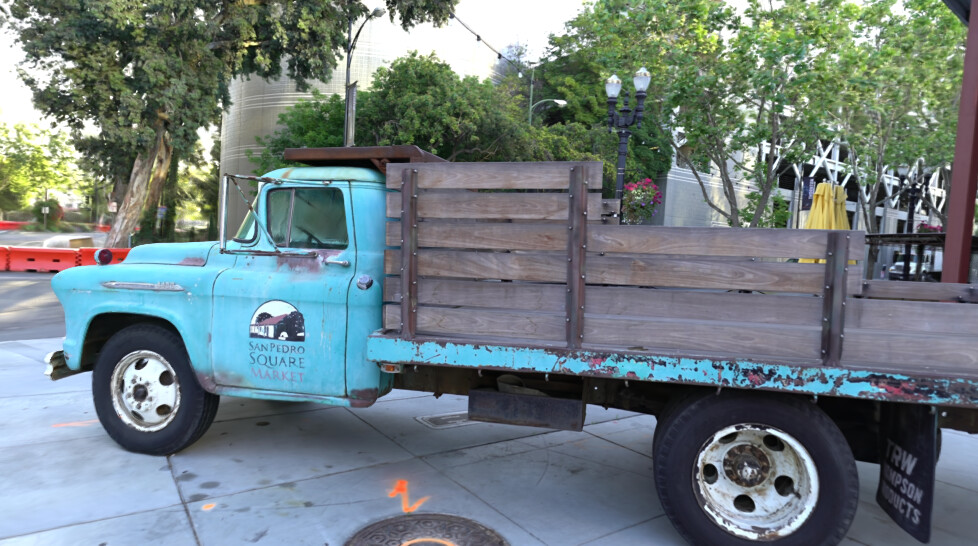}
    \end{minipage}%
    \begin{minipage}{0.08\linewidth}
        \centering
        \includegraphics[height=0.22\textheight]{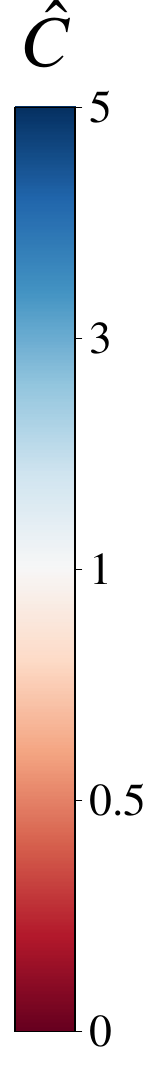} 
    \end{minipage}
    \caption{
\textbf{Confidence-Driven Gaussian Splatting:}
We show rendered images from our trained method, accompanied by the rendered confidence maps $\hat{C}$.
The confidence maps effectively isolate reflective surfaces, thin foliage, or rarely observed areas (such as the roof for {Barn}).
Importantly, the low-confidence regions are still well-reconstructed.
}
    \label{fig:confidence}
\end{figure}

To compensate for this, we take inspiration from recent feed-forward reconstruction models~\cite{wang2024duster} and weigh our photometric loss by a learned confidence value $\hat{C} \in \Rbb_+^{H\times W}$:
\begin{equation}
    \msub{\mathcal{L}}{conf} = \msub{\mathcal{L}}{rgb} \cdot \hat{C} - \beta \cdot \log \hat{C}, \label{eq:conf_loss}
\end{equation}
where $\beta \in \Rbb_+$ is a parameter balancing both terms.
When $\hat{C} = 1$, the loss reduces to $\msub{\mathcal{L}}{rgb}$.
Intuitively, this allows for trading photometric loss for uncertainty when $\hat{C} < 1$.
Importantly, the geometric losses remain untouched by this formulation; thus, the rendered confidence can be used to balance both terms.

\para{Rendering Confidence}
We endow each primitive with a learnable scalar value $\conf_i$, initialized to $0$, from which we derive the confidence $\confa_i$ using:
\begin{equation}
    \confa_i = \exp\left(\conf_i\right),
\end{equation}
ensuring an initial value of $1$.
We derive the final rendered confidence map $\hat{C}$ via alpha-blending (\cf \cref{fig:confidence} for a visualization):
\begin{equation}
    \hat{C}(\bm{r}) = \sum_{i=0}^{N-1} w_i(\bm{r}) \ \confa_i. \label{eq:conf_alphablending}
\end{equation}

\para{Confidence-Aware Densification}
Clearly, the confidence-based balancing from \cref{eq:conf_loss} changes the magnitude of the loss, and may even lead to negative values for $\msub{\mathcal{L}}{conf}$, when confidence is large and $\msub{\mathcal{L}}{rgb}$ approaches $0$. 
Because the densification scheme is built on the magnitudes of per-Gaussian gradients~\cite{ye2024AbsGau, yu2024gof}, it needs to be adapted.

In particular, we want to prevent over-densification for difficult-to-reconstruct regions;
these are usually characterized by low $\hat{C}$.
Within gradient densification, primitives are usually split or cloned when the accumulated positional gradients exceed a threshold $\msub{\tau}{grad} \in \Rbb_+$.
To discourage unconfident Gaussians from splitting, we divide this value by the per-primitive confidence:
\begin{figure}[ht!]
    \centering
    \begin{subfigure}[t]{0.48\textwidth}
    \includegraphics[width=0.48\linewidth, trim={6.5cm 0 1.5cm 0},clip]{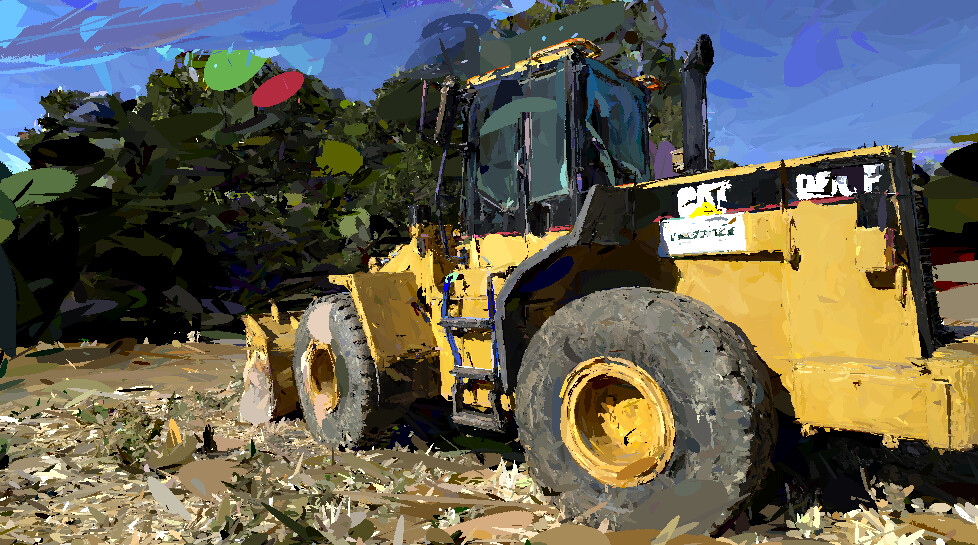}
    \includegraphics[width=0.48\linewidth, trim={6.5cm 0 1.5cm 0},clip]{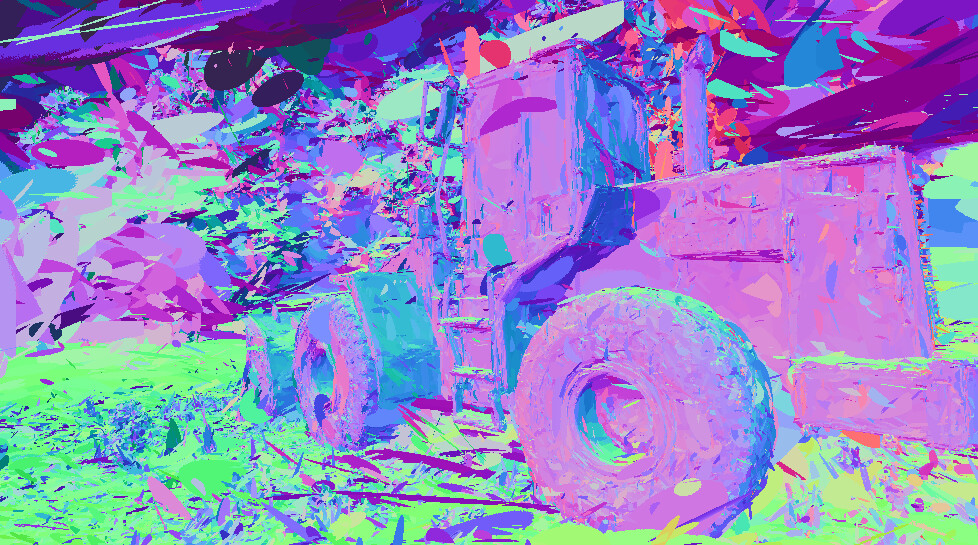}
    \caption{Without variance losses}
\end{subfigure}
    \begin{subfigure}[t]{0.48\textwidth}
    \includegraphics[width=0.48\linewidth, trim={6.5cm 0 1.5cm 0},clip]{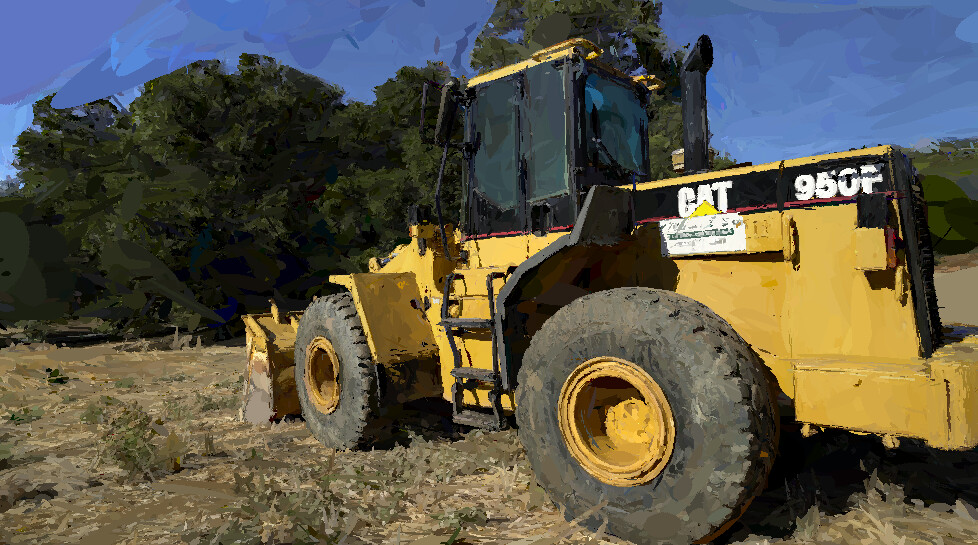}
    \includegraphics[width=0.48\linewidth, trim={6.5cm 0 1.5cm 0},clip]{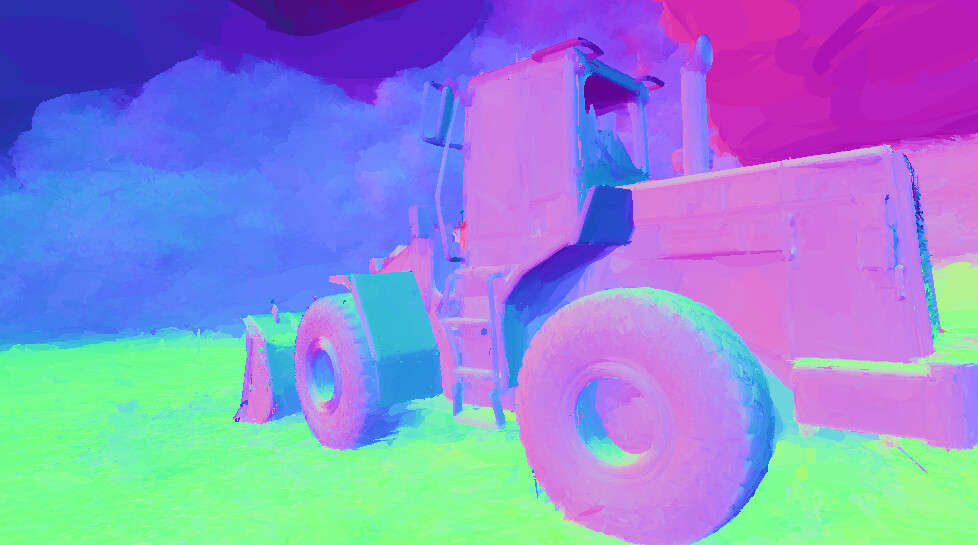}
    \caption{With variance losses}
\end{subfigure}
\caption{
\textbf{Effect of our proposed Variance Losses:}
Rendering color and normals only for the \emph{first-hit} Gaussians shows how our variance losses align individual Gaussians better to the true object surface, in both color and orientation.
}
\label{fig:variance}
\end{figure}
\begin{equation}
    \msub{\Bar{\tau}}{grad} = \frac{\msub{\tau}{grad}}{\min\left(\confa_i, 1\right)}.
\end{equation}
Importantly, we clamp the divisor by $1$ to ensure that confident Gaussians are not over-densified (\ie $\msub{\Bar{\tau}}{grad} \geq \msub{\tau}{grad}$).
In practice, we find that these modifications slightly reduce the overall primitive count when $\beta$ is set to a reasonable value.

\subsection{Variance Reducing Losses for Blending}
While confidence-steered optimization already improves mesh quality, we still observe cases where view-dependent appearance is represented with geometry;
most frequently, this is characterized by highly opaque primitives just behind surfaces, which are blocked in different amounts by semi-transparent primitives depending on the viewing angle.
This is in part enabled by the use of alpha-blending (see \cref{eq:alphablending}): Since only the blended final color is used to compute the loss, individual primitives have no incentive to correctly model radiance.

Previous work~\cite{ziyi2025radiancesurfaces} replaces the loss function of NeRF to apply the photometric loss to the individual color samples along the ray instead of the blended result, which effectively constrains such extreme cases.
We find that na\"ively applying this method to 3DGS significantly deteriorates reconstruction quality, because the vital D-SSIM loss term cannot be applied per-primitive. 
Instead, we introduce an additional weighted per-primitive $\mathcal{L}_2$-loss with the ground truth pixel color $\igt$ to complement the standard 3DGS training loss~\cite{jena2025sparfels}:
\begin{equation}
    \msub{\mathcal{L}}{color-var} = \sum_{i=0}^{N-1} w_i(\bm{r})  \left\lVert\text{sh}(\bm{\theta}_i, \bm{d})-\igt\right\lVert_2^2 . \label{eq:color_var_loss}
\end{equation}
Under the assumption that $\igt$ represents the rendered color, this is equivalent to minimizing the blended color variance along the ray. 
For a more in-depth exploration of the correlation between blending variance and confidence in radiance fields, we refer to Ewen \etal~\cite{ewen2025these}.
%

The per-pixel normals used in the depth-normal consistency loss~\cite{huang20242dgs, Radl2025SOF, yu2024gof} are also alpha-blended, resulting in poor alignment of individual Gaussians to the surface that can be remedied with the introduction of a normal variance loss
\begin{equation}
    \msub{\mathcal{L}}{normal-var} = \sum_{i=0}^{N-1} w_i(\bm{r})
    \left\lVert\bm{n}_i-\bm{N}\right\lVert_2^2,\label{eq:normal_var_loss}
\end{equation}
with blended pixel normal $\bm{N} = \sum_{i=0}^{N-1}  w_i(\bm{r}) \ \bm{n}_i$.
The resulting improved color and normal consistency is demonstrated in \cref{fig:variance}.
We note that the \textit{depth distortion} loss~\cite{barron2022mipnerf360} can also be interpreted as a variance loss on the blended depth. 

\section{Results and Evaluation}
\para{Implementation Details}
Due to its relative speed, we build our solution on top of the SOF codebase~\cite{Radl2025SOF}.
For mesh extraction, we also use Marching Tetrahedra with binary search~\cite{yu2024gof} to extract unbounded meshes, using the faster implementation from Radl \emph{et al.}~\cite{Radl2025SOF}.
Our final loss function is:
\begin{equation}
    \mathcal{L} = \msub{\mathcal{L}}{conf} + 
    \msub{\mathcal{L}}{geom} + 
    \msub{\lambda}{color-var} \msub{\mathcal{L}}{color-var} +
    \msub{\lambda}{normal-var} \msub{\mathcal{L}}{normal-var},
\end{equation}
where $\msub{\mathcal{L}}{geom}$ denotes the geometry losses as used in SOF (see~\cite{Radl2025SOF} for details).
We set $\msub{\lambda}{color-var} = 5 \times 10^{-1}$ and $ \msub{\lambda}{normal-var} = 5 \times 10^{-3}$ across all scenes.

For the confidence values $\conf_i$, we use a learning rate of $2.5 \times 10^{-4}$, and set $\beta = 7.5 \times 10^{-2}$ for all scenes.
We enable $\msub{\mathcal{L}}{conf}$ from iteration $500$, right at the start of densification; 
empirically, we have not observed improved training stability when enabling confidence learning later.

\para{Datasets and Comparisons}
To benchmark unbounded mesh extraction, we primarily rely on the Tanks \& Temples dataset~\cite{Knapitsch2017tanks}.
For bounded scenes, we include a more rigorous baseline for real-world surface extraction by conducting extensive evaluations on the high-fidelity ScanNet++ dataset~\cite{yeshwanthliu2023scannetpp}, moving beyond traditional small-scale object benchmarks~\cite{jensen2014large}.
Finally, we evaluate novel view synthesis and mesh extraction on the Mip-NeRF 360 dataset~\cite{barron2022mipnerf360}.

Our main comparison points are GOF~\cite{yu2024gof}, SOF~\cite{Radl2025SOF} and \milo~\cite{guedon2025milo} for unbounded meshes; we also include QGS~\cite{zhang2025qgs} and PGSR~\cite{chen2024pgsr} as representative bounded, multi-view baselines.
We reproduced all results using the latest available codebase.
To ensure a fair and reproducible comparison, we standardized the evaluation protocol across all methods.
Specifically, we adjusted the training and evaluation settings (\eg matching image resolutions and train/test splits) for several baselines (see \cref{app:reproducibility} for details).
Consequently, the reported results do not always reflect the numbers reported by the authors.
Importantly, we excluded methods relying on pre-trained monocular models~\cite{li2025vags, chen2024vcr, li2025geosvr} from this comparison, and refer to \cref{app:geosvr} for a dedicated discussion.

\subsection{Surface Reconstruction}
\para{Unbounded Meshes}
\begin{table}[ht!]
    \centering
    \setlength{\tabcolsep}{4pt}
    \caption{
\textbf{Full Geometry Reconstruction results} for the Tanks \& Temples dataset \cite{Knapitsch2017tanks}.
We evaluate the F1-score (higher is better) and report runtimes, measured on an RTX 4090.
Overall, our method achieves the best reconstruction quality of all tested methods (without leveraging multi-view constraints), while remaining efficient.
}
\resizebox{.98\linewidth}{!}{
\begin{tabular}{lrrrrrrr@{\hskip 0.5cm} lc}
\toprule
 & Barn & Caterpillar & Courthouse & Ignatius & Meetingroom & Truck & Average & Runtime \\
\midrule
PGSR & \cellcolor{tab_color!49}0.548 & \cellcolor{tab_color!32}0.437 & 0.238 & 0.728 & \cellcolor{tab_color!15}0.367 & \cellcolor{tab_color!49}0.658 & \cellcolor{tab_color!32}0.496 & 28min \\
QGS & \cellcolor{tab_color!15}0.536 & 0.374 & 0.183 & 0.733 & \cellcolor{tab_color!49}0.374 & \cellcolor{tab_color!32}0.645 & 0.474 & 41min\\
\midrule
GOF & 0.484 & 0.402 & 0.288 & 0.674 & 0.275 & 0.596 & 0.453 & 40min\\
SOF & 0.535 & \cellcolor{tab_color!15}0.408 & \cellcolor{tab_color!15}0.297 & \cellcolor{tab_color!15}0.736 & 0.309 & 0.558 & 0.474 & 17min\\
\milo & \cellcolor{tab_color!32}0.541 & 0.389 & \cellcolor{tab_color!32}0.322 & \cellcolor{tab_color!32}0.757 & 0.281 & 0.617 & \cellcolor{tab_color!15}0.485 & 60min\\
Ours & 0.534 & \cellcolor{tab_color!49}0.472 & \cellcolor{tab_color!49}0.333 & \cellcolor{tab_color!49}0.782 & \cellcolor{tab_color!32}0.372 & \cellcolor{tab_color!15}0.634 & \cellcolor{tab_color!49}0.521 & 18min\\
\bottomrule
\end{tabular}
}
    \label{tab:tnt_results}
\end{table}

Our main quantitative results concern mesh reconstruction for the Tanks \& Temples dataset~\cite{Knapitsch2017tanks}, which are presented in \cref{tab:tnt_results}.
We report the F1-score for each scene, the overall average, as well as the average optimization time across methods.
Our approach delivers superior overall accuracy, achieving a leading average F1-score of $\mathbf{0.521}$, outperforming all evaluated baselines with a significant margin.
Specifically, our method demonstrates exceptional robustness in challenging scenarios:
our decoupled appearance module effectively resolves the varying illumination present in Caterpillar and Ignatius, obtaining the best scores for these scenes.
Our confidence framework and variance losses result in robustness for particularly challenging scenes, such as Courthouse (where we have $>1100$ input images).
For Meetingroom and Truck, our method attains the highest scores out of all unbounded methods;
here, bounded baselines~\cite{zhang2025qgs, chen2024pgsr} rely on heavy multi-view constraints to regularize underconstrained regions such as the textureless floor.
In summary, our method attains state-of-the-art performance, extracts detailed unbounded meshes, and demonstrates highly competitive runtimes (optimization in $< 20$ minutes).
\begin{table}[ht!]
    \centering
    \setlength{\tabcolsep}{4pt}
    \caption{
\textbf{Full Geometry Reconstruction results} for our ScanNet++-v2 subset \cite{yeshwanthliu2023scannetpp}, where we report the F1-score.
Our method attains the highest scores across all scenes, underscoring our real-world applicability.
}
\resizebox{.98\linewidth}{!}{
\begin{tabular}{lrrrrrrr}
\toprule
 & \texttt{5a269ba6fe} & \texttt{08bbbdcc3d} & \texttt{39f36da05b} & \texttt{c263dfbf0} & \texttt{ef18cf0708} & \texttt{fb564c935d} & Average \\
\midrule
PGSR & \cellcolor{tab_color!32}0.668 & 0.666 & \cellcolor{tab_color!32}0.647 & 0.630 & \cellcolor{tab_color!32}0.542 & 0.632 & \cellcolor{tab_color!32}0.631 \\
QGS & \cellcolor{tab_color!15}0.646 & 0.622 & 0.571 & 0.522 & 0.504 & 0.572 & 0.573 \\
\midrule
GOF & 0.620 & 0.670 & 0.611 & \cellcolor{tab_color!15}0.672 & 0.504 & \cellcolor{tab_color!32}0.661 & 0.623 \\
SOF & 0.599 & \cellcolor{tab_color!32}0.681 & 0.616 & \cellcolor{tab_color!32}0.675 & 0.497 & 0.621 & 0.615 \\
\milo & 0.622 & \cellcolor{tab_color!15}0.675 & \cellcolor{tab_color!15}0.621 & 0.626 & \cellcolor{tab_color!32}0.542 & \cellcolor{tab_color!15}0.660 & \cellcolor{tab_color!15}0.624 \\
Ours & \cellcolor{tab_color!49}0.670 & \cellcolor{tab_color!49}0.729 & \cellcolor{tab_color!49}0.657 & \cellcolor{tab_color!49}0.715 & \cellcolor{tab_color!49}0.551 & \cellcolor{tab_color!49}0.684 & \cellcolor{tab_color!49}0.668 \\
\bottomrule
\end{tabular}
}
    \label{tab:snpp_results}
\end{table}

\para{Bounded Meshes}
We additionally evaluated bounded mesh extraction on a 6-scene subset of the ScanNet++ dataset~\cite{yeshwanthliu2023scannetpp} (see \cref{app:dataset} for selection details). 
We present our quantitative results in \cref{tab:snpp_results}, where we report the per-scene and average F1-score.
Our full method outperforms other works by a significant margin and attains the best performance for all scenes.
Crucially, these results demonstrate the applicability of our method in unconstrained, real-world scenarios: 
ScanNet++ is known to suffer from exposure variation, inconsistent lighting, and occasional blur, which our decoupled appearance module faithfully compensates for.

\begin{figure}[ht!]
    \centering
    
    \begin{tabular}{cccccc}
        \scriptsize{GT} &\scriptsize{Ours} & \scriptsize{\milo} & \scriptsize{SOF} & \scriptsize{QGS} & \scriptsize{PGSR} \\[-1pt]
        
        \includegraphics[width=0.15\textwidth, trim={9cm 0cm 4cm 4cm},clip]{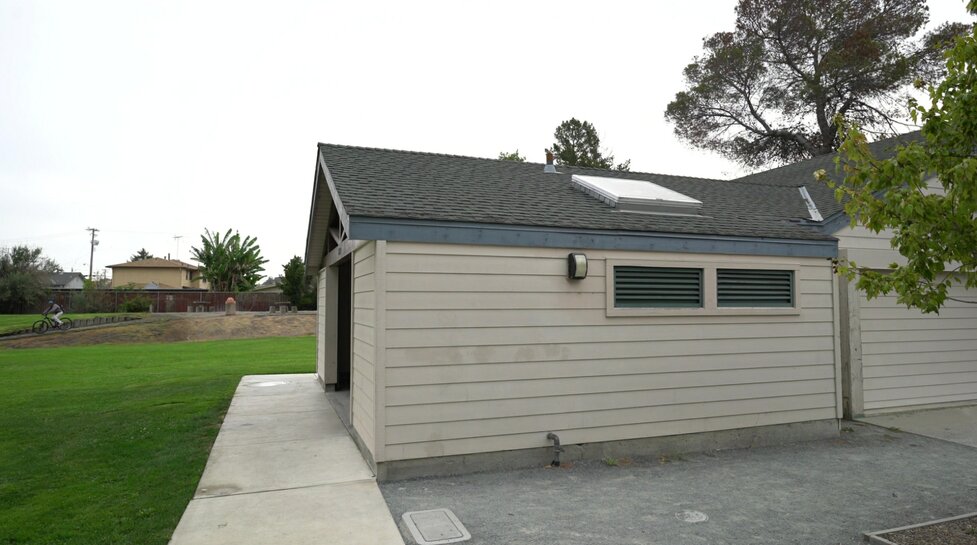} & 
        \includegraphics[width=0.15\textwidth, trim={9cm 0cm 4cm 4cm},clip]{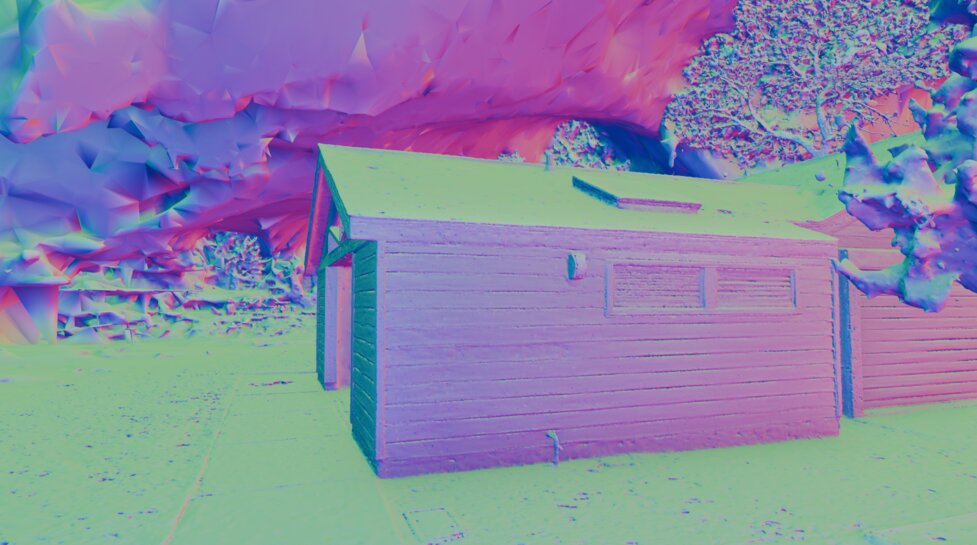} & 
        \includegraphics[width=0.15\textwidth, trim={9cm 0cm 4cm 4cm},clip]{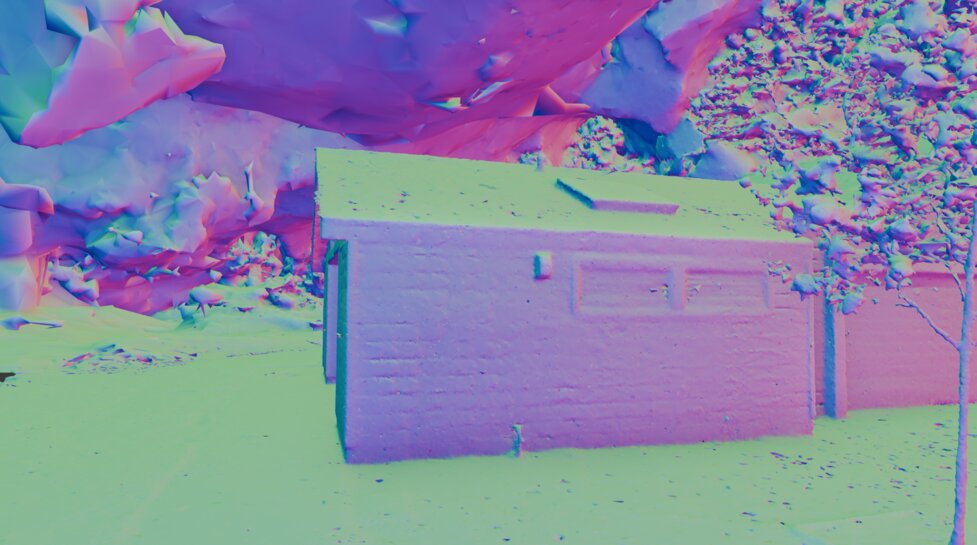} & 
        \includegraphics[width=0.15\textwidth, trim={9cm 0cm 4cm 4cm},clip]{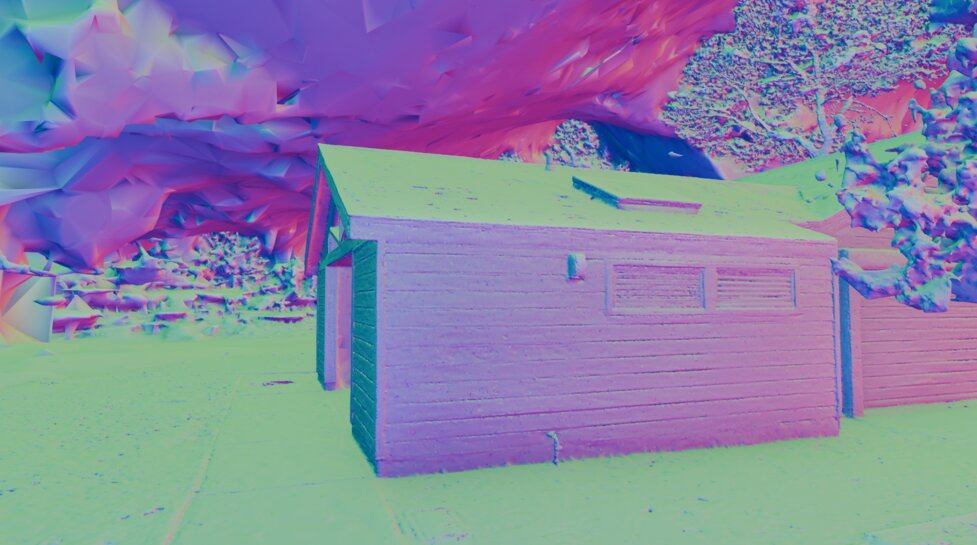} & 
        \includegraphics[width=0.15\textwidth, trim={9cm 0cm 4cm 4cm},clip]{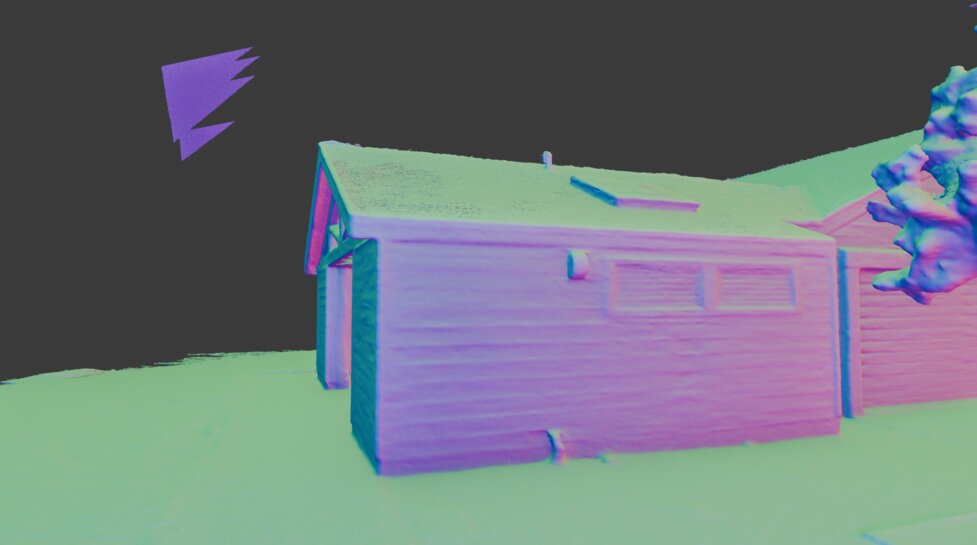} & 
        \includegraphics[width=0.15\textwidth, trim={9cm 0cm 4cm 4cm},clip]{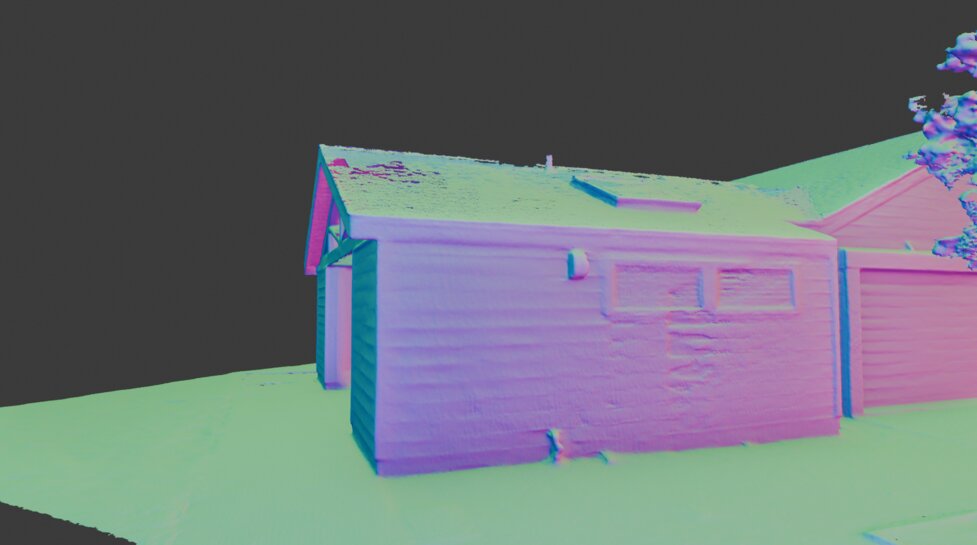} \\
        \includegraphics[width=0.15\textwidth, trim={9cm 5cm 6cm 1.5cm},clip]{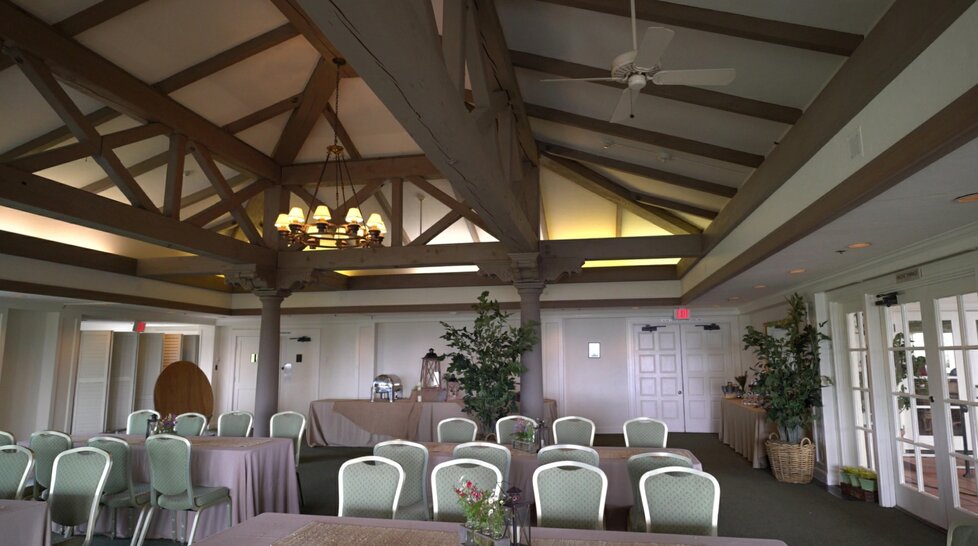} & 
        \includegraphics[width=0.15\textwidth, trim={9cm 5cm 6cm 1.5cm},clip]{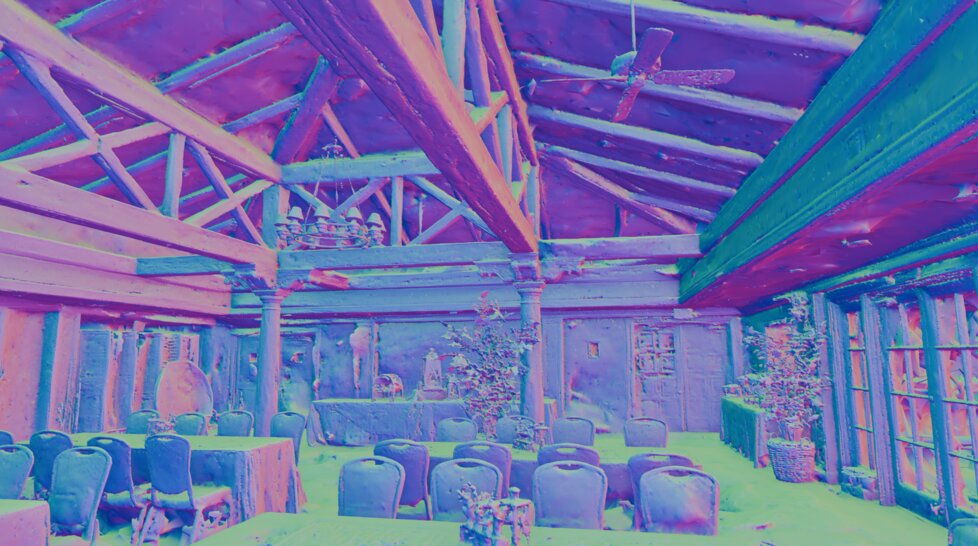} & 
        \includegraphics[width=0.15\textwidth, trim={9cm 5cm 6cm 1.5cm},clip]{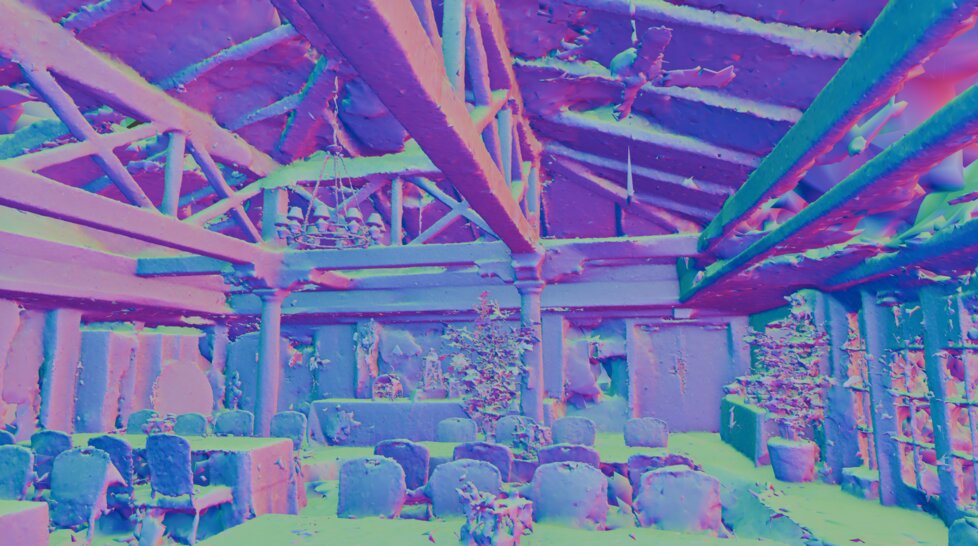} & 
        \includegraphics[width=0.15\textwidth, trim={9cm 5cm 6cm 1.5cm},clip]{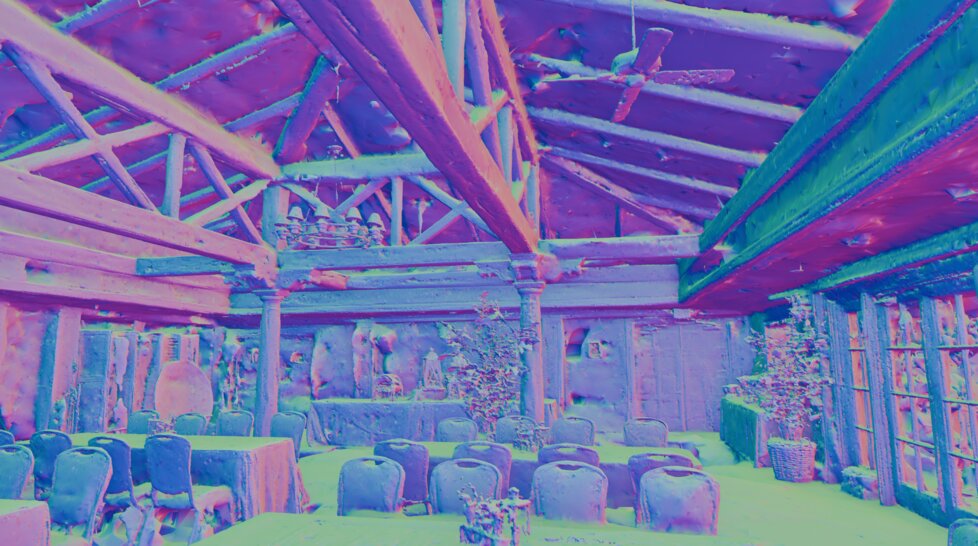} & 
        \includegraphics[width=0.15\textwidth, trim={9cm 5cm 6cm 1.5cm},clip]{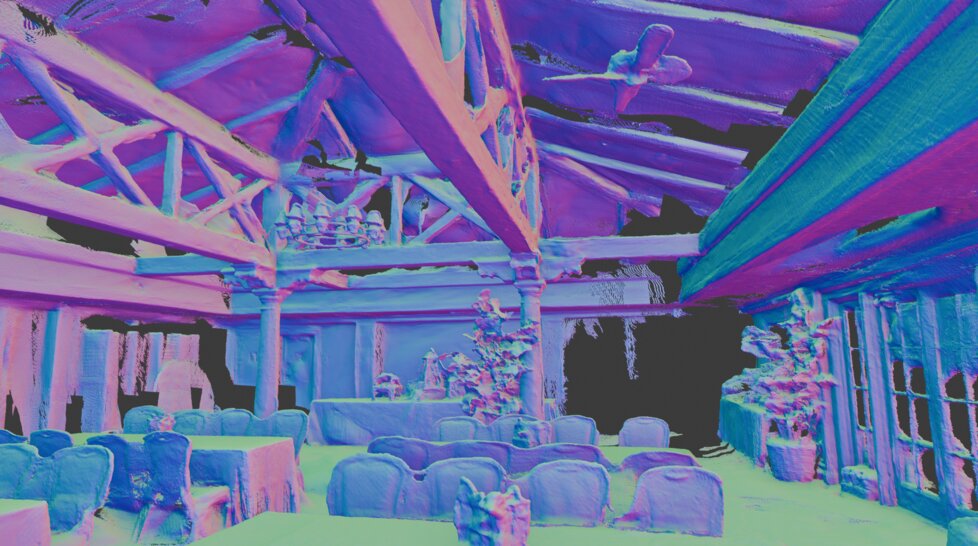} & 
        \includegraphics[width=0.15\textwidth, trim={9cm 5cm 6cm 1.5cm},clip]{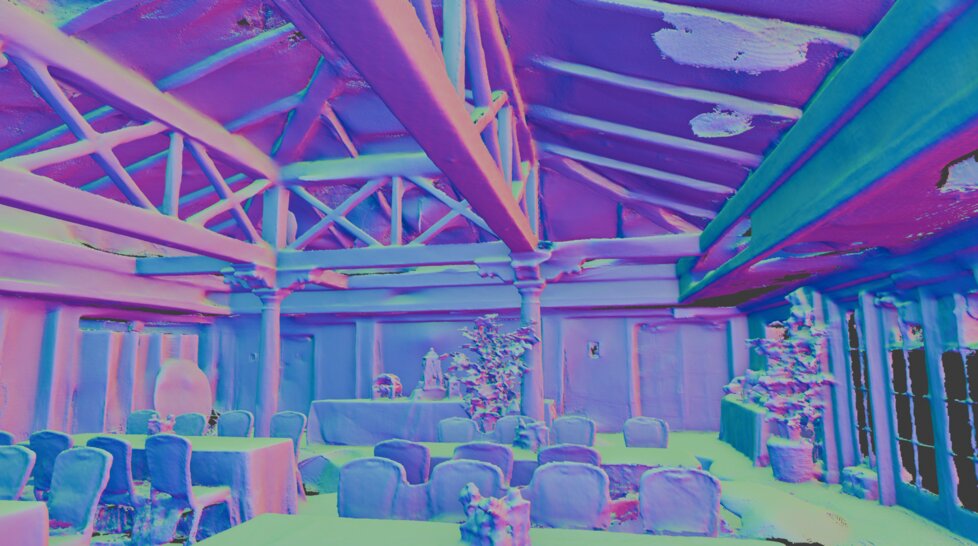} \\
        \includegraphics[width=0.15\textwidth, trim={9cm 7.5cm 14.5cm 5cm},clip]{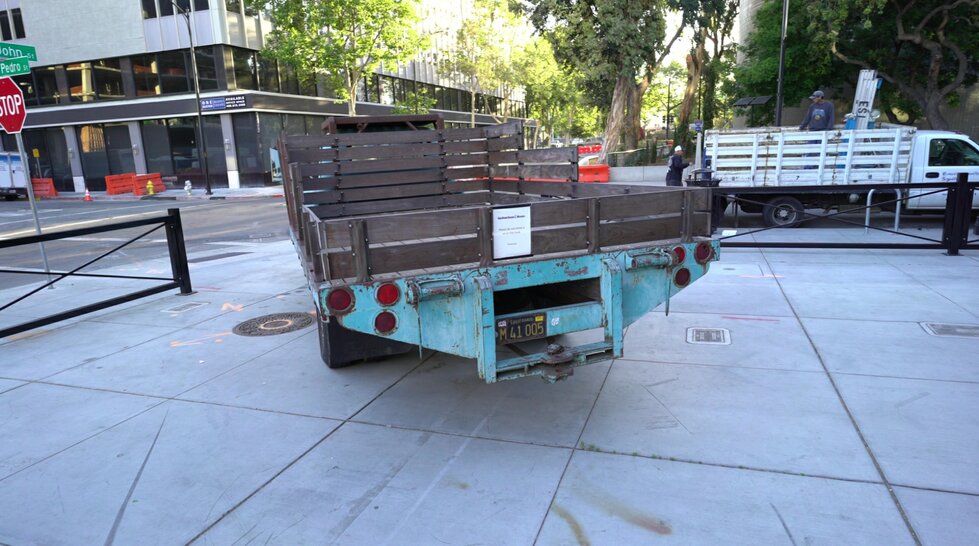} & 
        \includegraphics[width=0.15\textwidth, trim={9cm 7.5cm 14.5cm 5cm},clip]{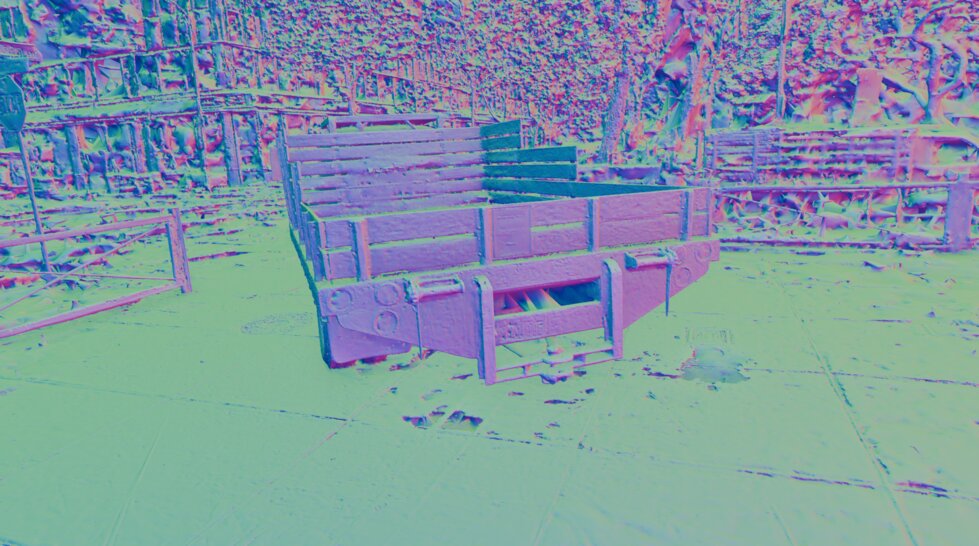} & 
        \includegraphics[width=0.15\textwidth, trim={9cm 7.5cm 14.5cm 5cm},clip]{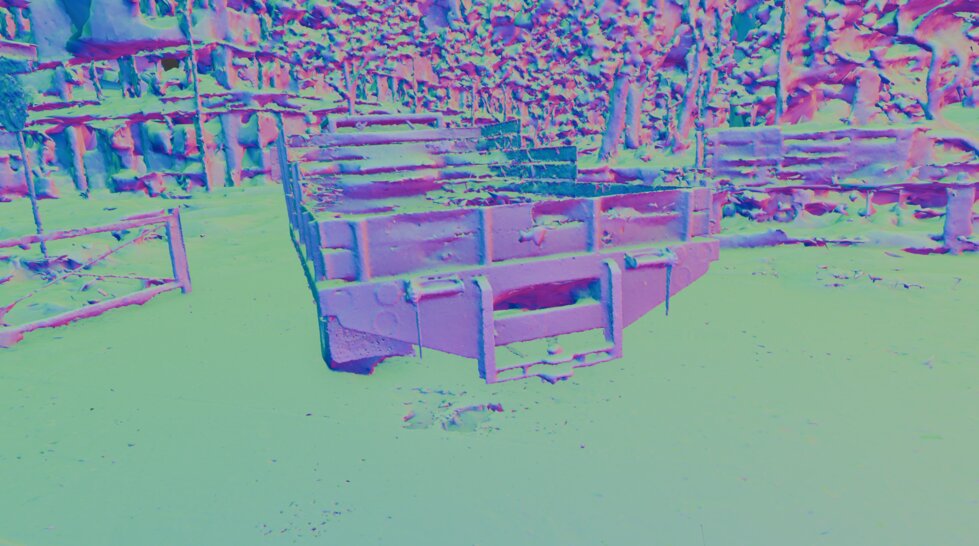} & 
        \includegraphics[width=0.15\textwidth, trim={9cm 7.5cm 14.5cm 5cm},clip]{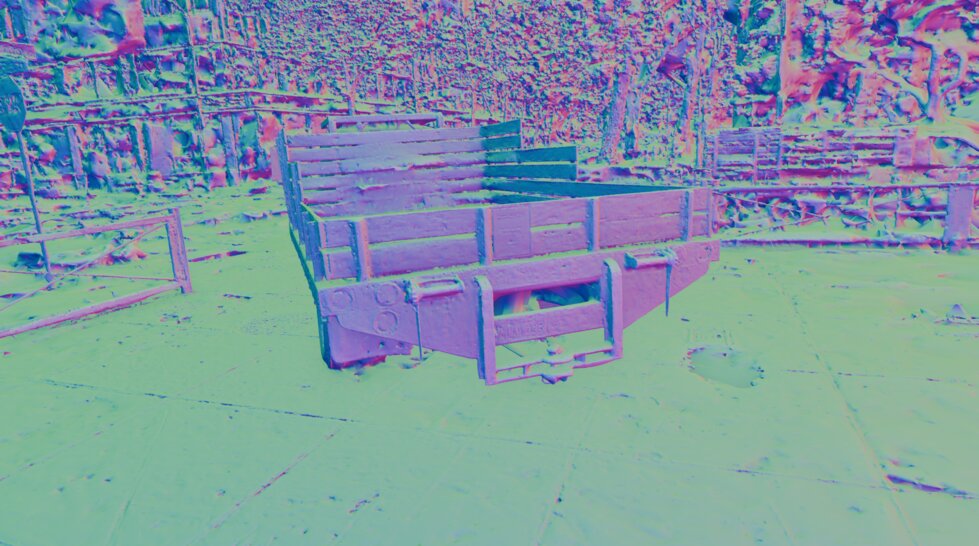} & 
        \includegraphics[width=0.15\textwidth, trim={9cm 7.5cm 14.5cm 5cm},clip]{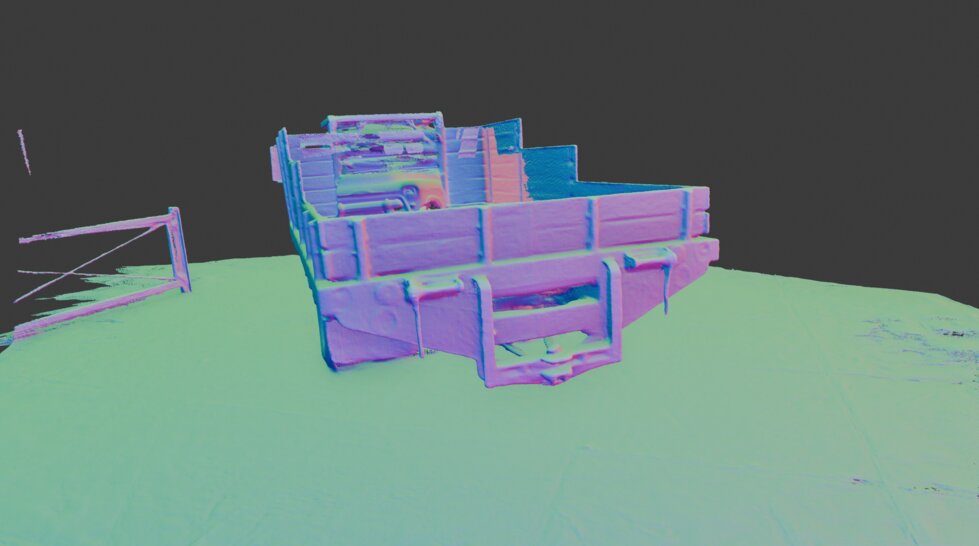} & 
        \includegraphics[width=0.15\textwidth, trim={9cm 7.5cm 14.5cm 5cm},clip]{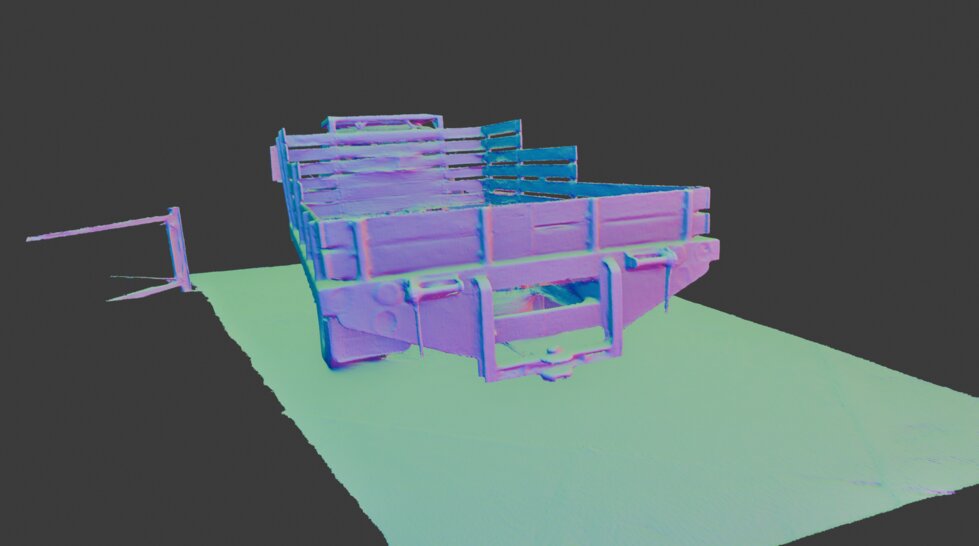} \\
        \includegraphics[width=0.15\textwidth, trim={0cm 4cm 15cm 6.5cm},clip]{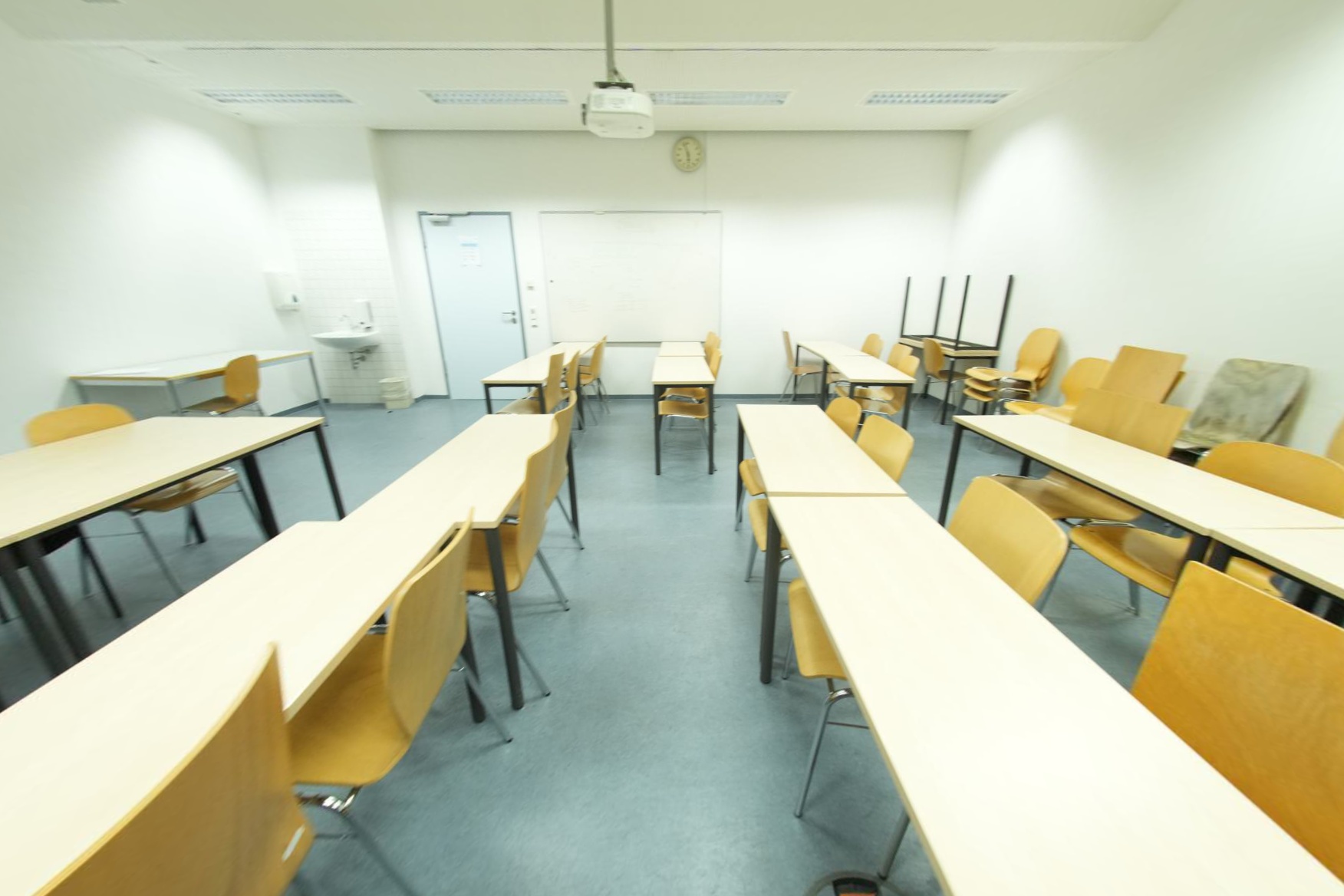} &
        \includegraphics[width=0.15\textwidth, trim={0cm 4cm 15cm 6.5cm},clip]{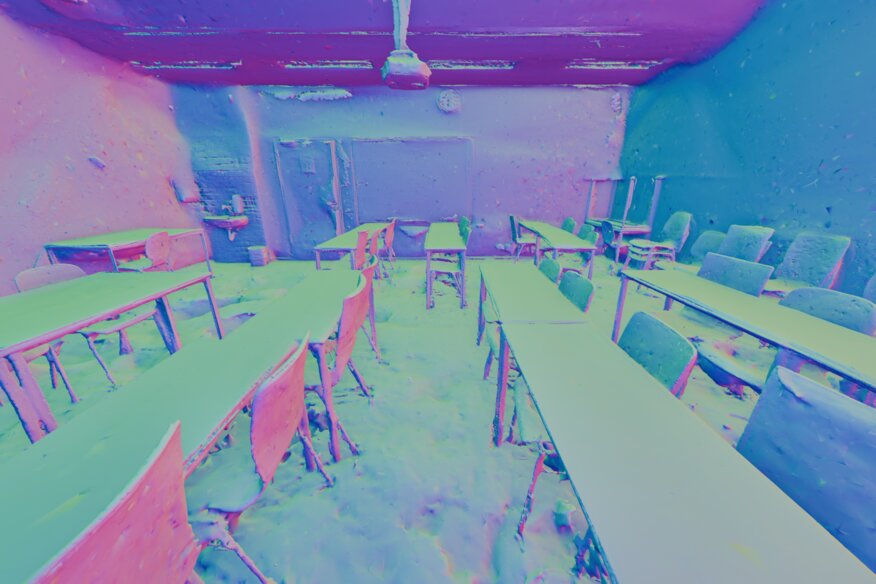} & 
        \includegraphics[width=0.15\textwidth, trim={0cm 4cm 15cm 6.5cm},clip]{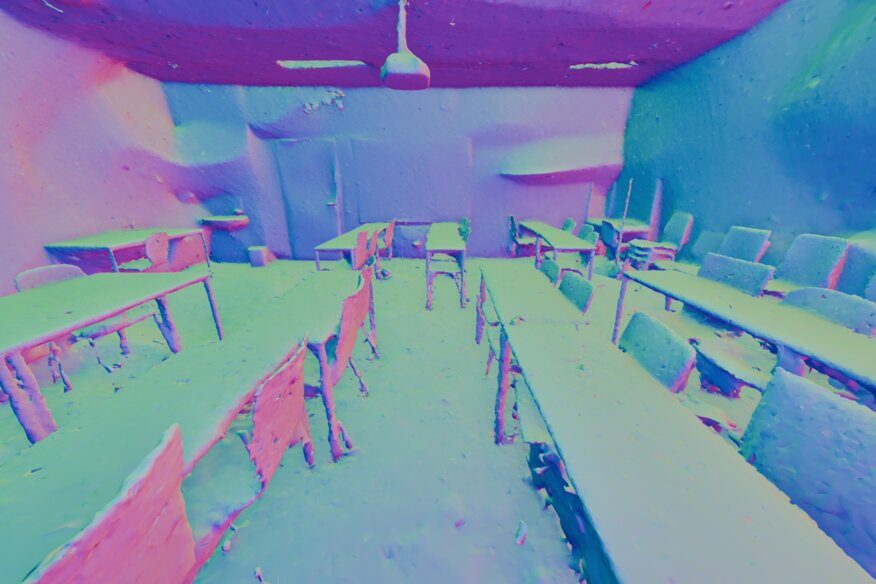} & 
        \includegraphics[width=0.15\textwidth, trim={0cm 4cm 15cm 6.5cm},clip]{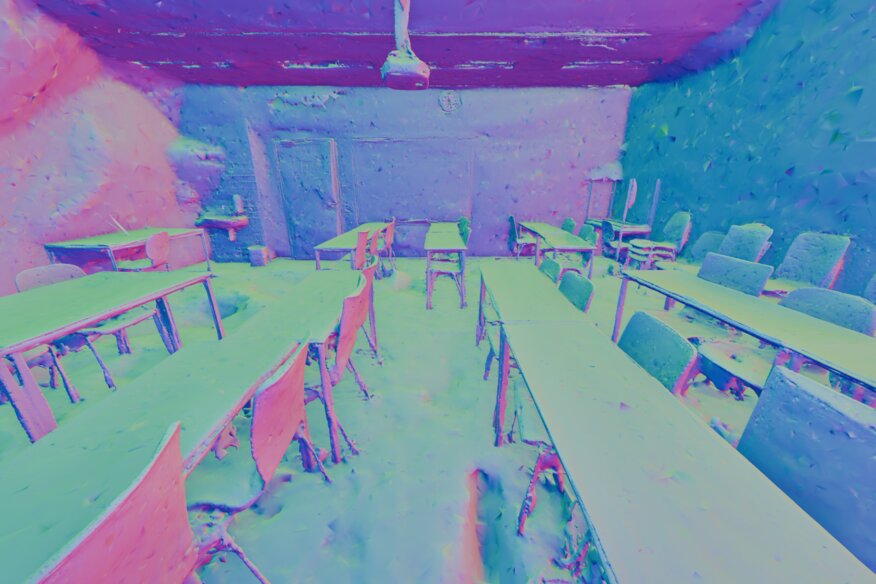} & 
        \includegraphics[width=0.15\textwidth, trim={0cm 4cm 15cm 6.5cm},clip]{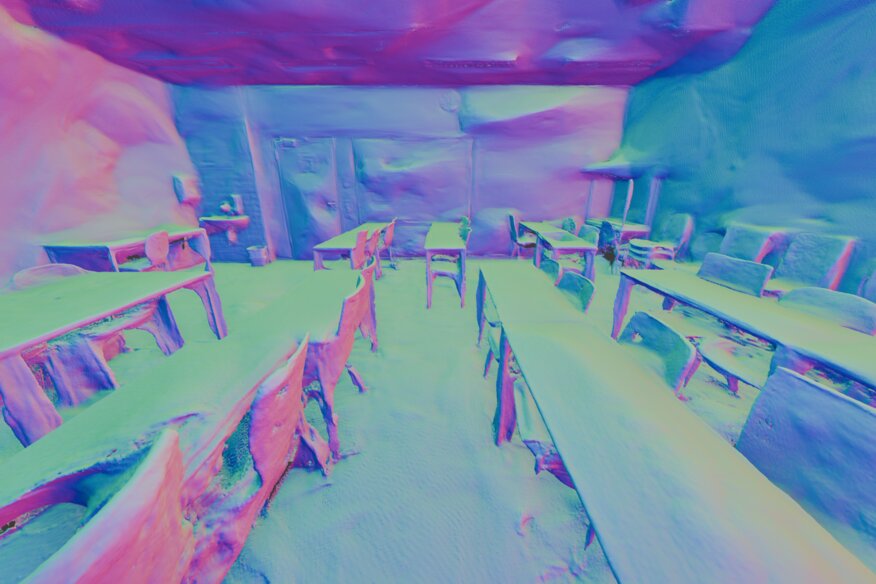} & 
        \includegraphics[width=0.15\textwidth, trim={0cm 4cm 15cm 6.5cm},clip]{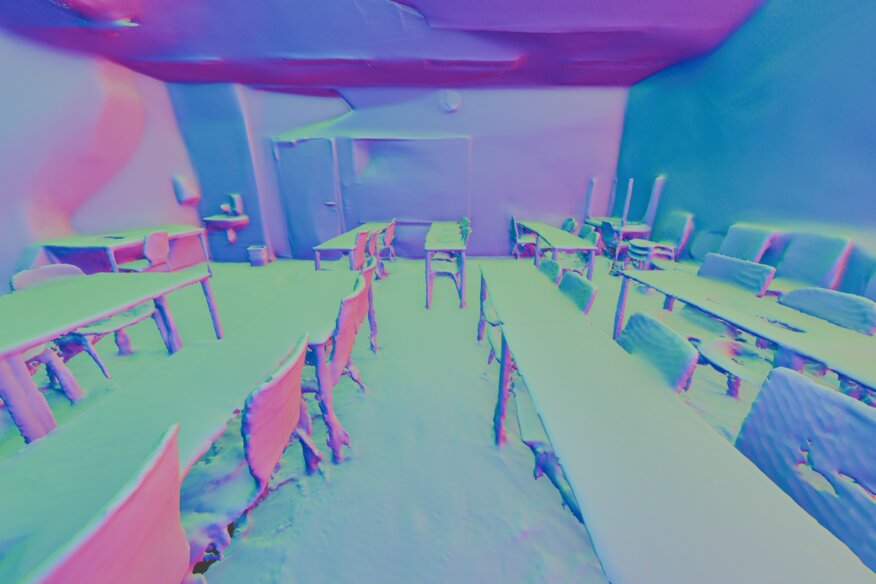} \\
        \includegraphics[width=0.15\textwidth, trim={0cm 1.5cm 9cm 2.5cm},clip]{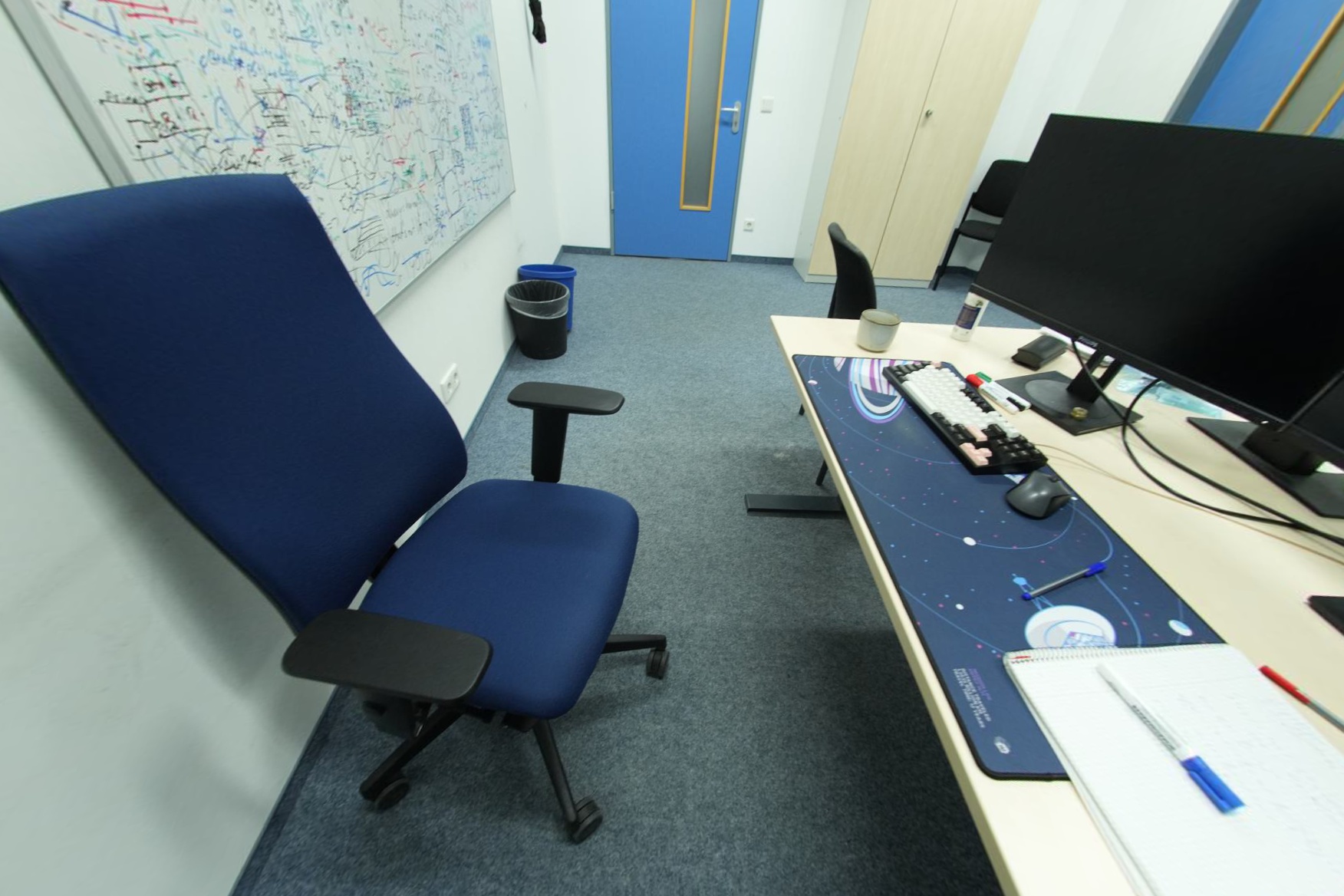} &
        \includegraphics[width=0.15\textwidth, trim={0cm 1.5cm 9cm 2.5cm},clip]{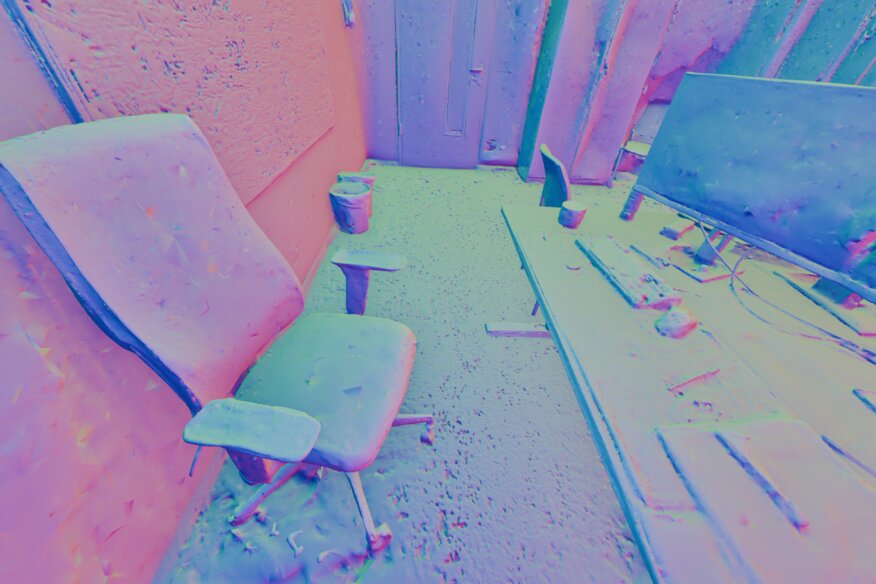} & 
        \includegraphics[width=0.15\textwidth, trim={0cm 1.5cm 9cm 2.5cm},clip]{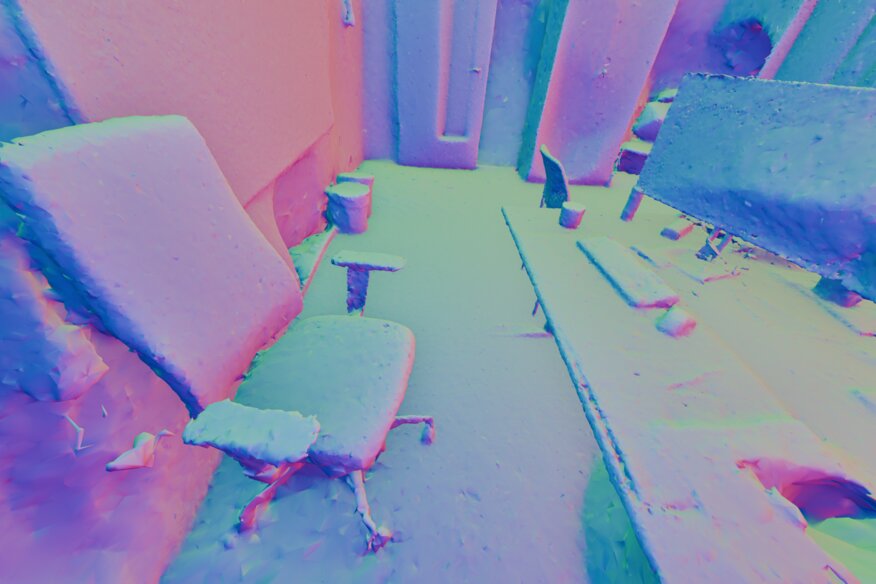} & 
        \includegraphics[width=0.15\textwidth, trim={0cm 1.5cm 9cm 2.5cm},clip]{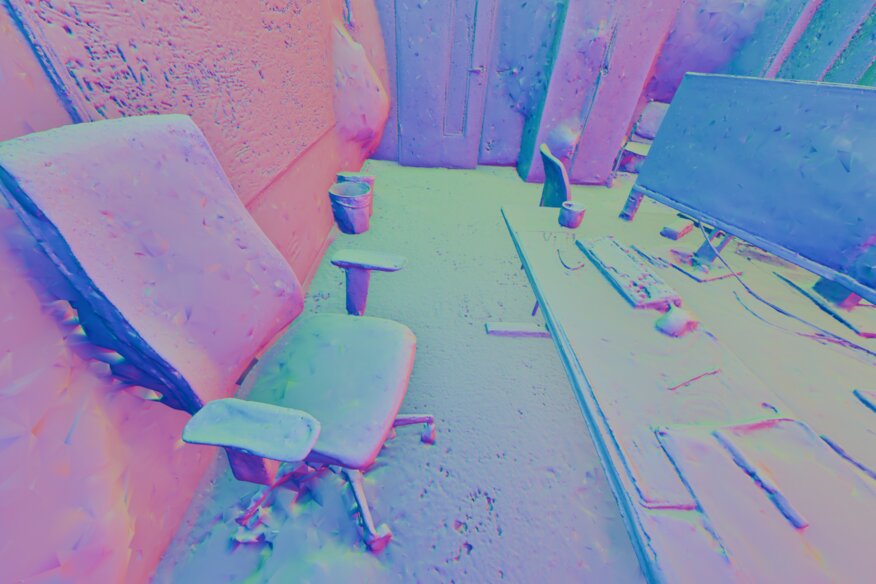} & 
        \includegraphics[width=0.15\textwidth, trim={0cm 1.5cm 9cm 2.5cm},clip]{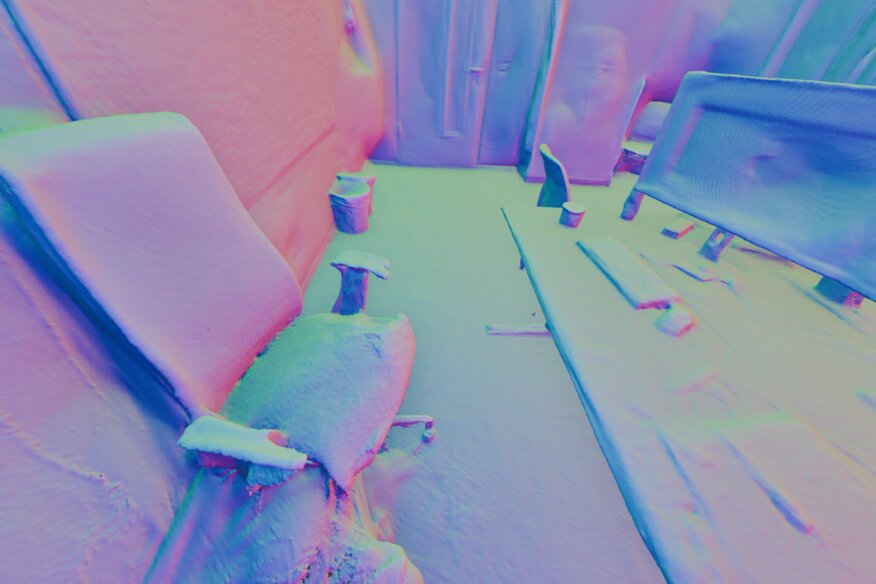} & 
        \includegraphics[width=0.15\textwidth, trim={0cm 1.5cm 9cm 2.5cm},clip]{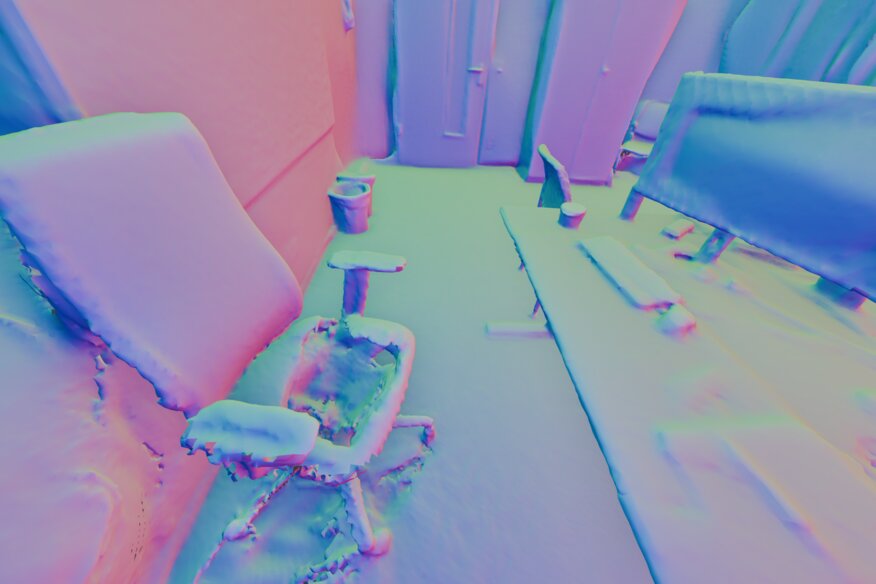} \\
        \includegraphics[width=0.15\textwidth, trim={7cm 21cm 13cm 0cm},clip]{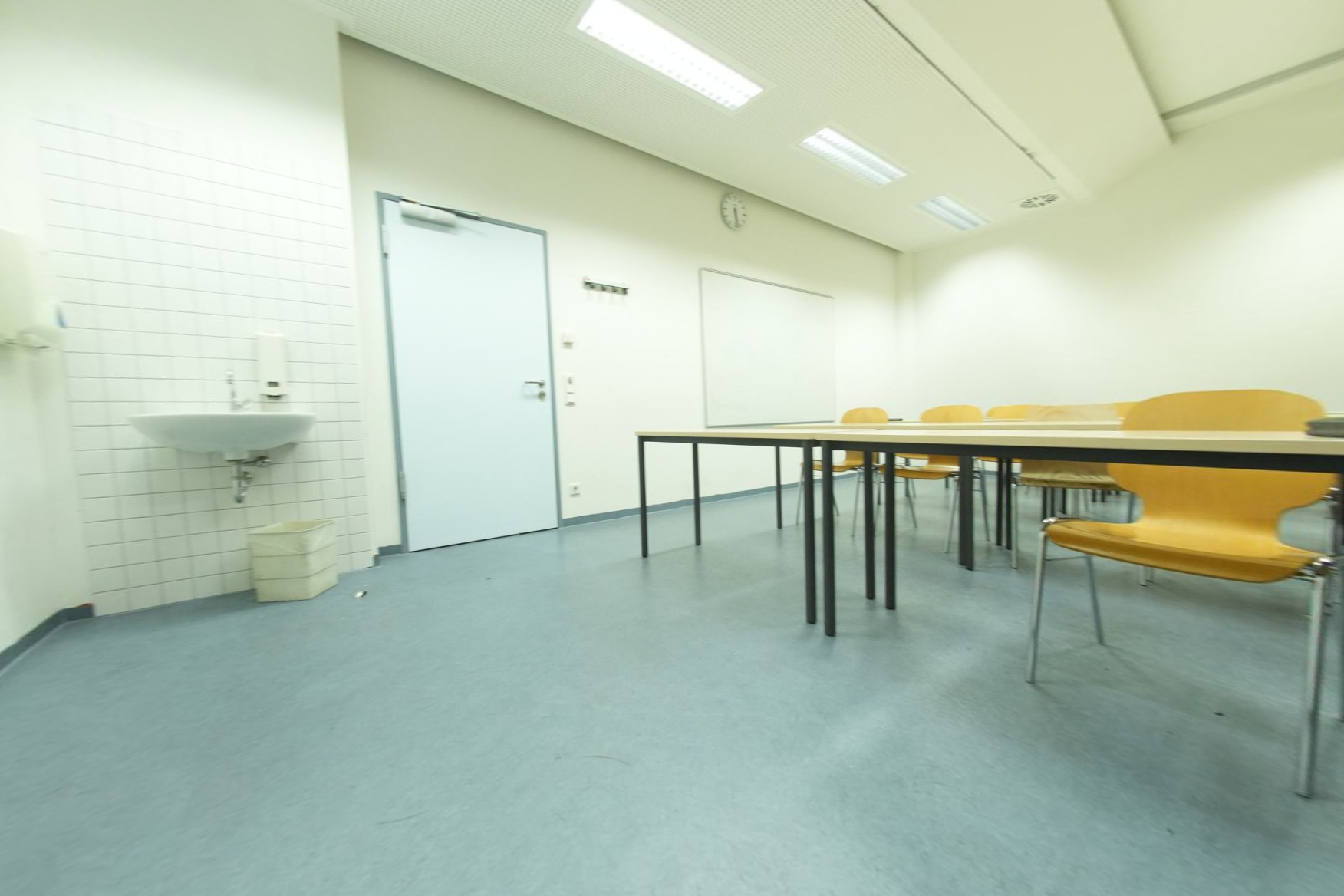} & 
        \includegraphics[width=0.15\textwidth, trim={3.5cm 10.5cm 6.5cm 0cm},clip]{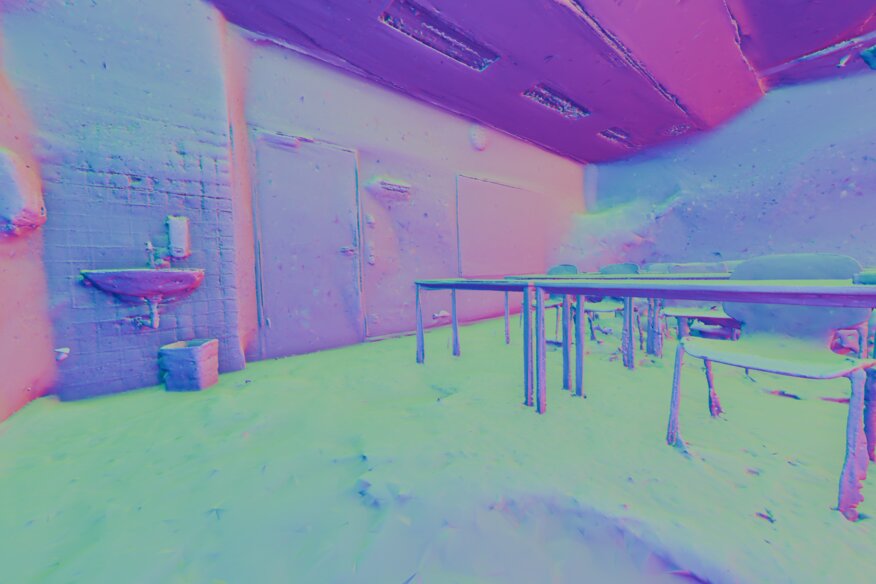} & 
        \includegraphics[width=0.15\textwidth, trim={3.5cm 10.5cm 6.5cm 0cm},clip]{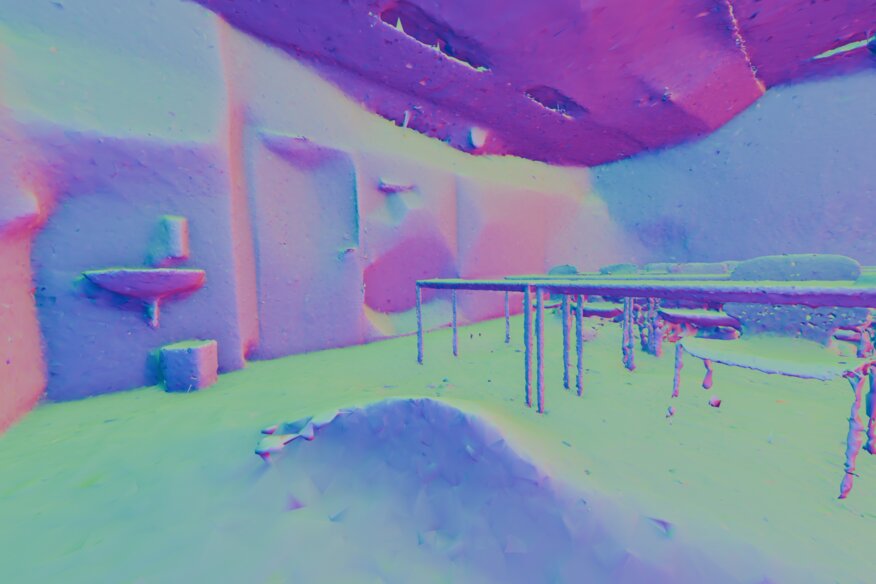} & 
        \includegraphics[width=0.15\textwidth, trim={3.5cm 10.5cm 6.5cm 0cm},clip]{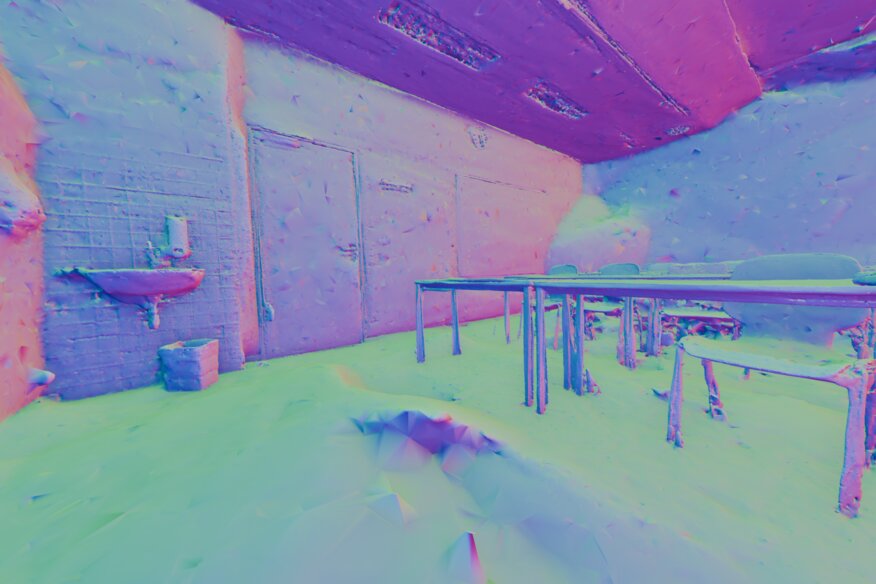} & 
        \includegraphics[width=0.15\textwidth, trim={3.5cm 10.5cm 6.5cm 0cm},clip]{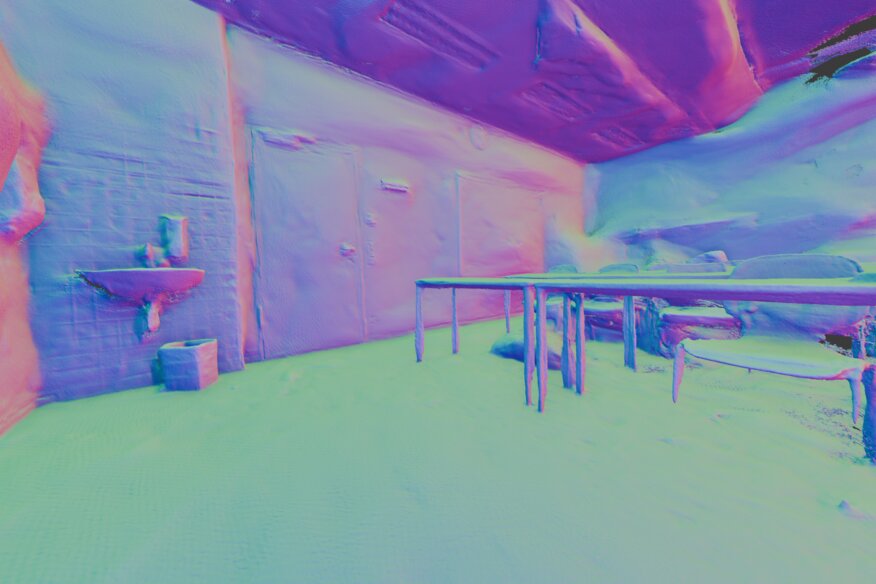} & 
        \includegraphics[width=0.15\textwidth, trim={3.5cm 10.5cm 6.5cm 0cm},clip]{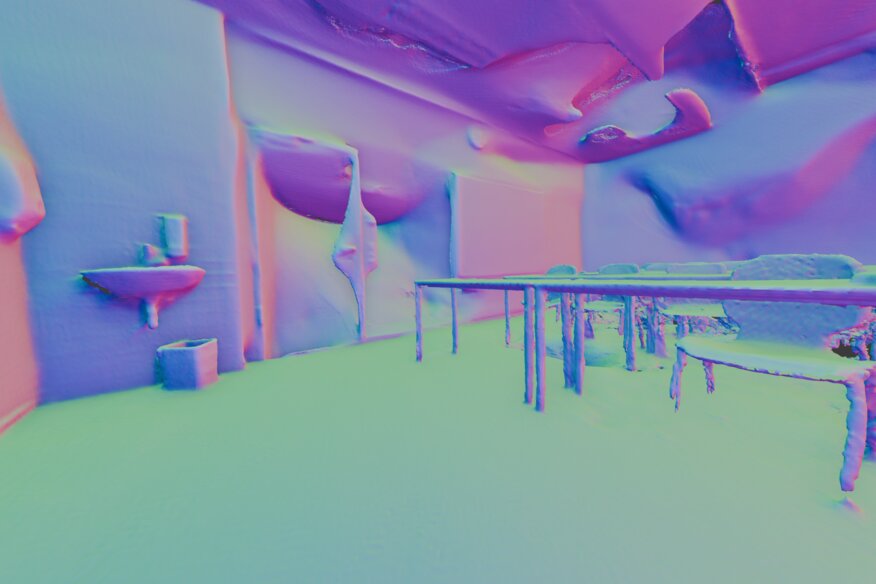} \\
        \includegraphics[width=0.15\textwidth, trim={12cm 0cm 3cm 9cm},clip]{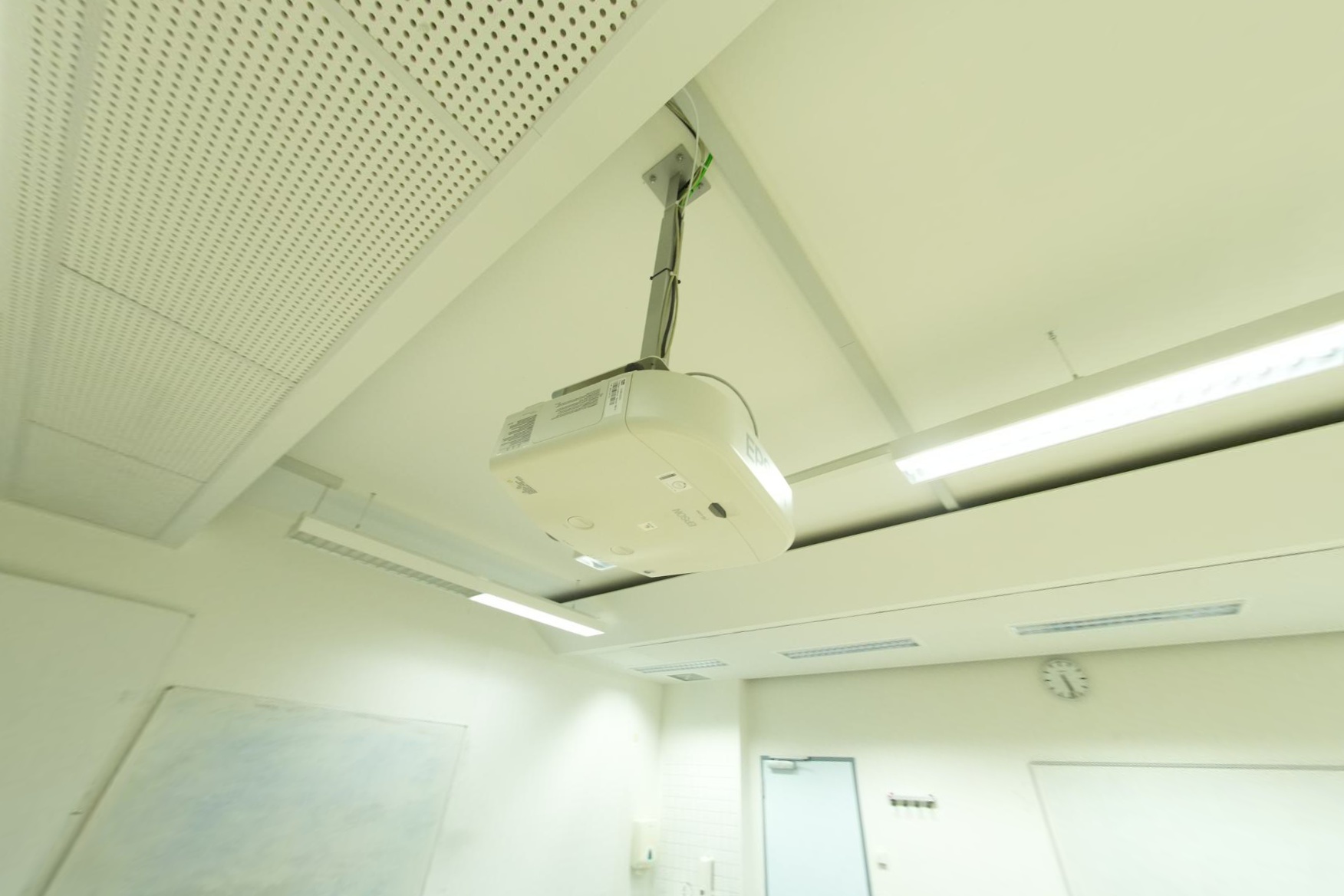} &
        \includegraphics[width=0.15\textwidth, trim={6cm 0cm 1.5cm 4.5cm},clip]{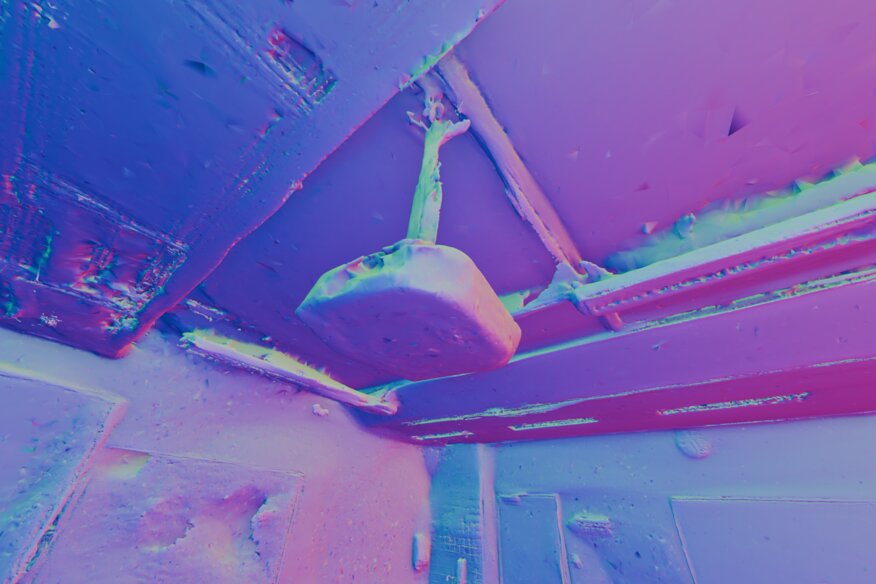} & 
        \includegraphics[width=0.15\textwidth, trim={6cm 0cm 1.5cm 4.5cm},clip]{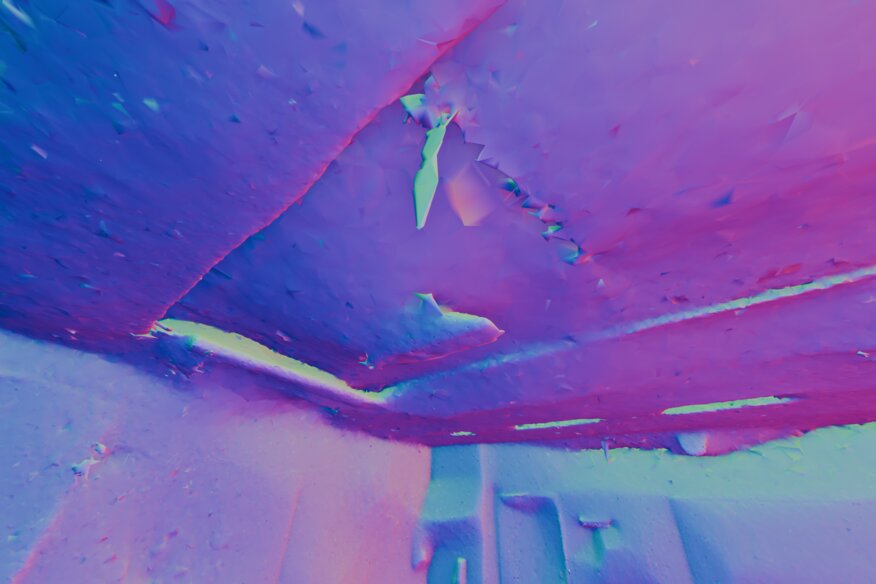} & 
        \includegraphics[width=0.15\textwidth, trim={6cm 0cm 1.5cm 4.5cm},clip]{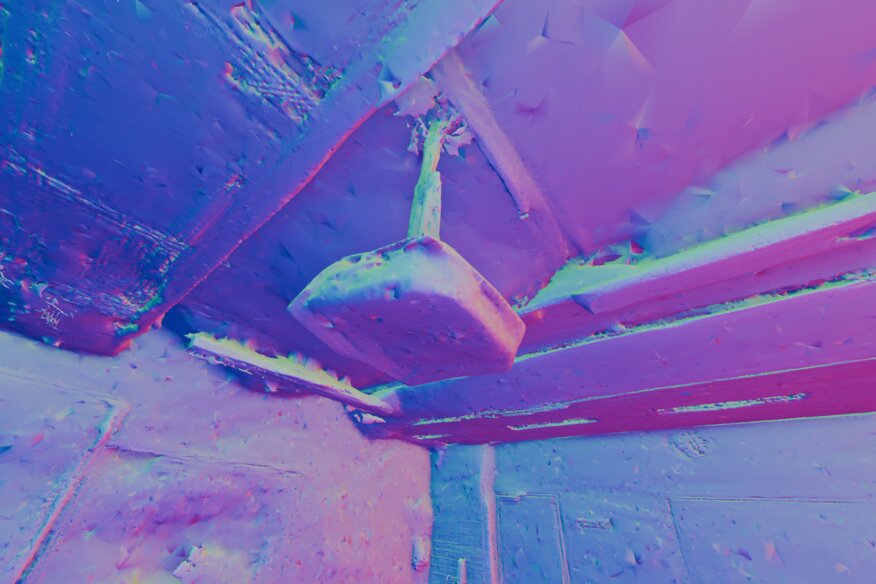} & 
        \includegraphics[width=0.15\textwidth, trim={6cm 0cm 1.5cm 4.5cm},clip]{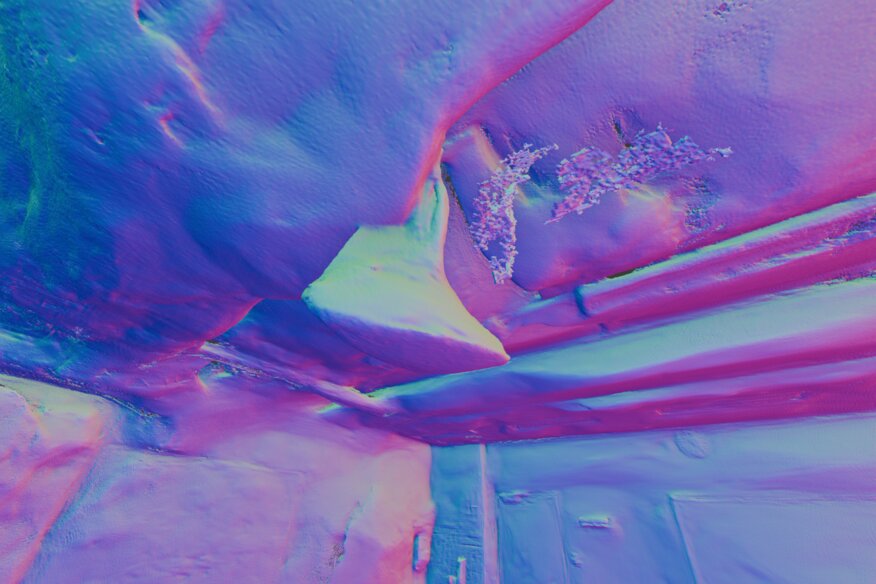} & 
        \includegraphics[width=0.15\textwidth, trim={6cm 0cm 1.5cm 4.5cm},clip]{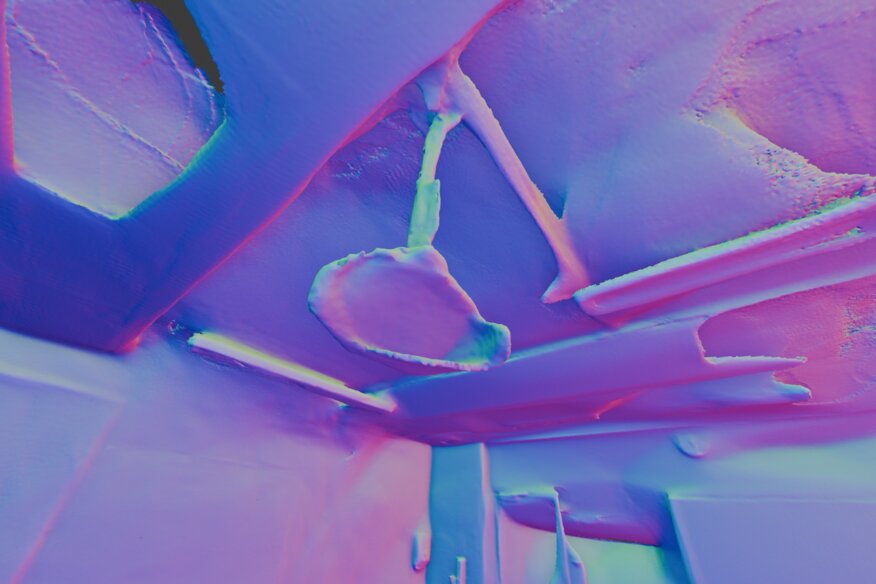} \\
        \includegraphics[width=0.15\textwidth, trim={5cm 12cm 0cm 0cm},clip]{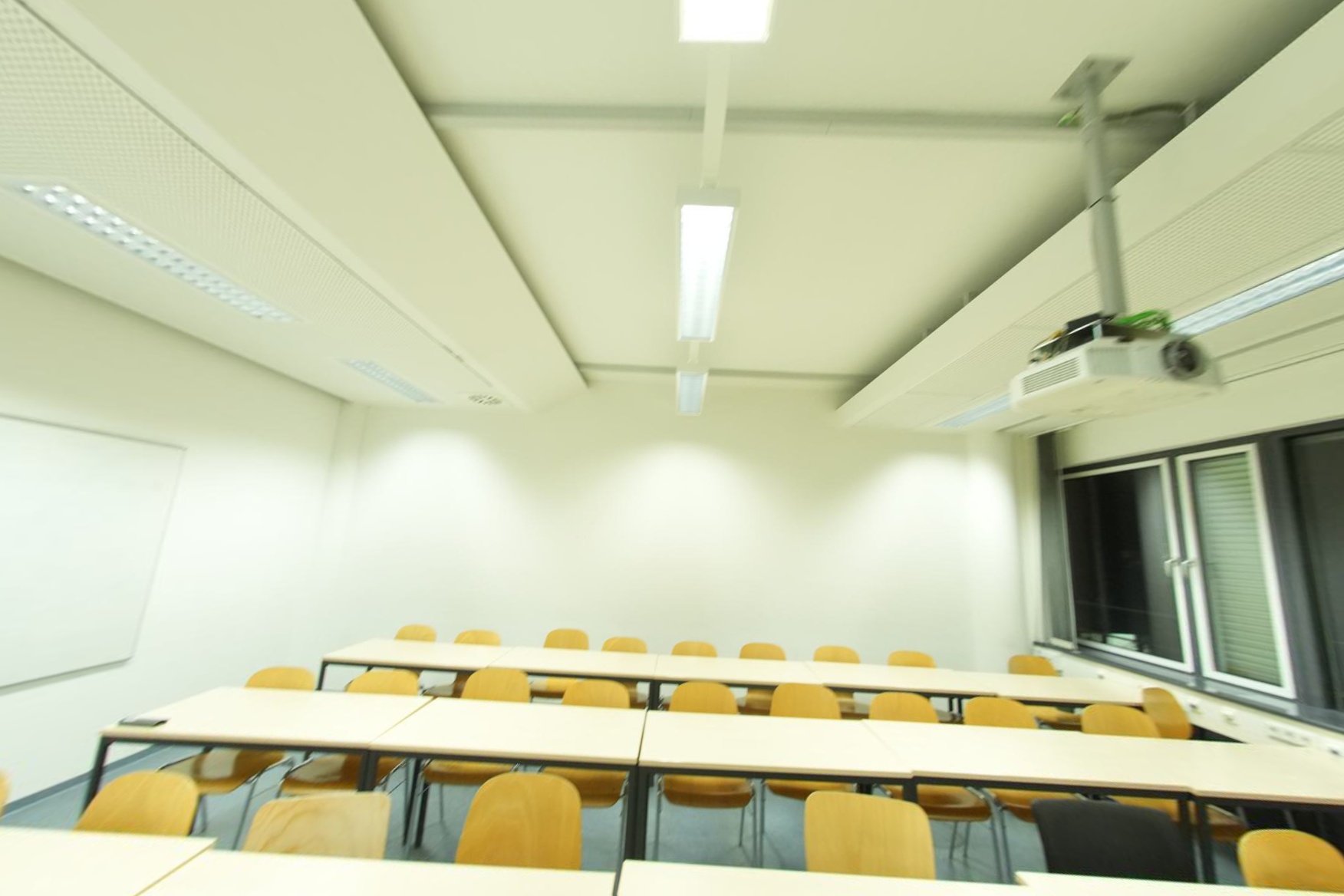} &
        \includegraphics[width=0.15\textwidth, trim={2.5cm 6cm 0cm 0cm},clip]{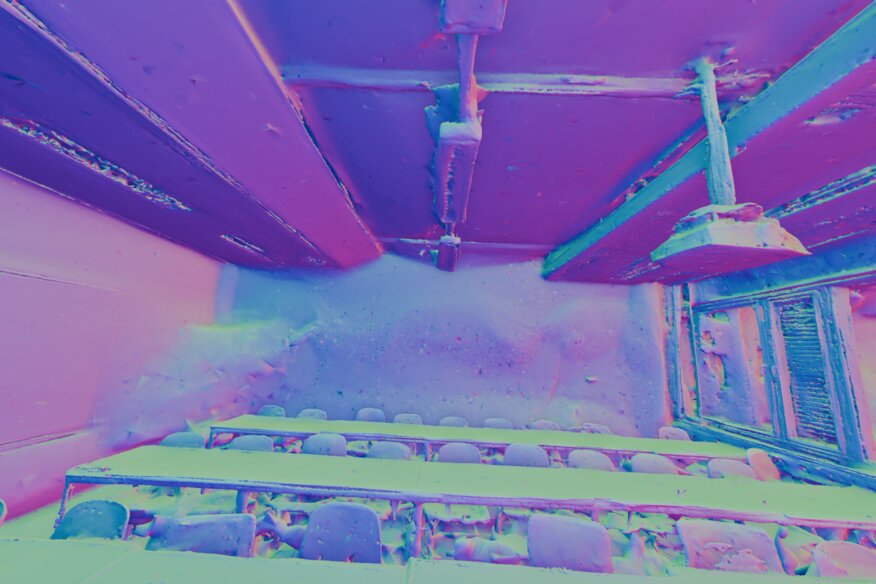} & 
        \includegraphics[width=0.15\textwidth, trim={2.5cm 6cm 0cm 0cm},clip]{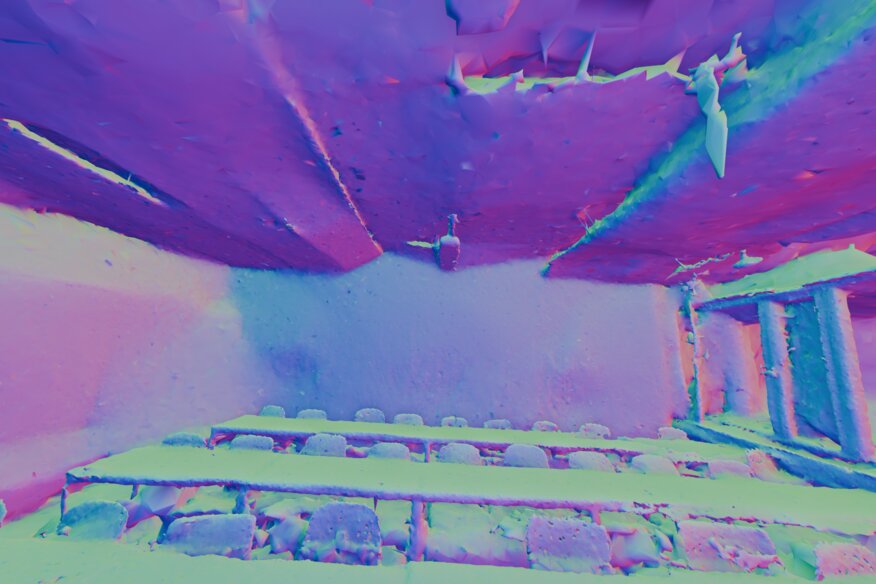} & 
        \includegraphics[width=0.15\textwidth, trim={2.5cm 6cm 0cm 0cm},clip]{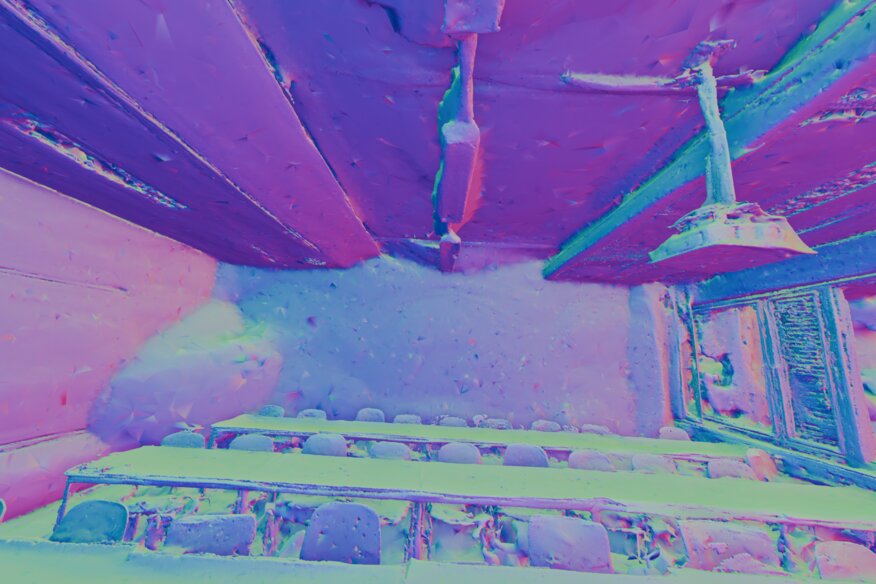} & 
        \includegraphics[width=0.15\textwidth, trim={2.5cm 6cm 0cm 0cm},clip]{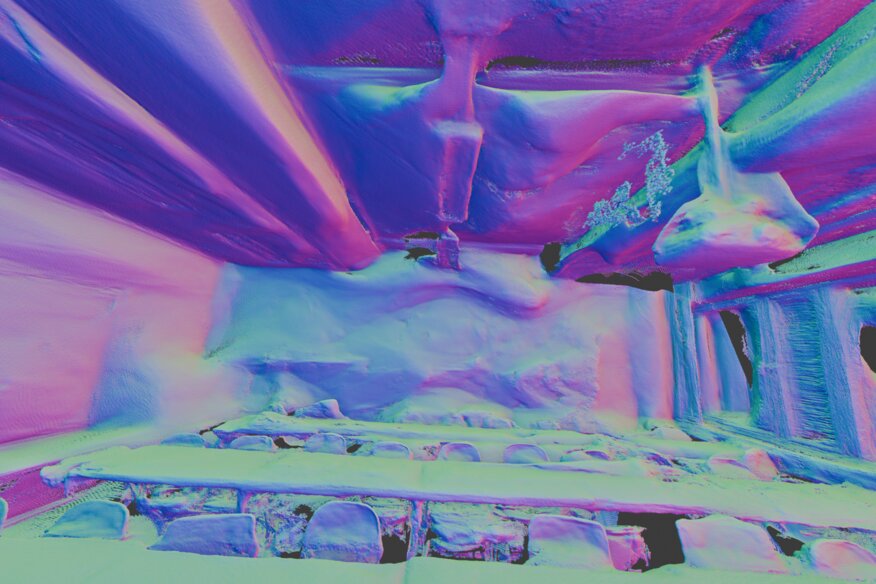} & 
        \includegraphics[width=0.15\textwidth, trim={2.5cm 6cm 0cm 0cm},clip]{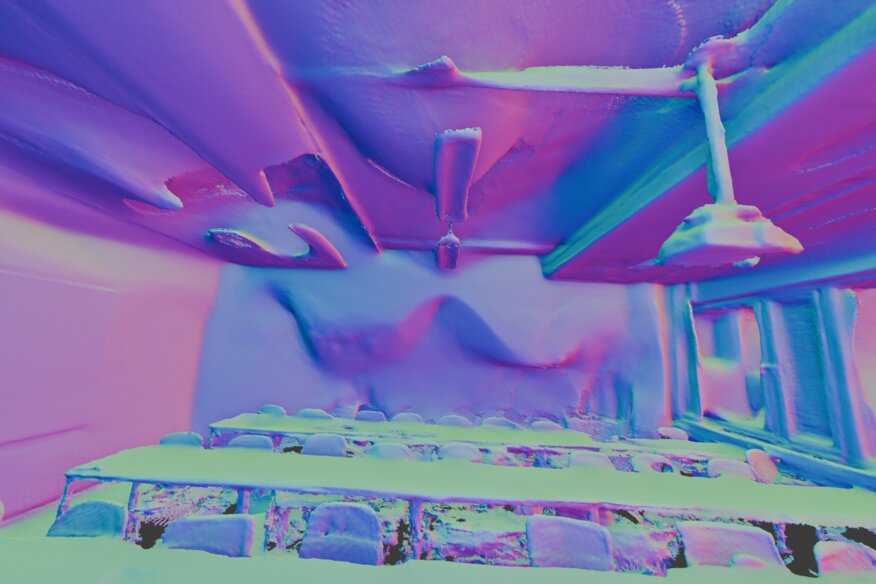} \\
        \includegraphics[width=0.15\textwidth, trim={0cm 6cm 16cm 12cm},clip]{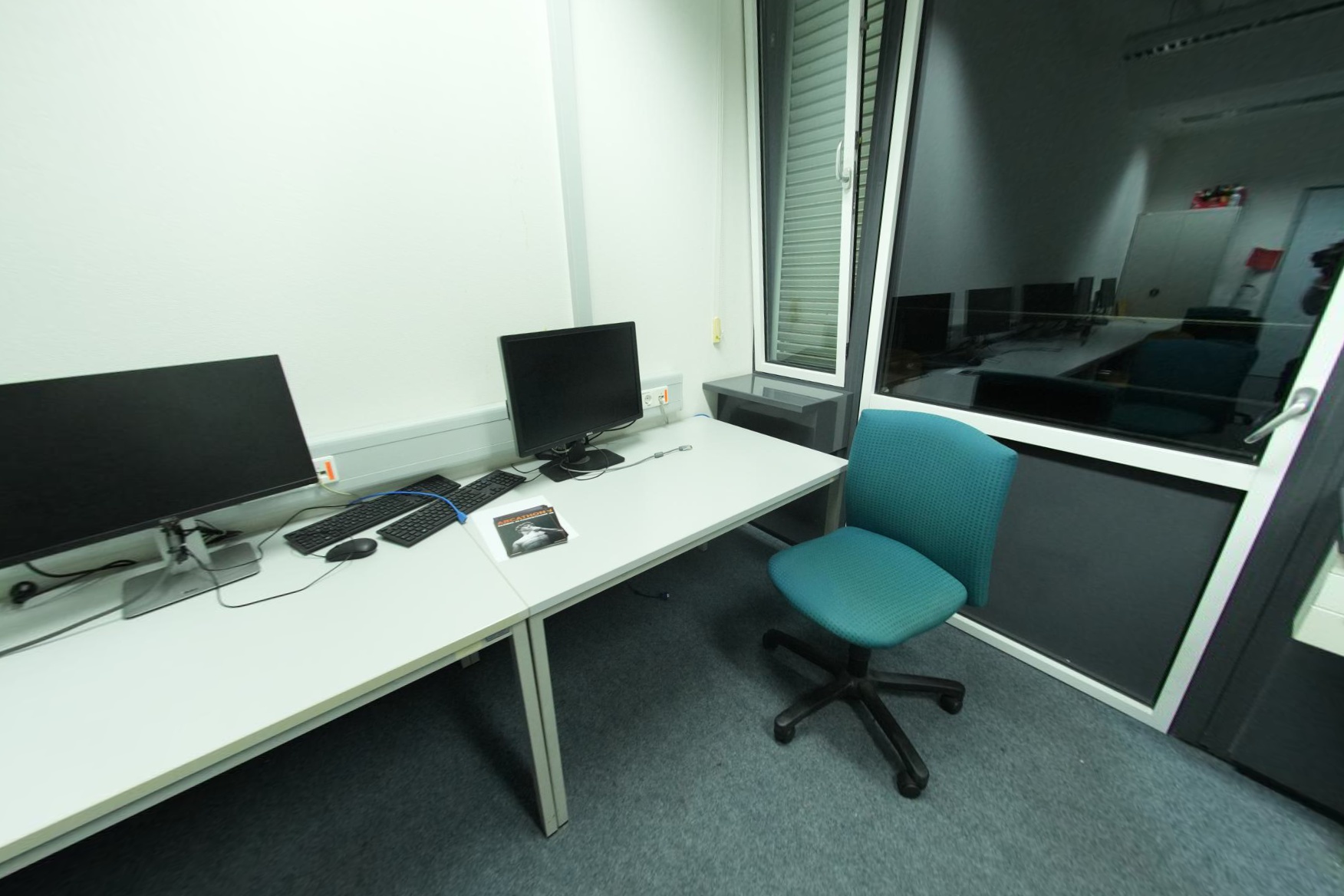} & 
        \includegraphics[width=0.15\textwidth, trim={0cm 3cm 8cm 6cm},clip]{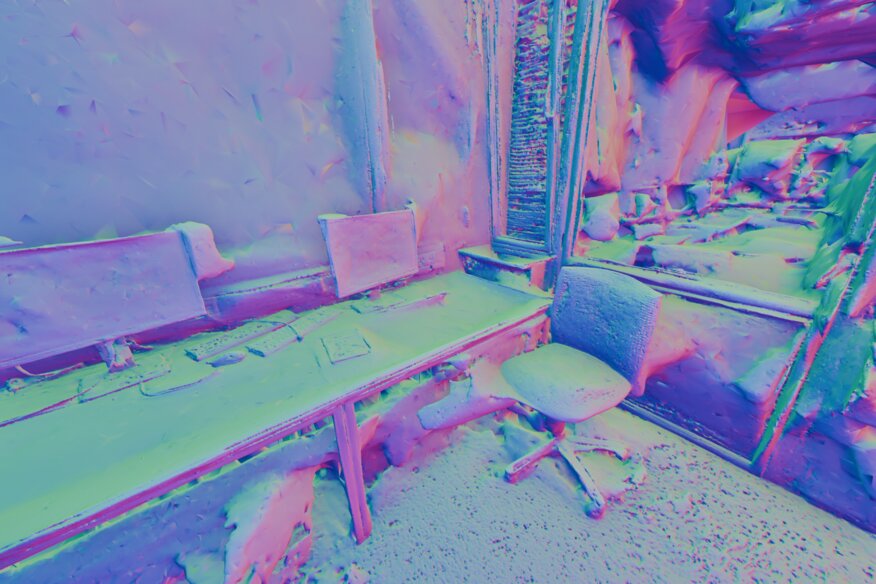} & 
        \includegraphics[width=0.15\textwidth, trim={0cm 3cm 8cm 6cm},clip]{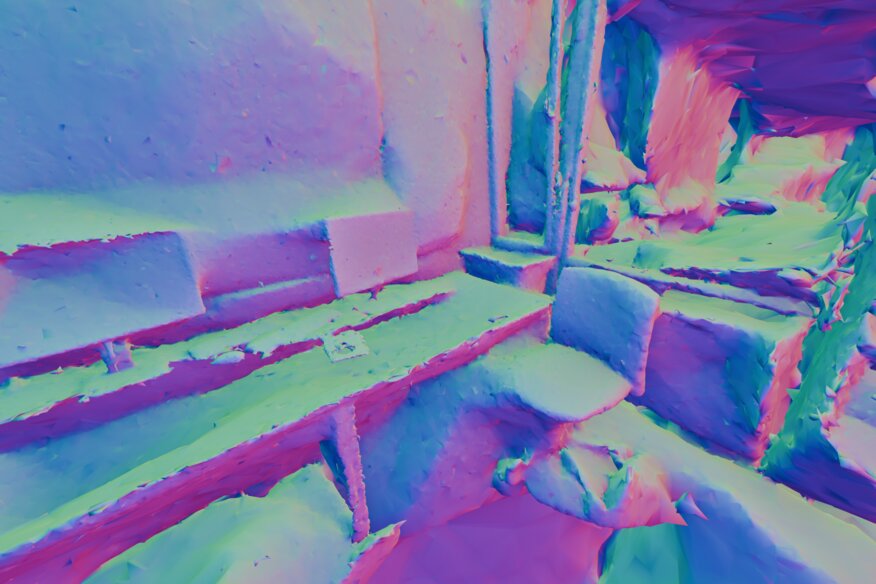} & 
        \includegraphics[width=0.15\textwidth, trim={0cm 3cm 8cm 6cm},clip]{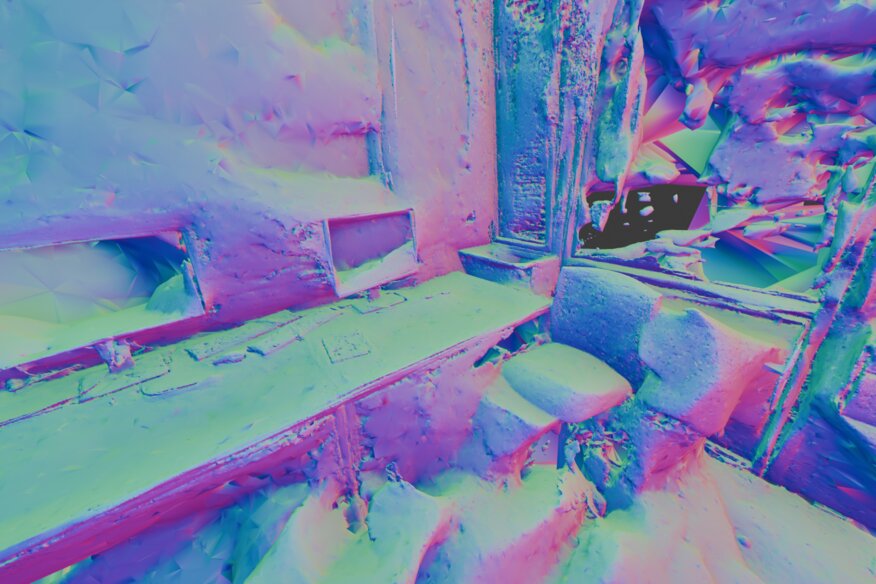} & 
        \includegraphics[width=0.15\textwidth, trim={0cm 3cm 8cm 6cm},clip]{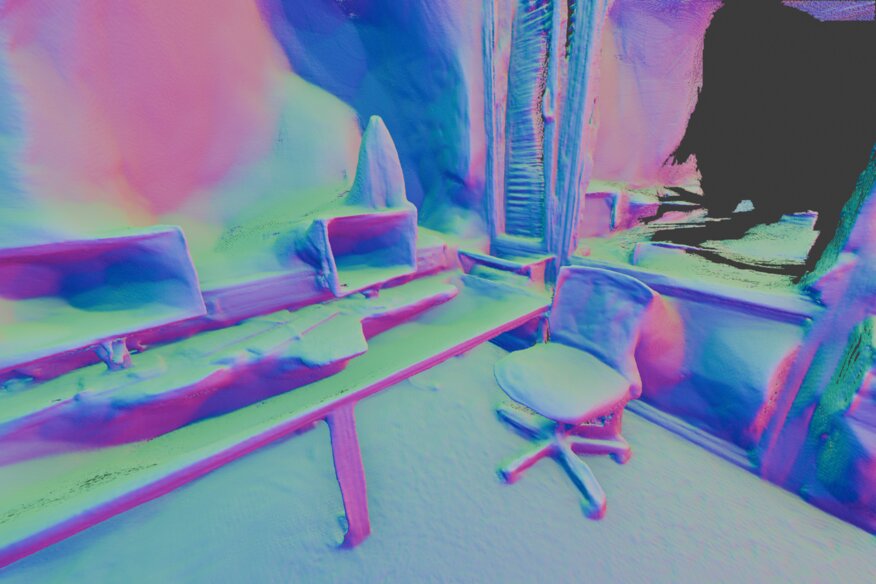} & 
        \includegraphics[width=0.15\textwidth, trim={0cm 3cm 8cm 6cm},clip]{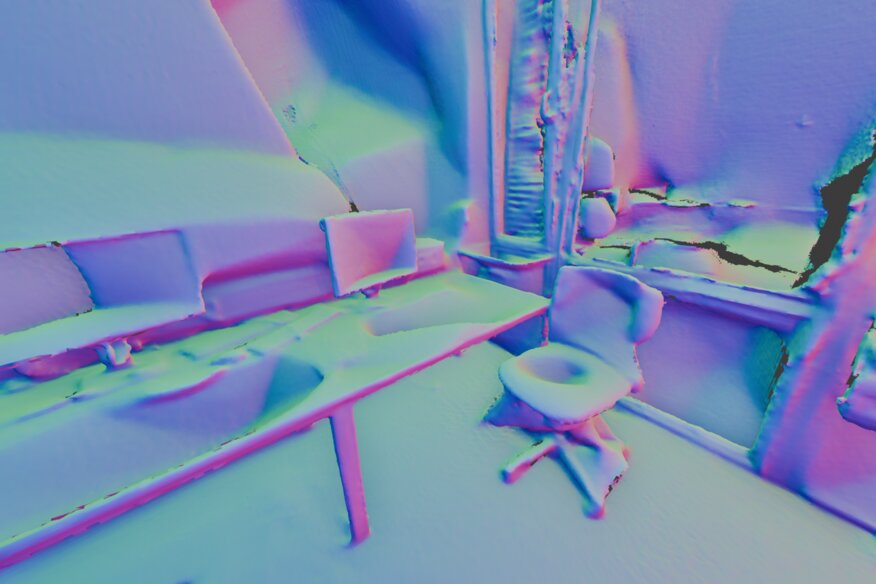} \\
    \end{tabular}
    
    \caption{
\textbf{Qualitative Mesh Comparison}:
Our approach produces more detailed meshes with fewer artifacts than other unbounded methods~\cite{guedon2025milo, Radl2025SOF}. 
Additionally, it achieves higher completeness and finer detail than bounded extraction works~\cite{chen2024pgsr, zhang2025qgs}.
}
    \label{fig:qualitative}
\end{figure}

\para{Qualitative Comparison}
\Cref{fig:qualitative} visualizes qualitative mesh comparisons against baseline approaches. 
When compared to bounded surface extraction methods~\cite{chen2024pgsr, zhang2025qgs}, our meshes are significantly more complete. 
These baselines rely on multi-view constraints that tend to aggressively over-smooth or entirely cull fine, complex geometry (\eg, thin structures and foliage).

\begin{figure}[h!]
    \centering
    
    \def\xminA{0.55} \def\yminA{0.3}
    \def\xmaxA{0.7} \def\ymaxA{0.45}
    
    \def\xminB{0.5} \def\yminB{0.7}
    \def\xmaxB{0.65} \def\ymaxB{0.85}

    \begin{subfigure}[b]{0.153\textwidth}
        \centering
        \adjincludegraphics[viewport={\xminA\width} {\yminA\height} {\xmaxA\width} {\ymaxA\height}, clip, width=\textwidth]{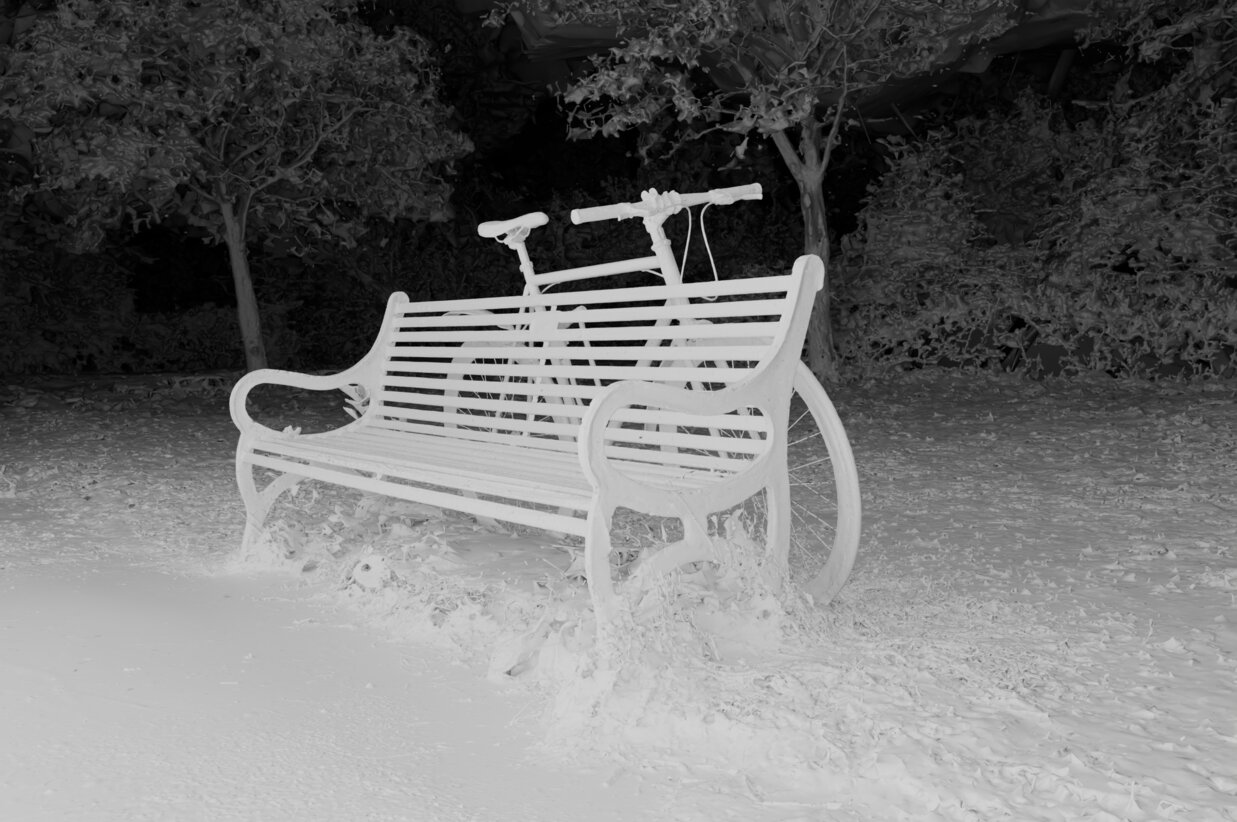}
        
        \vspace{2pt} 
        
        \adjincludegraphics[viewport={\xminA\width} {\yminA\height} {\xmaxA\width} {\ymaxA\height}, clip, width=\textwidth]{images/renders/Bicycle_GT.JPG}
    \end{subfigure}
    \hfill
    \begin{subfigure}[b]{0.32\textwidth}
        \centering
        \begin{tikzpicture}
            \node[anchor=south west, inner sep=0] (img1) at (0,0) {\includegraphics[width=\textwidth]{images/renders/Bicycle_Final.jpg}};
            
            \begin{scope}[x={(img1.south east)}, y={(img1.north west)}]
                \draw[tab_color, thin] (\xminA, \yminA) rectangle (\xmaxA, \ymaxA);
            \end{scope}
        \end{tikzpicture}
    \end{subfigure}%
    \hfill
    \begin{subfigure}[b]{0.32\textwidth}
        \centering
        \begin{tikzpicture}
            \node[anchor=south west, inner sep=0] (img2) at (0,0) {\includegraphics[width=\textwidth]{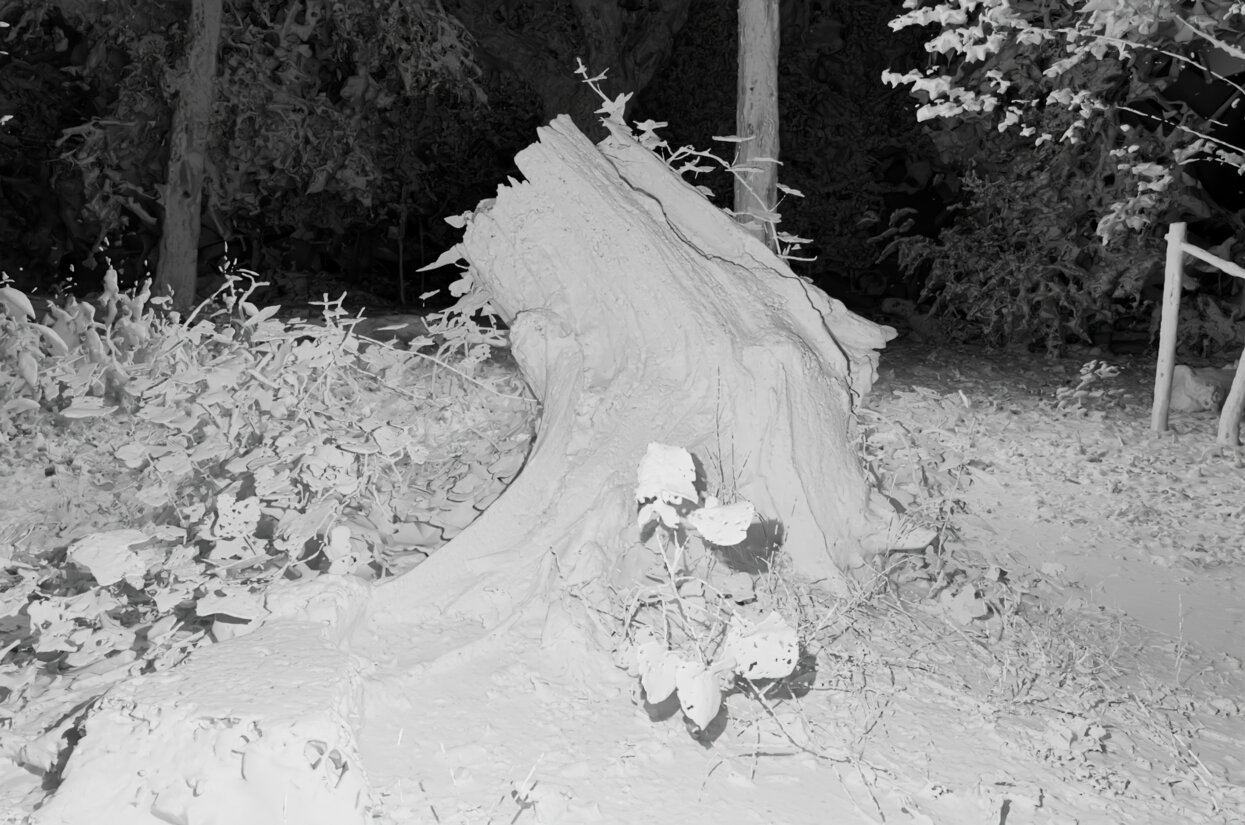}};
            
            \begin{scope}[x={(img2.south east)}, y={(img2.north west)}]
                \draw[tab_color, thin] (\xminB, \yminB) rectangle (\xmaxB, \ymaxB);
            \end{scope}
        \end{tikzpicture}
    \end{subfigure}%
    \hfill
    \begin{subfigure}[b]{0.153\textwidth}
        \centering
        \adjincludegraphics[viewport={\xminB\width} {\yminB\height} {\xmaxB\width} {\ymaxB\height}, clip, width=\textwidth]{images/renders/Stump_Final.jpg}
        
        \vspace{2pt}
        
        \adjincludegraphics[viewport={\xminB\width} {\yminB\height} {\xmaxB\width} {\ymaxB\height}, clip, width=\textwidth]{images/renders/Stump_GT.JPG}
    \end{subfigure}

    \caption{\textbf{Unbounded Mesh Rendering:} Our method reconstructs both fine details and high-quality backgrounds on challenging Mip-NeRF 360 scenes~\cite{barron2022mipnerf360}.}
    \label{fig:real-world-unbounded}
\end{figure}

Conversely, the state-of-the-art unbounded method \milo~\cite{guedon2025milo} suffers from similar artifacts. 
This limitation can be directly attributed to its mesh-aware bidirectional consistency loss, which is most easily minimized by generating overly smooth geometry.
Consequently, the resulting meshes lack intricate surface details.
Finally, compared to our direct baseline, SOF~\cite{Radl2025SOF}, our extracted meshes exhibit finer details while successfully avoiding the majority of visual artifacts. 
These enhancements are driven by our proposed confidence framework, which provides a dedicated mechanism to explicitly resolve severe photometric ambiguities.
Finally, we demonstrate mesh rendering of challenging unbounded scenes from Mip-NeRF 360~\cite{barron2022mipnerf360} in \cref{fig:real-world-unbounded}.

\subsection{Ablation Studies}
\para{Individual Components}
\begin{table}[t]
    \centering
    \caption{
\textbf{Ablation Study for our Method:}
We analyze the effects of our individual contribution on both the Tanks \& Temples~\cite{Knapitsch2017tanks} and the ScanNet++ dataset~\cite{yeshwanthliu2023scannetpp}.
As we can see, all our proposed components result in notable improvements.
    }
\label{tab:tnt_ablation}
    \setlength{\tabcolsep}{2pt}
\resizebox{.98\linewidth}{!}{
    \begin{tabular}{l ccc @{\hskip 0.8cm} ccc}
    \toprule
& \multicolumn{3}{c@{\hskip 0.8cm}}{\textbf{Tanks \& Temples}} & \multicolumn{3}{c}{\textbf{ScanNet++}} \\
\cmidrule(l r{0.9cm}){2-4} \cmidrule(lr){5-7}
    Method & Precision\textsuperscript{$\uparrow$} & Recall\textsuperscript{$\uparrow$} & F1-Score\textsuperscript{$\uparrow$} & Precision\textsuperscript{$\uparrow$} & Recall\textsuperscript{$\uparrow$} & F1-Score\textsuperscript{$\uparrow$} \\
    \midrule
SOF~\cite{Radl2025SOF} (baseline) 
& 0.542 & 0.438 & 0.474 & 0.565 & 0.677 & 0.615 \\
+ Improved Appearance 
& 0.557 & 0.459 & 0.493 & 0.577 & 0.685 & 0.625 \\
+ $\msub{\mathcal{L}}{conf}$ 
& \cellcolor{tab_color!15}0.568 & \cellcolor{tab_color!15}0.479 & \cellcolor{tab_color!15}0.509 & \cellcolor{tab_color!15}0.605 & \cellcolor{tab_color!15}0.715 & \cellcolor{tab_color!15}0.655 \\
+ $\msub{\mathcal{L}}{color-var}$ 
& \cellcolor{tab_color!32}0.571 & \cellcolor{tab_color!49}0.491 
& \cellcolor{tab_color!32}0.519 & \cellcolor{tab_color!32}0.606 & \cellcolor{tab_color!32}0.722 & \cellcolor{tab_color!32}0.658 \\
+ $\msub{\mathcal{L}}{normal-var}$ 
& \cellcolor{tab_color!49}0.580 & \cellcolor{tab_color!49}0.491 & \cellcolor{tab_color!49}0.521 & \cellcolor{tab_color!49}0.613 & \cellcolor{tab_color!49}0.734 & \cellcolor{tab_color!49}0.668 \\
    \bottomrule
    \end{tabular}
}
\end{table}

We investigate the impact of our contributions in \cref{tab:tnt_ablation}, where we considered the Tanks \& Temples as well as ScanNet++ dataset.
Adding our improved appearance module leads to better mesh quality across both datasets; 
on its own, it outperforms \milo~\cite{guedon2025milo}, the current state-of-the-art method for unbounded meshes.
Adding $\msub{\mathcal{L}}{conf}$ leads to further improvements, with a significant $\mathbf{0.03}$ gain for ScanNet++.
This substantial boost can be attributed to the inherent ambiguities of indoor scenes, such as untextured walls, inconsistent lighting, and occasional blur.
The color variance loss further improves performance by constraining extreme scenarios, in which view-dependent appearance is represented with geometry;
naturally, this yields the strongest gains on Tanks \& Temples, where such high-frequency effects are abundant.
\noindent 

\noindent 
Conversely, $\msub{\mathcal{L}}{normal-var}$ leads to notable improvements for both datasets, with a stronger impact for ScanNet++.
This perfectly aligns with the indoor nature of the dataset, which is heavily dominated by planar structures (\eg, walls, tables); 
in such environments, enforcing consensus between per-primitive normals acts as a highly effective regularizer.
Overall, our techniques lead to F1-score improvements of $\mathbf{0.047}/\mathbf{0.053}$.

\begin{table}[t]
    \centering
    \caption{
\textbf{Ablation Study on the Decoupled Appearance Module:}
We compare appearance embeddings via surface reconstruction on the Tanks \& Temples dataset~\cite{Knapitsch2017tanks}.
Our proposed decoupled configuration yields the strongest results.
}
\label{tab:appearance}
    \setlength{\tabcolsep}{2pt}
\resizebox{.98\linewidth}{!}{
    \begin{tabular}{l ccc @{\hskip 0.8cm} l ccc}
    \toprule
& \multicolumn{3}{c@{\hskip 0.8cm}}{\textbf{Baselines (Standard)}} & & \multicolumn{3}{c}{\textbf{Ours (Ablations)}} \\
\cmidrule(l r{0.9cm}){2-4} \cmidrule(lr){6-8}
    Method & Precision\textsuperscript{$\uparrow$} & Recall\textsuperscript{$\uparrow$} & F1-Score\textsuperscript{$\uparrow$} & Method & Precision\textsuperscript{$\uparrow$} & Recall\textsuperscript{$\uparrow$} & F1-Score\textsuperscript{$\uparrow$} \\
    \midrule
VastG.~\cite{lin2024vastgaussian} & 0.541 & 0.440 & 0.475 & 
VastG. (Imp.) & \cellcolor{tab_color!15}0.557 & 0.445 & 0.484 \\
PGSR~\cite{chen2024pgsr} & 0.545 & 0.441 & 0.478 & 
w/o $l(\cdot)$ & \cellcolor{tab_color!49}0.559 & 0.441 & 0.483 \\
H3DGS~\cite{kerbl2024hierarchical} & 0.553 & 0.448 & 0.484 & 
w/o SSIM Dec. & \cellcolor{tab_color!49}0.559 & \cellcolor{tab_color!15}0.453 & \cellcolor{tab_color!32}0.490 \\
PPISP~\cite{deutsch2026ppisp} & 0.556 & \cellcolor{tab_color!32}0.456 & \cellcolor{tab_color!15}0.489 & 
\textbf{Ours} & \cellcolor{tab_color!15}0.557 & \cellcolor{tab_color!49}0.459 & \cellcolor{tab_color!49}0.493 \\
    \bottomrule
    \end{tabular}
}
\end{table}

\para{Decoupled Appearance}
We evaluate multiple variants of our decoupled appearance module in \cref{tab:appearance}, validating our SSIM-decoupled design;
here, we disabled all other components to isolate the effects of the appearance embeddings.
As baselines, we consider the appearance embeddings as used in VastGaussian~\cite{lin2024vastgaussian}, PGSR~\cite{chen2024pgsr}, Hierarchical-3DGS (H3DGS)~\cite{kerbl20233dgs}, as well as the concurrent work PPISP~\cite{deutsch2026ppisp}. 
Our architectural enhancements to the VastGaussian baseline (VastG. (Imp.)) already yield a notable improvement in F1-score.
Disregarding the luminance term (w/o $l(\cdot)$) attains similar results compared to VastG. (Imp.), demonstrating that this loss term alone does not benefit scene reconstruction unless appearance is also explicitly compensated for.
Using the $\iapp$ for $\msub{\mathcal{L}}{D-SSIM}$ (w/o SSIM Decoupling) greatly improves surface extraction results, however, we obtain even stronger results when using $\irender$ for contrast and structure terms (\textbf{Ours}).
Both of these variants slightly outperform PPISP, which targets inference; conversely, we optimize for 3D coherence and faithful mesh extraction.
\newcommand{\overlaylabelbeta}[2]{%
    \begin{tikzpicture}
        \node[anchor=south west, inner sep=0] (image) at (0,0) {\includegraphics[width=0.31\linewidth]{#1}};
        \begin{scope}[x={(image.south east)},y={(image.north west)}]
            \node[anchor=south west, 
                  fill=black, fill opacity=0.5, 
                  text opacity=1, 
                  inner sep=2pt, 
                  rounded corners=1pt] 
            at (0.01, 0.01) { 
                \sffamily\scriptsize{\textcolor{white}{#2}} 
            };
        \end{scope}
    \end{tikzpicture}%
}

\para{Confidence}
\begin{wrapfigure}{r}{0.45\textwidth}
    \centering
    \vspace{-10pt}
    \includegraphics[width=\linewidth]{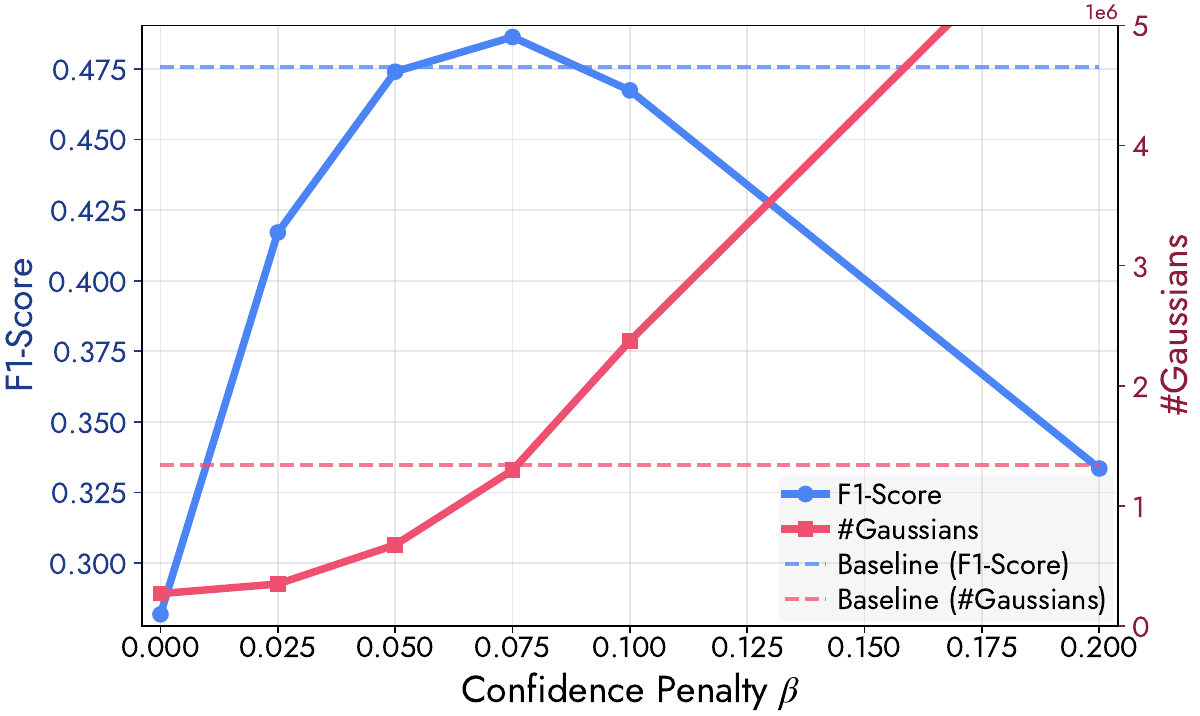} \\
    \overlaylabelbeta{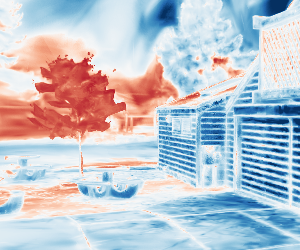}{$\beta = 0.05$}
    \overlaylabelbeta{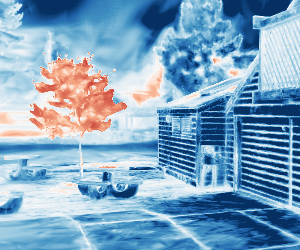}{$\beta = 0.075$}
    \overlaylabelbeta{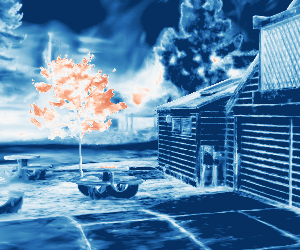}{$\beta = 0.1$}
    \vspace{-5pt}
\caption{
\textbf{$\beta$-Ablation Study:}
We evaluate surface reconstruction quality for varying choices of $\beta$.
Our chosen value of $7.5 \times 10^{-2}$ achieves the best performance.
}
    \vspace{-10pt}
    \label{fig:beta_ablation}
\end{wrapfigure}
%
Finally, we ablate our choice of $\beta$, which is the primary hyperparameter in our implementation.
To this end, we present results for the Tanks \& Temples dataset on surface reconstruction, using $\beta \in [0, 0.2]$, \cf in \cref{fig:beta_ablation}.
Here, we only used our improved appearance model while removing our variance losses to keep the confidence isolated.
As can be seen, using $\beta=0$ provides no incentive to the model to maximize $\hat{C}$, as $\msub{\mathcal{L}}{conf}$ reduces to $\msub{\mathcal{L}}{rgb} \cdot \hat{C}$, resulting in under-reconstruction with very few points.
Using a value of $\beta = 0.2$ results in the opposite effects, where the penalty is too harsh to allow for any uncertainty;
this results in a huge number of primitives, as the large $\msub{\mathcal{L}}{rgb}$ causes massive absolute gradients, negatively impacting densification.

Interestingly, a choice of $\beta=0.05$ may represent a \emph{light} variant of our model;
achieving strong mesh extraction results, while keeping the number of primitives low.
Ultimately, our chosen $\beta = 0.075$ achieves a good balance, achieving the highest F1-score, while using slightly fewer primitives than the baseline (w/o confidence).
We refer to \cref{app:confidence} for a more thorough discussion, and more visualizations.
\section{Discussion, Limitations and Future Work}

Our proposed method effectively extracts high-fidelity surfaces, but it is not without limitations. 
Fundamentally, our framework relies on the assumption of a moderately dense capture setting.
While our confidence-driven losses successfully resolve photometric ambiguities and greatly reduce artifacts, occasional holes or cavities may still manifest in severely occluded or unobserved regions.
We believe that for such scenarios, or specifically in sparse-view capture settings, integrating multi-view geometric constraints~\cite{chen2024pgsr} or monocular priors~\cite{chen2024vcr, li2025geosvr} effectively compensates for the lack of densely captured data.

Furthermore, our method struggles to reconstruct detailed geometry for distant background regions.
Due to the insufficient view density and lack of ``close-up'' captures, the 3DGS reconstruction is insufficiently densified in these areas.
We believe that pre-trained multi-view diffusion models~\cite{fischer2025flowr, wu2025difix3d}, along with tailored densification recipes, could provide additional details for such regions.

Our confidence framework opens exciting avenues for broader downstream applications.
We hypothesize that this formulation could readily be adapted into a heuristic-free densification framework for general 3DGS:
As demonstrated in \cref{fig:beta_ablation}, the confidence penalty parameter $\beta$ already acts as a hyperparameter that balances mesh quality and primitive count.
We also believe that this approach may benefit other downstream 3DGS applications, such as 3D inpainting or super-resolution training.
\section{Conclusion}
In this work, we presented a confidence-steered framework tailored for high-fidelity surface extraction from 3D Gaussian Splatting.
By introducing learnable, per-primitive confidence values alongside color and normal variance losses, our method dynamically mitigates the ambiguities caused by high-frequency view-dependent appearance.
Furthermore, based on a rigorous analysis of the popular D-SSIM loss, we proposed an architecturally improved, decoupled appearance embedding that explicitly compensates for unconstrained real-world illumination changes. 
Ultimately, our contributions allow us to achieve state-of-the-art results for unbounded mesh extraction, all while strictly avoiding the inherent computational overhead of prior-dependent methods to maintain a highly competitive runtime (optimization in $<20$ minutes). 

\section*{Acknowledgements}
We thank Diego Gomez and Ziyi Zhang for their help with evaluation, as well as Nissim Maruani for his help with illustrations.

%
%
\bibliographystyle{splncs04}
\bibliography{main}

\appendix



\section{Implementation Details}
To aid reproducibility, this section contains additional implementation details for our method.
%

\subsection{Decoupled D-SSIM}
\label{app:decoupled}
%
We made the following implementation changes to the decoupled appearance module from VastGaussian~\cite{lin2024vastgaussian}:
First, we replaced the sigmoid activation function with $\exp(\cdot)$, which allows the appearance embedding to compensate for overexposed images.
This change also enables a straightforward zero-initialization by setting $\mathbf{W}=\mathbf{0},\ \mathbf{b}=\mathbf{0}$ for the final convolutional layer, which stabilizes training in early iterations.
We detach $\text{ds}_{32}(\irender)$ when used as input for the CNN, preventing gradients of the appearance embedding from propagating to the underlying 3DGS model.
Whereas VastGaussian~\cite{lin2024vastgaussian} center-crops the input images to ensure they are perfectly divisible into $32\times 32$ tiles for downsampling, our approach utilizes reflection-padding instead.
For the CNN-input, we concatenate an additional, lightweight positional encoding, containing $u,v$-coordinates as well as a radial distance $r$:
\begin{equation}
    r(u,v) = \sqrt{u^2 + v^2},
\end{equation}
with $u,v \in [-1,1]$.
This allows the network to compensate for vignetting~\cite{goldman2010vignette}.
Thus, the final input to the CNN is $[\cancel{\nabla}\text{ds}_{32}(\irender), \bm{\rho}_i, u, v, r(u,v)]$, 
where the input dimensionality of the CNN is increased to $70$, compared to $67$ for VastGaussian.

For optimal performance, we design a custom fused CUDA kernel based on~\cite{mallick2024taming}, which is up to $5\times$ faster than the na\"ive PyTorch implementation.

\subsection{Confidence}
For numerical stability, the rendered confidence values are clamped to $\hat{C} \in [0.001,\ 5.0]$ to ensure that the gradients remain in a reasonable range.
Recall our definition of $\msub{\mathcal{L}}{conf}$:
\begin{equation}
    \msub{\mathcal{L}}{conf} = \msub{\mathcal{L}}{rgb} \cdot \hat{C} - \beta \cdot \log \hat{C}.
\end{equation}
The gradient w.r.t. $\hat{C}$ is:
\begin{equation}
    \frac{\partial \msub{\mathcal{L}}{conf}}{\partial \hat{C}} = \msub{\mathcal{L}}{rgb} - \frac{\beta}{\hat{C}}.
\end{equation}
As $\hat{C} \rightarrow 0$, this quantity explodes; 
with our minimum confidence value, the gradient reaches at most $ -75$ with $\beta = 7.5 \times 10 ^{-2}$.
The maximum value was chosen to ensure that the gradients flowing back to individual Gaussians remain reasonably scaled, preventing further densification of already well-reconstructed regions.
In addition, note that we detach $w_i(\bm{r})$ in \cref{eq:conf_alphablending}, as we do not want the Gaussians to modify their opacity guided by confidence gradients.

\para{Confidence-Steered Densification}
We find that our adapted densification scheme is vital to control primitive counts, resulting in model sizes comparable to our baseline SOF~\cite{Radl2025SOF}.
The ablation study in \cref{tab:tnt_ablation_suppl} shows that disabling our confidence-steered densification yields a slightly lower F1-score with a $20\%$ increase in primitive count.
\begin{table}[ht!]
    \centering
    \setlength{\tabcolsep}{4pt}
    \caption{
\textbf{Extended Ablation Studies} for the Tanks \& Temples dataset \cite{Knapitsch2017tanks}.
We investigate the impact of our confidence-steered densification, 
our choice of using $\msub{\mathcal{L}}{conf}$ from iteration $500$, and verify our chosen hyperparameters.
}
\resizebox{.98\linewidth}{!}{
\begin{tabular}{lrrrrrrrc}
\toprule
 & Barn & Caterpillar & Courthouse & Ignatius & Meetingroom & Truck & Average & \# Gaussians \\
\midrule
Ours & 0.534 & 0.472 & 0.333 & 0.782 & 0.372 & 0.634 & \cellcolor{tab_color!32}0.521 & \cellcolor{tab_color!32}1.27M\\
Ours w/o adapted densification & 0.537 & 0.463 & 0.338 & 0.769 & 0.369 & 0.623 & 0.516 & 1.53M\\
\midrule
Ours & 0.534 & \cellcolor{tab_color!49} 0.472 & \cellcolor{tab_color!49} 0.333 & \cellcolor{tab_color!32} 0.782 & \cellcolor{tab_color!49} 0.372 & \cellcolor{tab_color!32} 0.634 & \cellcolor{tab_color!49} 0.521 & \cellcolor{tab_color!49}1.27M\\
Ours w/ $\msub{\mathcal{L}}{conf}$ from iter. 2.5K & 0.532 & \cellcolor{tab_color!32} 0.465 & \cellcolor{tab_color!15} 0.330 & \cellcolor{tab_color!49} 0.783 & \cellcolor{tab_color!15} 0.363 & \cellcolor{tab_color!49} 0.637 & \cellcolor{tab_color!32} 0.518 & \cellcolor{tab_color!32}1.34M\\
Ours w/ $\msub{\mathcal{L}}{conf}$ from iter. 5K & \cellcolor{tab_color!49} 0.537 & \cellcolor{tab_color!32} 0.465 & 0.323 & \cellcolor{tab_color!15} 0.780 & \cellcolor{tab_color!32} 0.366 & \cellcolor{tab_color!15} 0.632 & \cellcolor{tab_color!15} 0.517 & 1.42M \\
Ours w/ $\msub{\mathcal{L}}{conf}$ from iter. 10K & \cellcolor{tab_color!49} 0.537 & 0.447 & 0.328 & 0.764 & 0.359 & 0.616 & 0.508 & 1.43M \\
Ours w/ $\msub{\mathcal{L}}{conf}$ from iter. 15K & \cellcolor{tab_color!15} 0.536 & 0.435 & \cellcolor{tab_color!32} 0.332 & 0.759 & 0.349 & 0.594 & 0.501 & \cellcolor{tab_color!15}1.36M \\
\midrule
Ours & \cellcolor{tab_color!32} 0.534 & \cellcolor{tab_color!32} 0.472 & \cellcolor{tab_color!32} 0.333 & \cellcolor{tab_color!49} 0.782 & \cellcolor{tab_color!49} 0.372 & \cellcolor{tab_color!15} 0.634 & \cellcolor{tab_color!49} 0.521 & 1.27M\\
Ours w/ $\msub{\lambda}{normal-var}=0.05$ & \cellcolor{tab_color!15}0.531 & \cellcolor{tab_color!49} 0.476 & 0.326 & 0.765 & 0.349 & \cellcolor{tab_color!49} 0.643 & \cellcolor{tab_color!15} 0.515 & 1.25M\\
Ours w/ $\msub{\lambda}{normal-var}=0.0005$ & 0.528 & 0.460 & \cellcolor{tab_color!49} 0.338 & \cellcolor{tab_color!32} 0.776 & \cellcolor{tab_color!32} 0.369 & \cellcolor{tab_color!15} 0.634 & \cellcolor{tab_color!32} 0.518 & 1.22M\\
Ours w/ $\msub{\lambda}{color-var}=1.0$ & 0.528 & \cellcolor{tab_color!15} 0.464 & \cellcolor{tab_color!15} 0.328 & \cellcolor{tab_color!15} 0.771 & 0.364 & \cellcolor{tab_color!32} 0.639 & \cellcolor{tab_color!15} 0.515 & 1.26M\\
Ours w/ $\msub{\lambda}{color-var}=0.1$ & \cellcolor{tab_color!49} 0.535 & 0.462 & 0.319 & 0.761 & \cellcolor{tab_color!15} 0.368 & 0.626 & 0.512 & 1.24M\\
\bottomrule
\end{tabular}
}
    \label{tab:tnt_ablation_suppl}
\end{table}

By default, $\msub{\mathcal{L}}{conf}$ is activated in iteration $500$;
we validate this choice in \cref{tab:tnt_ablation_suppl}.
We observe that the impact on the final F1-score is minor, while primitive counts increase slightly.
Notably, our final configuration achieves superior F1-scores than our baseline SOF~\cite{Radl2025SOF}, while using $6\%$ fewer primitives.

\subsection{Variance Losses}
To further evaluate the robustness of our framework, we ablate the weights of our proposed variance losses (\cf \cref{tab:tnt_ablation_suppl}). 
Specifically, we vary the normal variance weight $\msub{\lambda}{normal-var}$ ($0.005$ by default) by an order of magnitude $\{0.0005, 0.05\}$, and adjust the color variance weight $\msub{\lambda}{color-var}$ ($0.5$ by default) to $\{0.1, 1.0\}$. 

Overall, the reconstruction accuracy remains exceptionally stable across these variations, reinforcing that our variance losses do not rely on precise, scene-specific tuning. 
We observe that the color variance term is slightly more sensitive to weight variations.
However, the performance degrades gracefully, resulting in only a minor F1-score drop of $0.009$. 
Ultimately, our default configuration ($\msub{\lambda}{color-var}=0.5, \msub{\lambda}{color-var} = 0.005$) yields the highest overall performance.

\section{Reproducibility}
\label{app:reproducibility}
In this section, we include further details on the used datasets and how we reproduced the results for baseline methods, paving the way for a fair and reproducible comparison.

\subsection{Datasets}
\label{app:dataset}
To enable a fair and consistent comparison, we reproduce all presented results in the paper using the latest, publicly available codebases.
During this process, we identified a discrepancy in data preprocessing: while some methods utilize the datasets provided by GOF~\cite{yu2024gof}, others rely on custom COLMAP scripts, such as the one from PGSR~\cite{chen2024pgsr}. 
To maintain strict reproducibility, we standardized our comparison by using the publicly accessible data provided by GOF across all experiments.

\para{Sensitivity to Data Preprocessing}
To demonstrate the impact of custom data preprocessing, we conduct an experiment on the Barn scene from the Tanks \& Temples dataset~\cite{Knapitsch2017tanks}, using the preprocessing pipeline from PGSR~\cite{chen2024pgsr}.
We selected this scene due to the significant discrepancy between reported and reproduced results; for instance, while \milo~\cite{guedon2025milo} reports an F1-score of $\mathbf{0.59}$, our reproduction yielded only $\mathbf{0.541}$.
While our method previously obtained an F1-score of $\mathbf{0.534}$ for Barn, using the custom preprocessed data leads to a score of $\mathbf{0.616}$ (an improvement of $\mathbf{+0.082}$ that is solely caused by a difference in the input data).

This result clearly shows that comparing metrics across different papers may be misleading, if the underlying data preprocessing is not standardized.
Consequently, all reported results were evaluated under identical conditions.

\para{ScanNet++}
\newcommand{\snppanno}[2]{%
    \stackinset{l}{1pt}{t}{1pt}{
        \begin{tikzpicture}
            \node[fill=black, fill opacity=0.5, text opacity=1, 
                  inner sep=1.5pt, rounded corners=0.5pt] 
            {\sffamily\fontsize{2}{3}\selectfont{\textcolor{white}{#2}}};
        \end{tikzpicture}%
    }{#1}
}
\begin{figure*}[ht!]
\scriptsize\sffamily
\setlength{\tabcolsep}{1pt}%
\setlength{\fboxsep}{0pt}%
\setlength{\fboxrule}{0.5pt}%
\renewcommand{\arraystretch}{1.1}%
\resizebox{.99\linewidth}{!}{
\begin{tabular}{ccc}

\vspace{-0.2cm}\\
\snppanno{\includegraphics[width=0.19\linewidth]{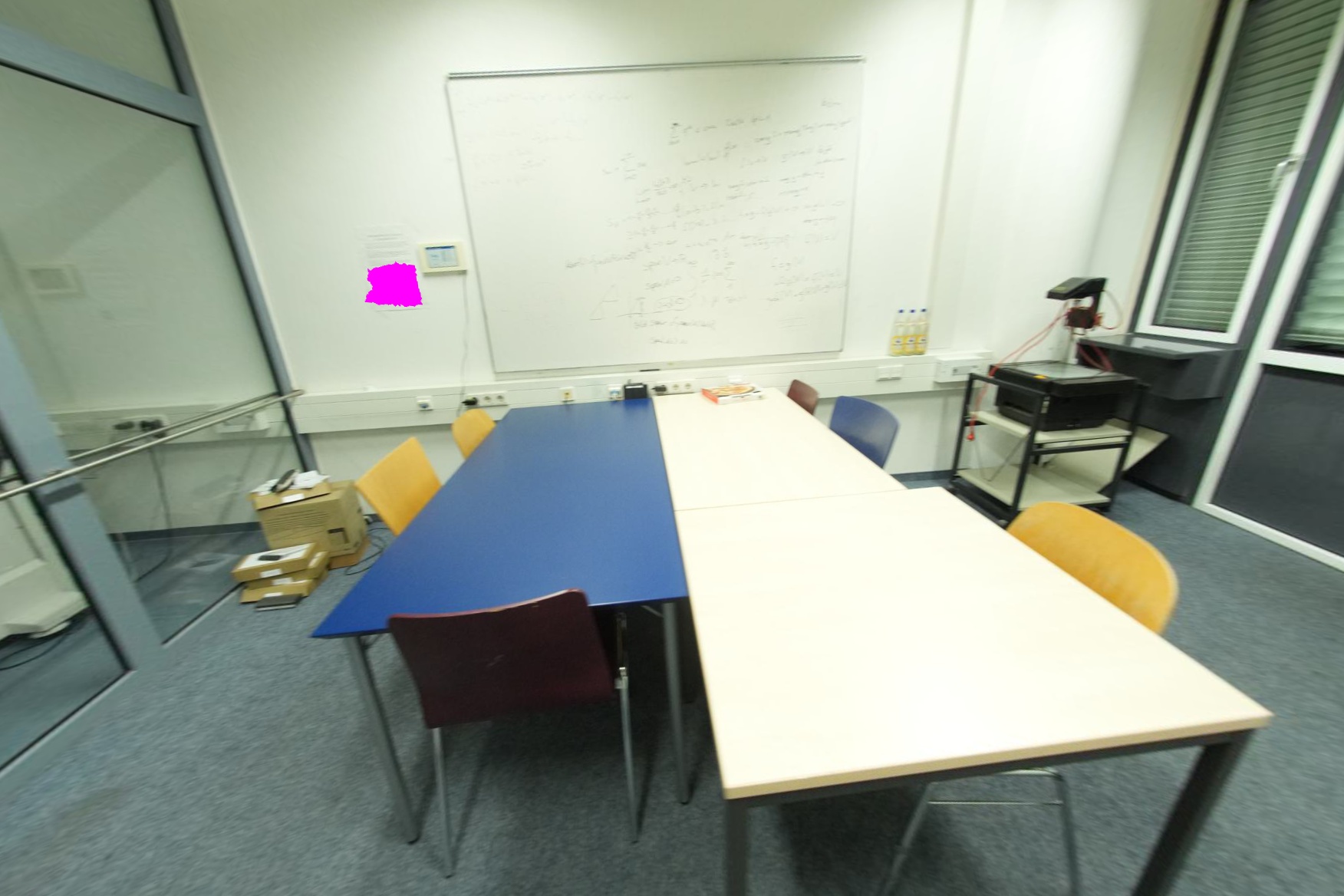}}{\texttt{5a269ba6fe} (189)} &
\snppanno{\includegraphics[width=0.19\linewidth]{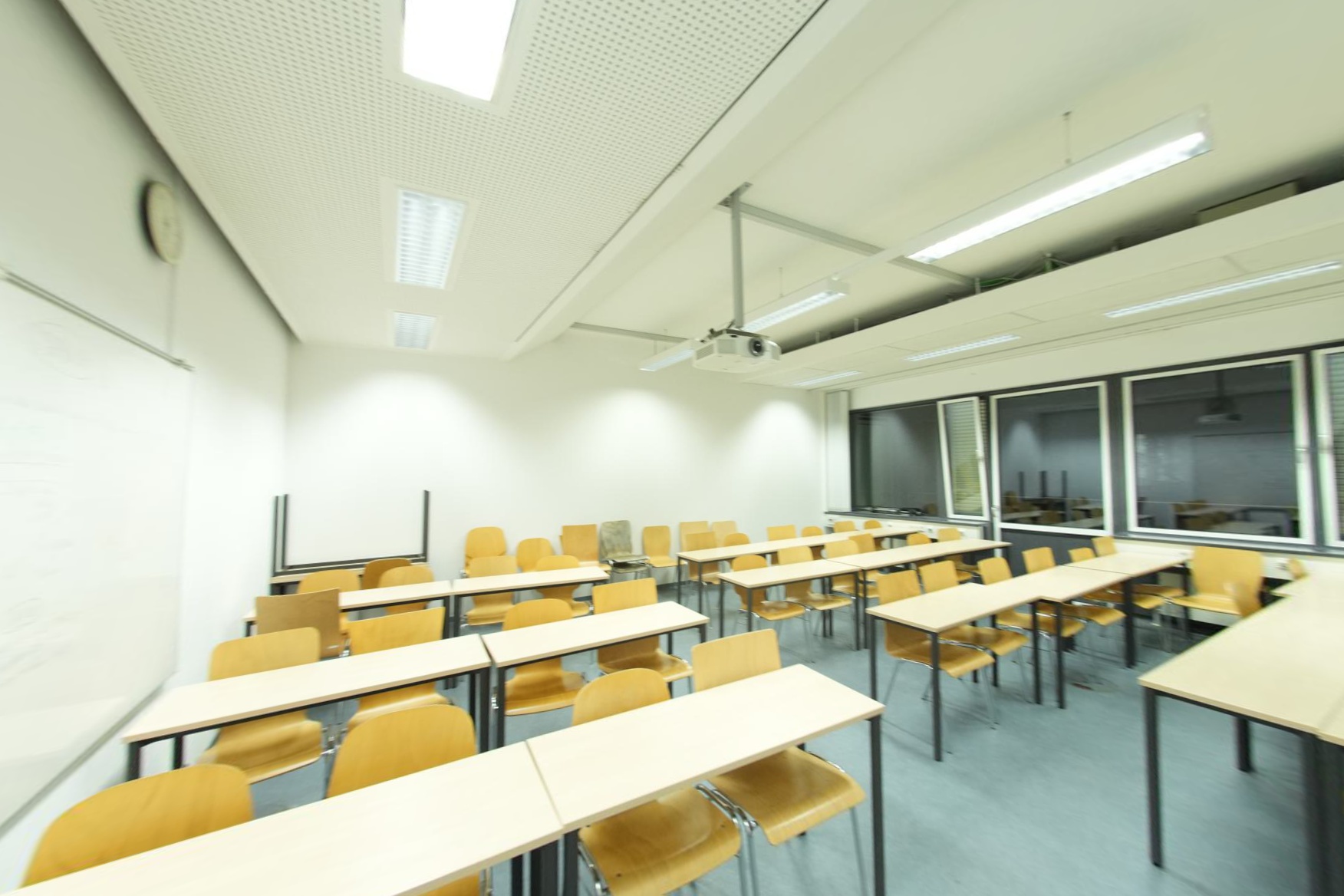}}{\texttt{08bbbdcc3d} (295)} &
\snppanno{\includegraphics[width=0.19\linewidth]{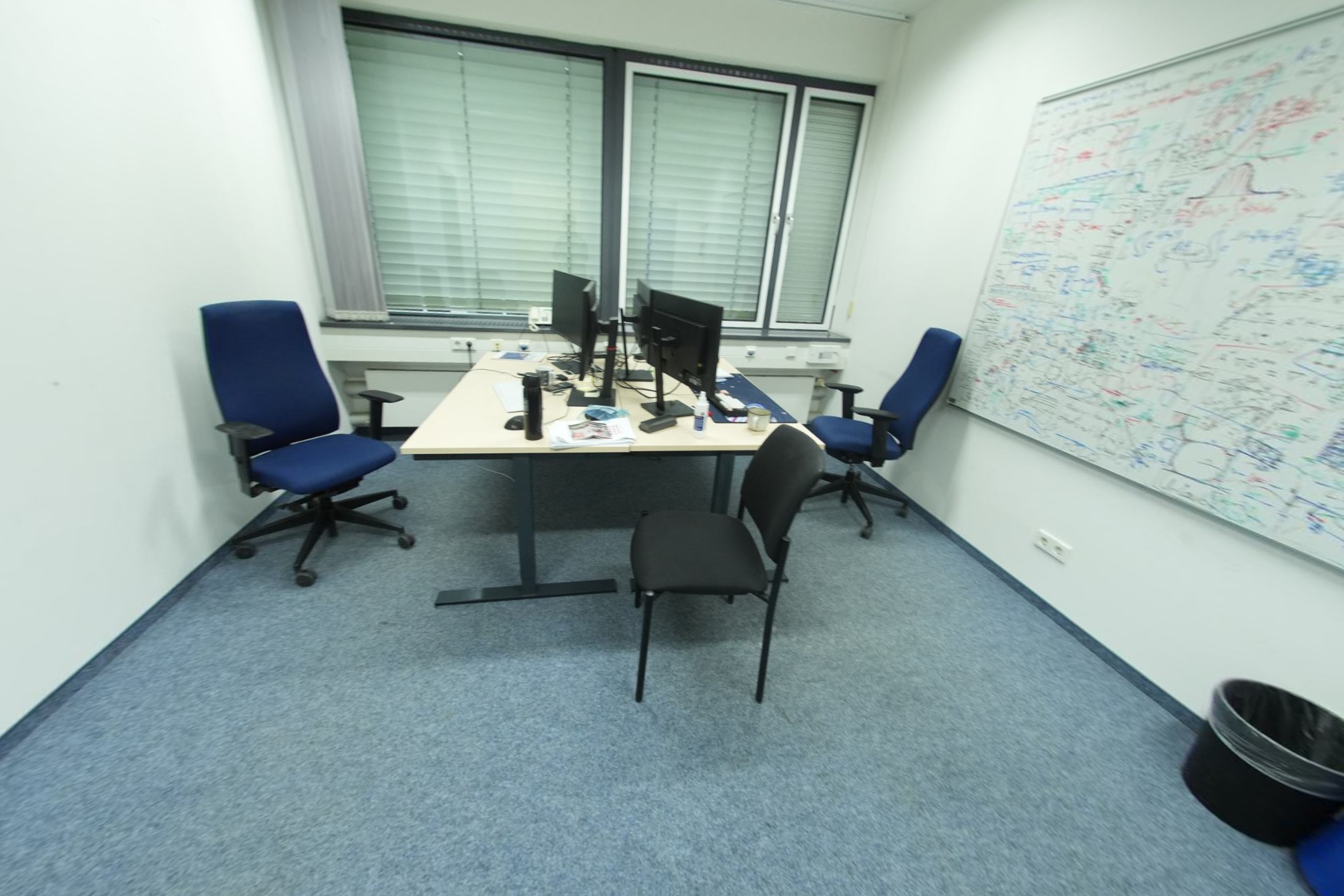}}{\texttt{39f36da05b} (369)} \\ \addlinespace[1pt]

\snppanno{\includegraphics[width=0.19\linewidth]{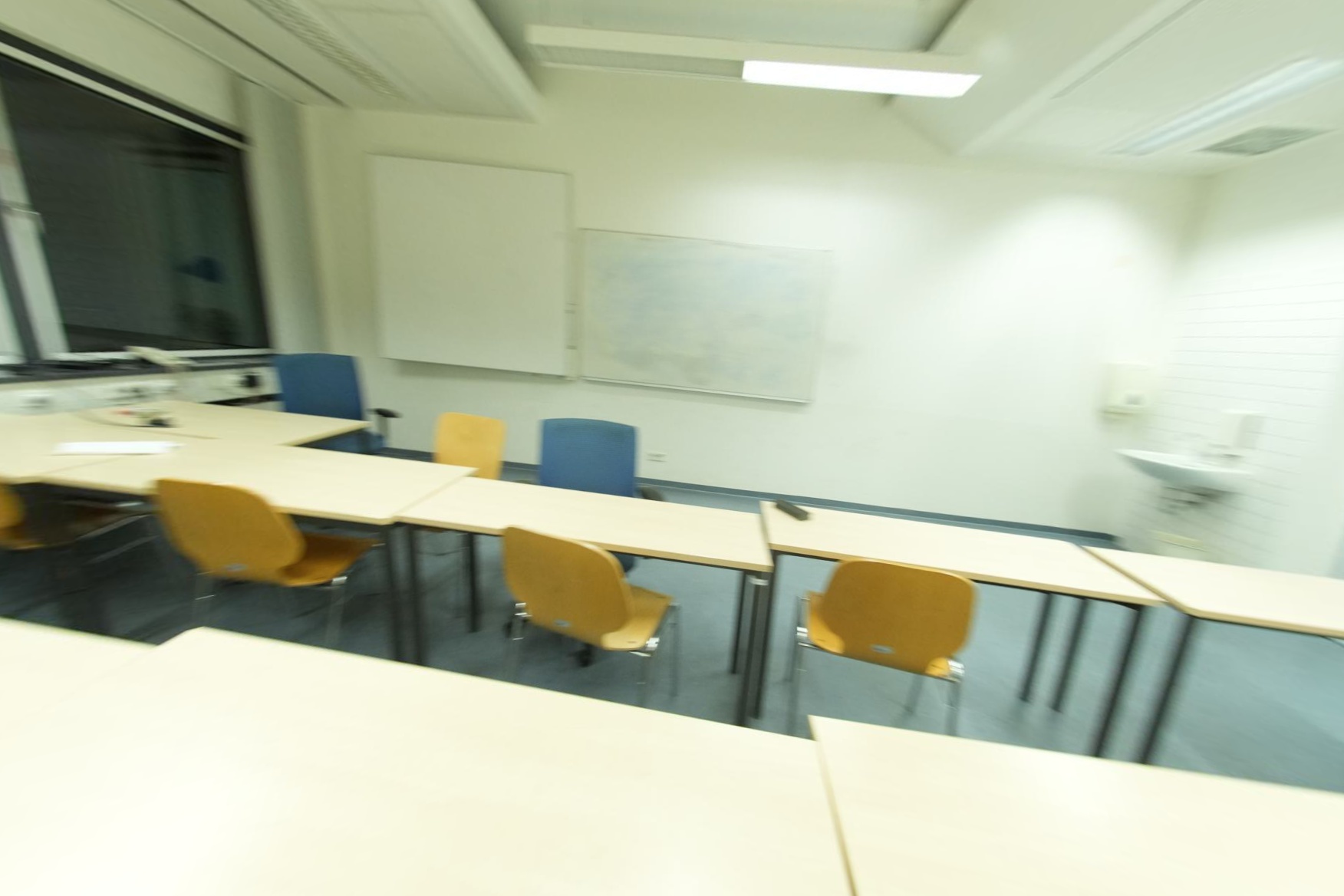}}{\texttt{dc263dfbf0} (280)} 
&
\snppanno{\includegraphics[width=0.19\linewidth]{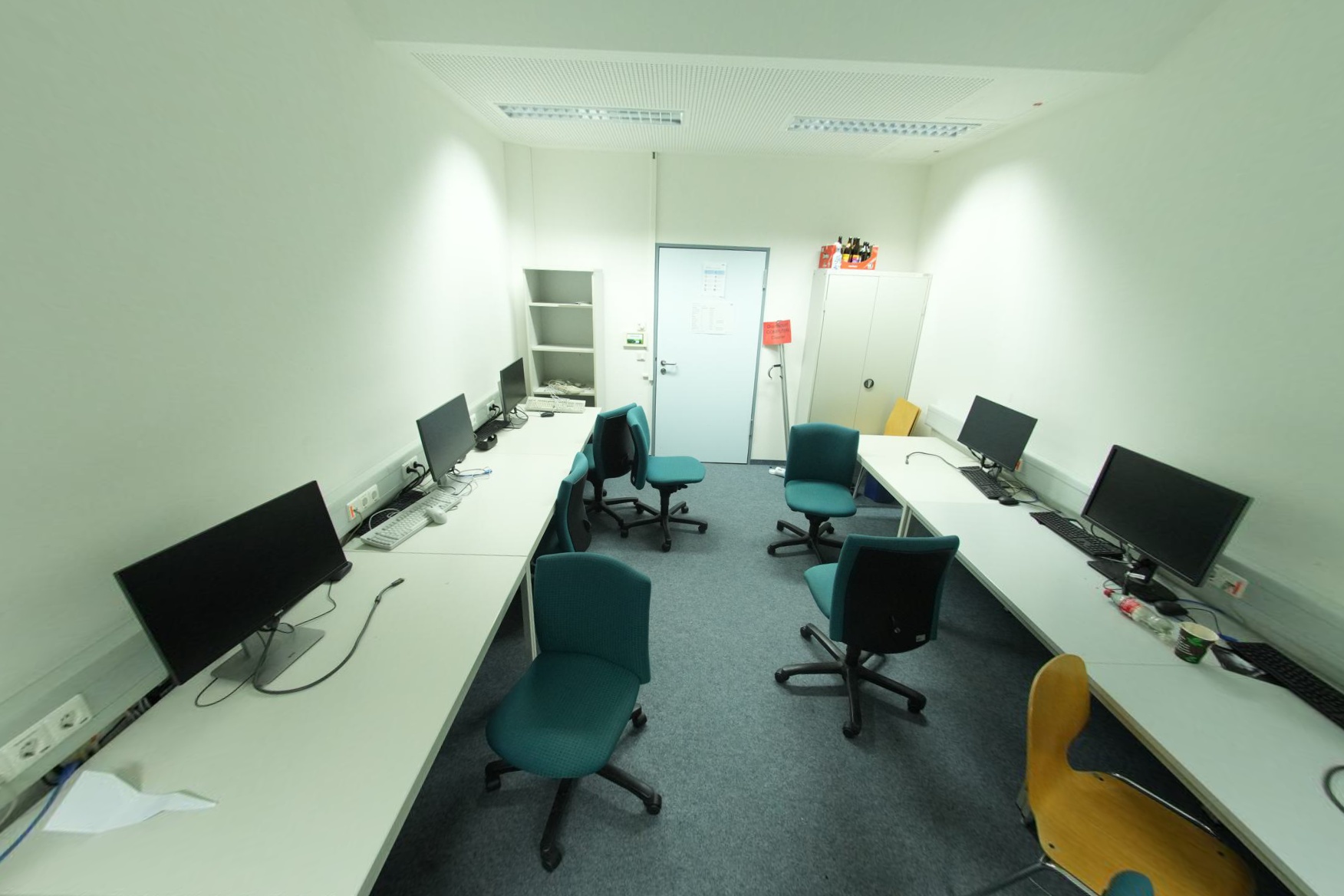}}{\texttt{ef18cf0708} (289)}
&
\snppanno{\includegraphics[width=0.19\linewidth]{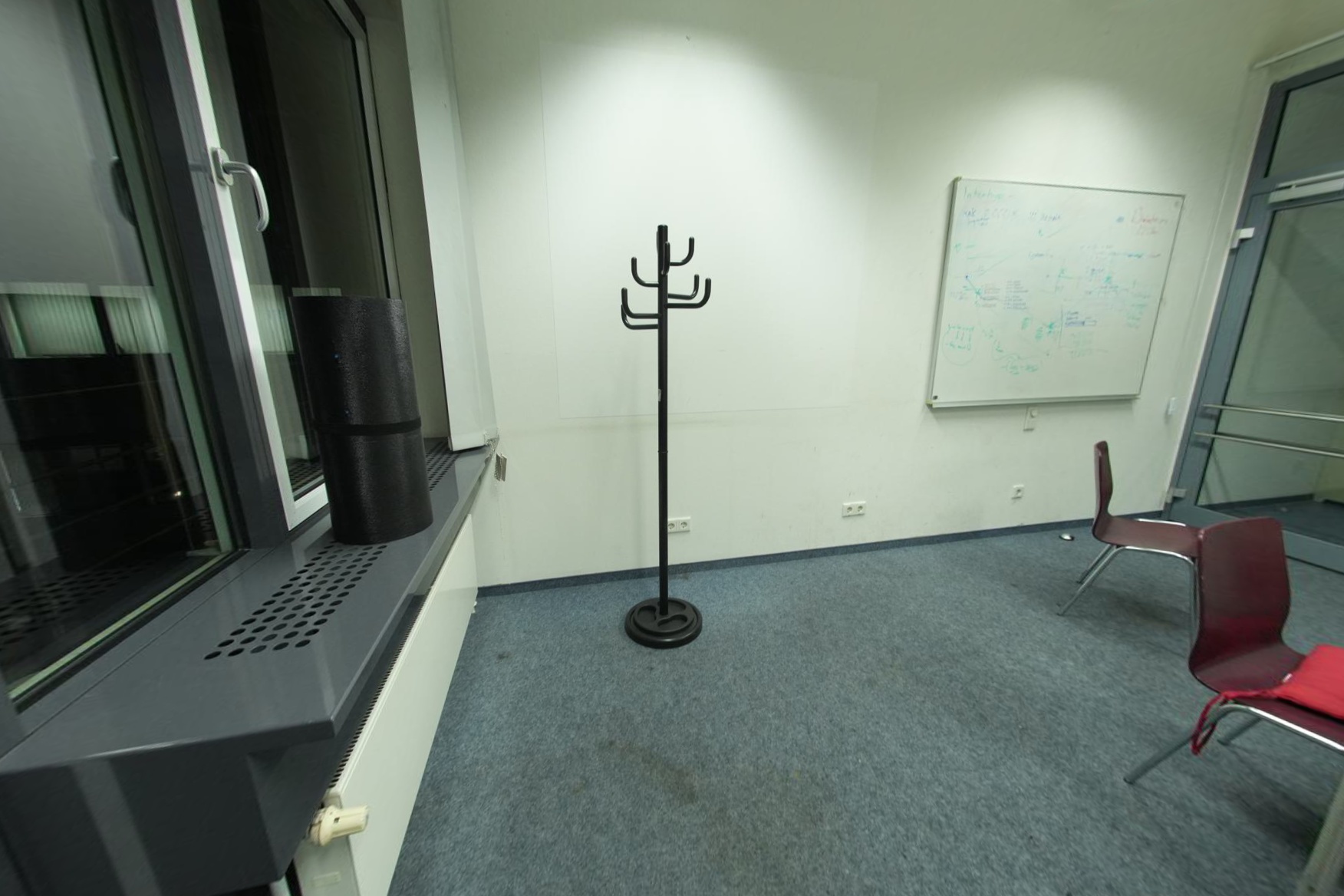}}{\texttt{fb564c935d} (161)}  \\
\end{tabular}
}
  \caption{\label{fig:snpp}%
    \textbf{ScanNet++ dataset~\cite{yeshwanthliu2023scannetpp}}:
We show example images from our selected scenes, with the number of images per-scene inset.
All selected scenes contain challenging reflections (\eg \texttt{5a269ba6fe, fb564c935d}), textureless areas, and motion blur.
  }
\end{figure*}
We also include comparisons on the ScanNet++ dataset~\cite{yeshwanthliu2023scannetpp} to evaluate our method in challenging, real-world settings.
Given the computationally prohibitive scale of evaluating all baselines across the entire benchmark, 
we selected a representative subset of six scenes (\cf \cref{fig:snpp}) ensuring a diverse coverage of reflections, textureless areas, and motion blur.
Each scene contains between $161$ and $369$ images.
Across all experiments, we used undistorted images at half resolution during optimization (image resolution of $584\times 876$, utilizing the \texttt{--r 2} flag).

For evaluation, we compute the F1-score, using the evaluation script adapted from Tanks \& Temples, setting the inlier threshold $\tau = 0.05$ throughout.
We follow standard benchmark practice and compute metrics exclusively within the bounding volume of the ground-truth mesh.
Other than that, the evaluation protocol strictly follows Tanks \& Temples.

\subsection{Implemented Concurrent Works}
\para{PGSR}
To reproduce PGSR~\cite{chen2024pgsr}, we use the latest available public codebase\footnote{\url{https://github.com/zju3dv/PGSR}}.
We find that in contrast to other methods~\cite{Radl2025SOF, yu2024gof}, PGSR includes the test views during training, which provides a significant advantage compared to other baselines.
Further, the argument \texttt{--ncc\_scale=0.5} enables the use of full-resolution images for the multi-view photometric loss, compared to the standard half-resolution images utilized by all other methods.
Thus, we add the \texttt{--eval} flag and set \texttt{ncc\_scale=1.0} for a fair comparison.

We also find that the TSDF fusion implementation makes use of the \emph{ground-truth point cloud} during mesh extraction.
Concretely, an axis-aligned bounding box of the ground-truth point cloud is used to clip the 3D points extracted from the final mesh and to derive the per-scene voxel size. 
This use of privileged information during mesh extraction leads to an advantage for PGSR.
We did not modify this protocol for our evaluation, but note that ground-truth point clouds are not available in practical scenarios, where PGSR would require careful tuning of voxel size and per-scene bounds.
Our approach, on the other hand, is completely agnostic to the scene scale, as it is solely dependent on the 3DGS point cloud.

\para{\milo}
To reproduce \milo~\cite{guedon2025milo}, we used the latest available public codebase\footnote{\url{https://github.com/Anttwo/MILo}}.
For all presented results, we used the GOF rasterization backend with the \emph{base} configuration;
this configuration was also reported in~\cite{guedon2025milo}.
Upon closer inspection of the provided data and evaluation protocol, we find that the images used for Tanks \& Temples differ in resolution from our data ($1954\times1090$ for Ours, $1500\times835$ for \milo);
we confirmed this in private correspondence with the authors.

Consequently, we reproduced their results using our data, following the same hyperparameter settings as in their work; we note that the difference to the reported metrics is nonetheless marginal.

\para{QGS}
To reproduce QGS~\cite{zhang2025qgs}, we use the latest available public codebase\footnote{\url{https://github.com/will-zzy/QGS}} with multi-view losses enabled; 
these losses were adapted from PGSR~\cite{chen2024pgsr}.
We find that for training, QGS also ignores the train/test split and uses full-resolution images.
For a fair comparison, we add the \texttt{--eval} flag and set \texttt{ncc\_scale=1.0}.

In contrast to PGSR, QGS does not rely on ground-truth information for TSDF fusion, and uses a constant voxel size for all scenes.

\para{Other Methods}
We also reproduced the results for GOF~\cite{yu2024gof} and SOF~\cite{Radl2025SOF}, using the latest publicly available 
codebases\footnote{\url{https://github.com/autonomousvision/gaussian-opacity-fields}}\footnote{\url{https://github.com/r4dl/SOF/}}.
No modifications were necessary, as both methods already used the data provided by GOF.

\subsection{Appearance Embeddings}
For completeness, we quickly describe the appearance embeddings used for comparison.

\para{PGSR Appearance Embedding}
PGSR~\cite{chen2024pgsr} uses learnable coefficient $\{a_i, b_i\} \in \Rbb^2$, which define a per-image mapping:
\begin{equation}
    \iapp = \exp(a_i)\irender + b_i.
\end{equation}
Importantly, the exposure-adjusted image is only used in the $\mathcal{L}_1$-loss, given $\msub{\mathcal{L}}{SSIM} < 0.5$.
While this method is efficient, it cannot compensate for \eg vignetting, explaining the benefit of our method.

\para{Hierarchical 3DGS Appearance Embedding}
Hierarchical 3DGS~\cite{kerbl2024hierarchical} instead employs a learnable, per-image affine mapping ${\bm{A}_i, \bm{b}_i}$:
\begin{equation}
    \iapp = \bm{A}_i \irender + \bm{b}_i,
\end{equation}
which corresponds to a global mapping, similarly to PGSR. 
Note that $\bm{A}_i \in \Rbb^{3\times 3}, \bm{b}_i \in \Rbb^3$.
In contrast to PGSR, the use of a full transformation matrix enables the model to apply distinct scaling to each color channel, allowing it to capture true chromaticity shifts rather than being restricted to luminance.

\para{PPISP}
PPISP~\cite{deutsch2026ppisp} uses a physically-plausible model of learnable, per-image parameters, explicitly compensating for exposure, vignetting, color correction as well as the camera response function;
see~\cite{deutsch2026ppisp} for details.
This more sophisticated mapping performs better than the previously discussed variants due to its ability to compensate for effects such as vignetting.

Importantly, PPISP is designed with inference in mind, proposing a custom controller module for novel view synthesis;
in contrast, we are interested in explaining camera-dependent effects to enable coherent 3D scene reconstruction --- which is vital for faithful mesh extraction.

\section{Mathematical Derivations}
In this section, we present the necessary derivations to implement the presented variance losses.
\subsection{Color Variance Loss}
The color variance loss $\msub{\mathcal{L}}{color-var}$ represents the weighted RGB variance for each pixel:
\begin{equation}
    \msub{\mathcal{L}}{color-var} = \sum_{i=0}^{N-1} w_i(\bm{r})  \left\lVert\text{sh}(\bm{\theta}_i, \bm{d})-\igt\right\lVert_2^2 .
\end{equation}
The gradients for the view-dependent color can be simply computed using power and chain rule:
\begin{equation}
\frac{\partial \msub{\mathcal{L}}{color-var}}{\partial \text{sh}(\bm{\theta}_i, \bm{d})} = 2 w_i(\bm{r})  (\text{sh}(\bm{\theta}_i, \bm{d}) - \igt).
\end{equation}
We can compute the gradient with regard to $\alpha_i$ (omitting $\bm{r}$ for notational convenience) as follows:
\begin{align}
    \frac{\partial \msub{\mathcal{L}}{color-var}}{\partial w_i} &= \left\lVert\text{sh}(\bm{\theta}_i, \bm{d})-\igt\right\lVert_2^2,\\
    \frac{\partial w_i}{\partial \alpha_i} &=  T_i,\\
    \frac{\partial \msub{\mathcal{L}}{color-var}}{\partial \alpha_i} =  
    \frac{\partial \msub{\mathcal{L}}{color-var}}{\partial w_i}  \frac{\partial w_i}{\partial \alpha_i} &= 
 \left\lVert\text{sh}(\bm{\theta}_i, \bm{d})-\igt\right\lVert_2^2 T_i.
\end{align}
This gradient does not account for the effect of $\alpha_i$ on the Gaussians $j > i$, which include $(1-\alpha_i)$ in $T_j$. Because $\alpha_i < 1$ by definition, $\frac{\partial w_j}{\partial \alpha_i} = -\frac{w_j}{1 - \alpha_i}$ holds and
\begin{equation}
    \sum_{j=i+1}^{N-1} \left\lVert\text{sh}(\bm{\theta}_j, \bm{d})-\igt\right\lVert_2^2 \left( - \frac{w_j }{1 - \alpha_i } \right) = - \frac{1}{1 - \alpha_i} \sum_{j=i+1}^{N-1} w_j  \left\lVert\text{sh}(\bm{\theta}_j, \bm{d})-\igt\right\lVert_2^2.
\end{equation}
The final gradient for $\alpha_i$ is then:
\begin{equation}
    \frac{\partial \msub{\mathcal{L}}{color-var}}{\partial \alpha_i} = \left\lVert\text{sh}(\bm{\theta}_i, \bm{d})-\igt\right\lVert_2^2 T_i - \frac{1}{1 - \alpha_i} \sum_{j=i+1}^{N-1} w_j  \left\lVert\text{sh}(\bm{\theta}_j, \bm{d})-\igt\right\lVert_2^2 
\end{equation}
For efficiency, $\sum_{j=i+1}^{N-1} w_j(\bm{r})  \left\lVert\text{sh}(\bm{\theta}_j, \bm{d})-\igt\right\lVert_2^2$ is computed on-the-fly during the backward pass, via repeated subtraction from the final variance loss. 
Note that as we build upon SOF~\cite{Radl2025SOF}, we inherit the front-to-back backward pass from StopThePop~\cite{radl2024stopthepop}.
\subsection{Normal Variance Loss}
The gradient derivation for the normal variance loss is mostly analogous, but uses the blended normal $\bm{N} = \sum_{i=0}^{N-1} w_i(\bm{r}) \bm{n}_i$ as mean, since no ground truth is available. By reformulating the loss as the complement of the blended normal magnitude, it can be efficiently computed in the forward pass:
\begin{align}
\msub{\mathcal{L}}{normal-var} 
&= \sum_{i=0}^{N-1} w_i(\bm{r}) \left\lVert\bm{n}_i-\bm{N}\right\lVert_2^2 \\  
&= \sum_{i=0}^{N-1} w_i(\bm{r}) \left( \underbrace{\left\lVert\bm{n}_i\right\lVert_2^2}_{=1} - 2\bm{n}_i \cdot \bm{N} + \left\lVert\bm{N}\right\lVert_2^2\right)\\
&= \underbrace{\sum_{i=0}^{N-1} w_i(\bm{r})}_{=1-T} - 2 \underbrace{\left( \sum_{i=0}^{N-1} w_i(\bm{r}) \bm{n}_i \right) \cdot \bm{N}}_{=\left\lVert\bm{N}\right\lVert_2^2} + \left\lVert\bm{N}\right\lVert_2^2 \underbrace{\sum_{i=0}^{N-1} w_i(\bm{r})}_{=1-T} \\
 &= (1-T_N) - (1 + T_N) \left\lVert\bm{N}\right\lVert_2^2.
\end{align}
The gradients w.r.t. $\bm{n}_i$ and $\alpha_i$ are then analogous to the color variance loss:
\begin{equation}
\frac{\partial \msub{\mathcal{L}}{normal-var}}{\partial n_i} = 2 w_i(\bm{r})  (\bm{n}_i - \bm{N})
\end{equation}
\begin{equation}
    \frac{\partial \msub{\mathcal{L}}{normal-var}}{\partial \alpha_i} = \left\lVert \bm{n}_i - \bm{N}\right\lVert_2^2 T_i - \frac{1}{1 - \alpha_i} \sum_{j=i+1}^{N-1} w_j  \left\lVert \bm{n}_i - 
 \bm{N} \right\lVert_2^2 
\end{equation}
\section{Extended Evaluation}
In this section, we report additional evaluation results, such as geometry reconstruction for small, bounded scenes~\cite{jensen2014large} and Novel View Synthesis results for the Mip-NeRF 360 dataset~\cite{barron2022mipnerf360}.

\subsection{Comparison with prior-dependent Methods}
\label{app:geosvr}
Here, we compare our method to the prior-dependent GeoSVR~\cite{li2025geosvr}, which builds on the recent SVRaster~\cite{Sun2024SVRaster}.
In particular, this method utilizes multi-view constraints adapted from PGSR~\cite{chen2024pgsr}, and relies on a monocular depth estimator.
Additionally, GeoSVR is not able to extract unbounded meshes due to its reliance on TSDF fusion (using the same evaluation protocol as PGSR).
We reproduced their results on our unified data and present the comparison in \cref{tab:geosvr}.
%
\begin{table}[]
    \centering
    \setlength{\tabcolsep}{4pt}
    \caption{
\textbf{Comparison with prior-dependent GeoSVR~\cite{li2025geosvr}} using the Tanks \& Temples dataset~\cite{Knapitsch2017tanks}.
Our method is able to extract unbounded meshes while keeping processing time low, whereas GeoSVR attains a higher F1-score.
}
    \label{tab:geosvr}
    \resizebox{.98\linewidth}{!}{
\begin{tabular}{lrrrrrrr@{\hskip 0.5cm} l}
\toprule
 & Barn & Caterpillar & Courthouse & Ignatius & Meetingroom & Truck & Average & Runtime \\
\midrule
Ours & 0.534 & \cellcolor{tab_color!32} 0.472 & \cellcolor{tab_color!32} 0.333 & \cellcolor{tab_color!15} 0.782 & 0.372 & 0.634 & \cellcolor{tab_color!15} 0.521 & 18min\\
GeoSVR & \cellcolor{tab_color!49} 0.557 & \cellcolor{tab_color!49} 0.495 & \cellcolor{tab_color!49} 0.347 & \cellcolor{tab_color!49} 0.804 & \cellcolor{tab_color!32} 0.382 & \cellcolor{tab_color!49} 0.688 & \cellcolor{tab_color!49} 0.546 & 42min\\
GeoSVR (w/ QGS Eval) & \cellcolor{tab_color!15} 0.544 & 0.432 & \cellcolor{tab_color!15} 0.328 & \cellcolor{tab_color!32} 0.787 & \cellcolor{tab_color!49} 0.408 & \cellcolor{tab_color!15} 0.651 & \cellcolor{tab_color!32} 0.525 & 42min\\
\bottomrule
\end{tabular}
}
\end{table}
%
Although GeoSVR achieves a superior F1-score compared to our method, this does not come without trade-offs:
Mainly, the reliance on multi-view techniques and a monocular depth prior is costly in terms of performance, more than doubling optimization time ($42$ vs. $18$ minutes);
Note that this explicitly excludes the additional preprocessing overhead required to obtain depth maps from the monocular model.
Because GeoSVR relies on TSDF fusion, it is unable to produce unbounded meshes, leading to lower overall mesh completeness.

Following PGSR, GeoSVR also uses ground-truth point clouds to derive an optimal, scene-specific voxel size for TSDF fusion; 
in practical scenarios, such privileged information about scene scale is strictly unavailable, necessitating a costly, per-scene hyperparameter search.
To evaluate GeoSVR under realistic conditions, we re-extracted their meshes using the TSDF fusion protocol from QGS~\cite{zhang2025qgs}, which employs a fixed, scene-agnostic voxel size of $0.006$ for all scenes.
Under this evaluation setting, GeoSVR's average F1-score drops significantly to $0.525$.

\subsection{Extended Comparison with Bounded Methods}
We also compare our results against PGSR~\cite{chen2024pgsr}, which serves as a benchmark for bounded reconstruction methods. 
These techniques typically rely on TSDF fusion, where performance is heavily dictated by voxel size.

Because averaging over multiple observations is an inherent part of TSDF fusion, the resulting meshes tend to be overly smooth.
And as the input is restricted to depth renderings from the training views, rarely observed regions are often missing from the final mesh.
These gaps in the reconstruction sometimes even aid the final F1-score for the Tanks \& Temples dataset~\cite{Knapitsch2017tanks}, as ground-truth point clouds are missing important fine details (\cf \cref{fig:completeness} for a visualization);
conversely, this negatively impacts the more complete meshes of unbounded methods quantitatively.
\newcommand{\tntanno}[2]{%
    \stackinset{l}{1pt}{t}{1pt}{%
        \begin{tikzpicture}
            \node[fill=black, fill opacity=0.5, text opacity=1, 
                  inner sep=1.5pt, rounded corners=0.5pt] 
            {\sffamily\fontsize{5}{4}\selectfont{\textcolor{white}{#2}}};
        \end{tikzpicture}%
    }{#1}%
}

\newcommand{\boxedtntanno}[2]{%
    \tntanno{%
        \begin{tikzpicture}
            \node[anchor=south west, inner sep=0] (img) at (0,0) {#1};
            \begin{scope}[x={(img.south east)}, y={(img.north west)}]
                \draw[white, thin] (\xmin, \ymin) rectangle (\xmax, \ymax);
            \end{scope}
        \end{tikzpicture}%
    }{#2}%
}

\begin{figure}
    \centering
    \scriptsize\sffamily
    \begin{minipage}{0.32\textwidth}
        \centering \small Ground Truth Mesh
    \end{minipage}\hfill
    \begin{minipage}{0.32\textwidth}
        \centering \small Precision: Ours
    \end{minipage}\hfill
    \begin{minipage}{0.32\textwidth}
        \centering \small Precision: PGSR
    \end{minipage}
    \vspace{2pt}
    
    \def\xmin{0.05} \def\ymin{0.1} \def\xmax{0.5} \def\ymax{0.5}%
    \boxedtntanno{\includegraphics[width=0.32\linewidth]{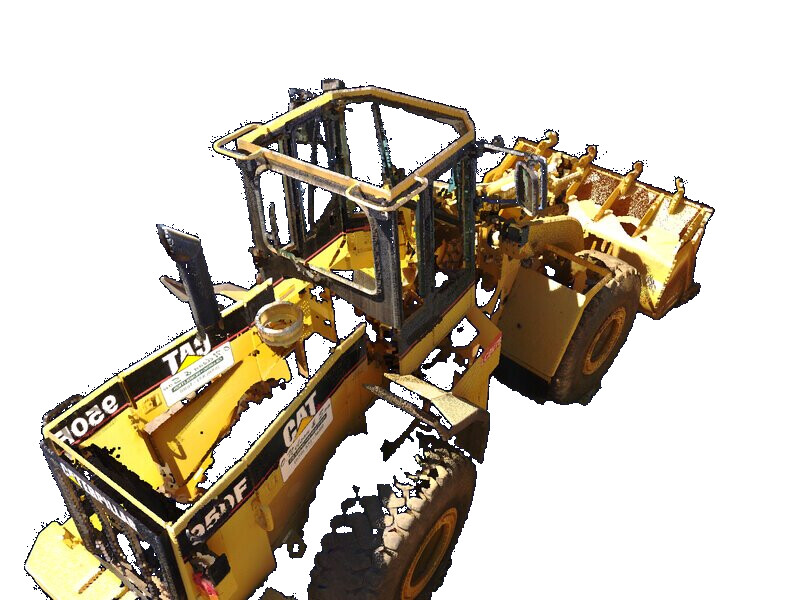}}{Caterpillar}
    \boxedtntanno{\includegraphics[width=0.32\linewidth]{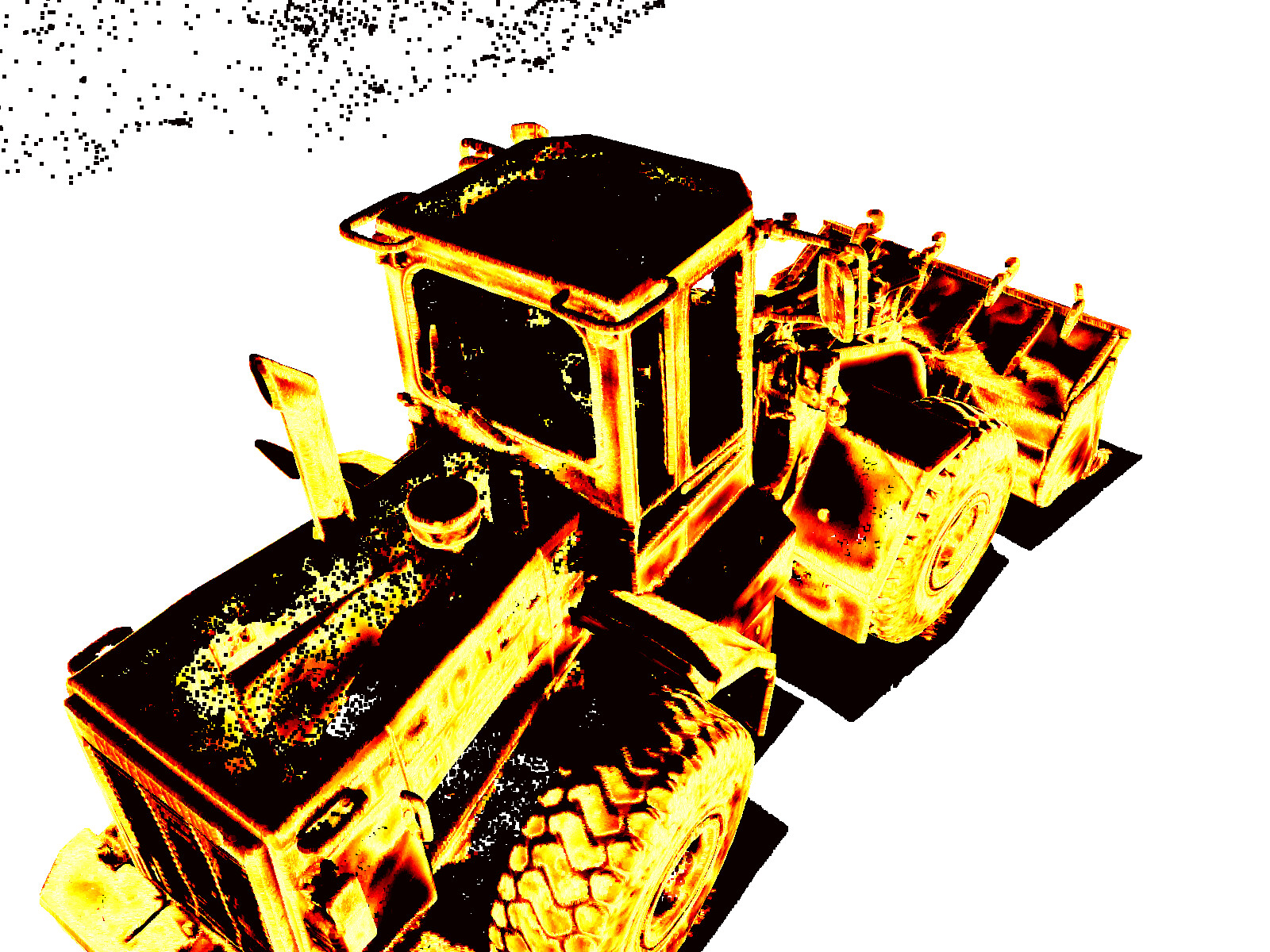}}{F1-Score: 0.472}
    \boxedtntanno{\includegraphics[width=0.32\linewidth]{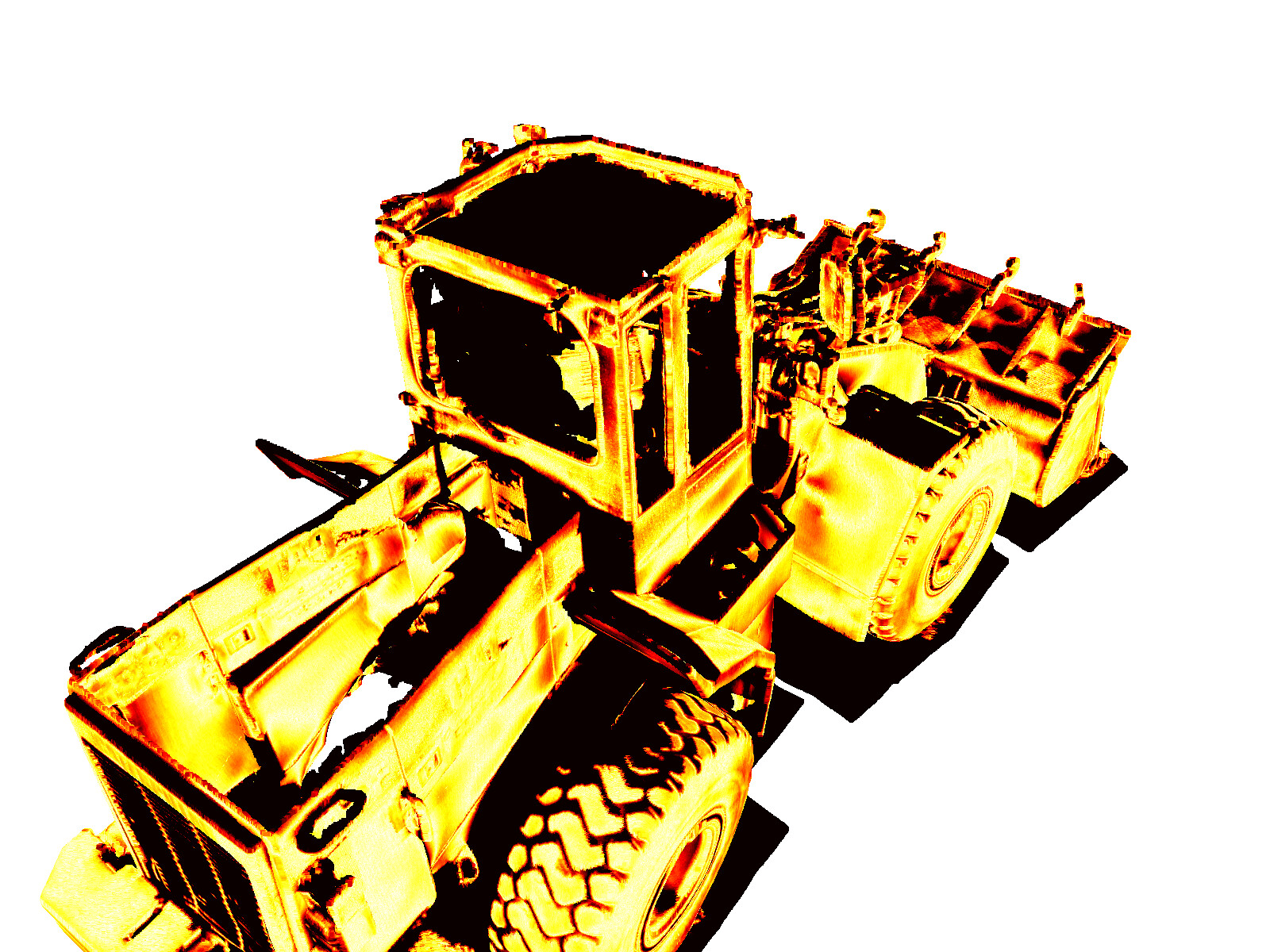}}{F1-Score: 0.437} \\
    
    \def\xmin{0.05} \def\ymin{0.1} \def\xmax{0.4} \def\ymax{0.6}%
    \boxedtntanno{\includegraphics[width=0.32\linewidth]{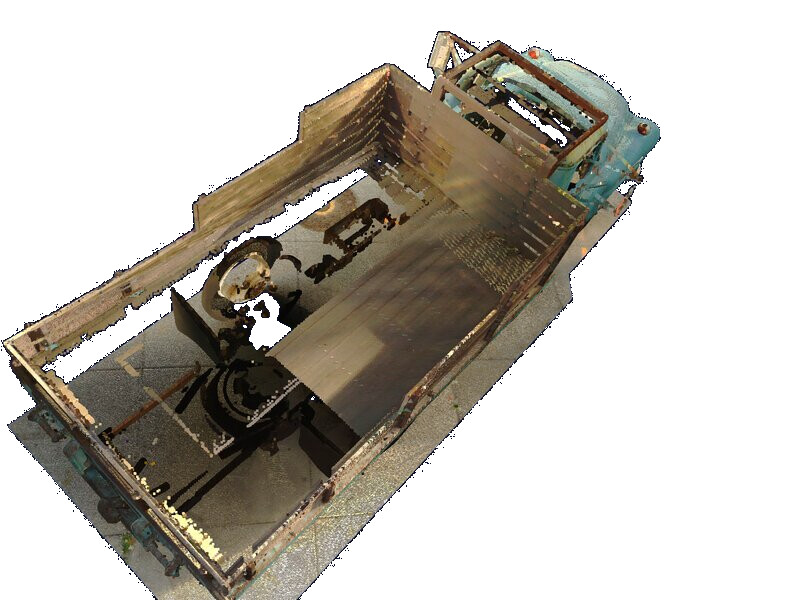}}{Truck}
    \boxedtntanno{\includegraphics[width=0.32\linewidth]{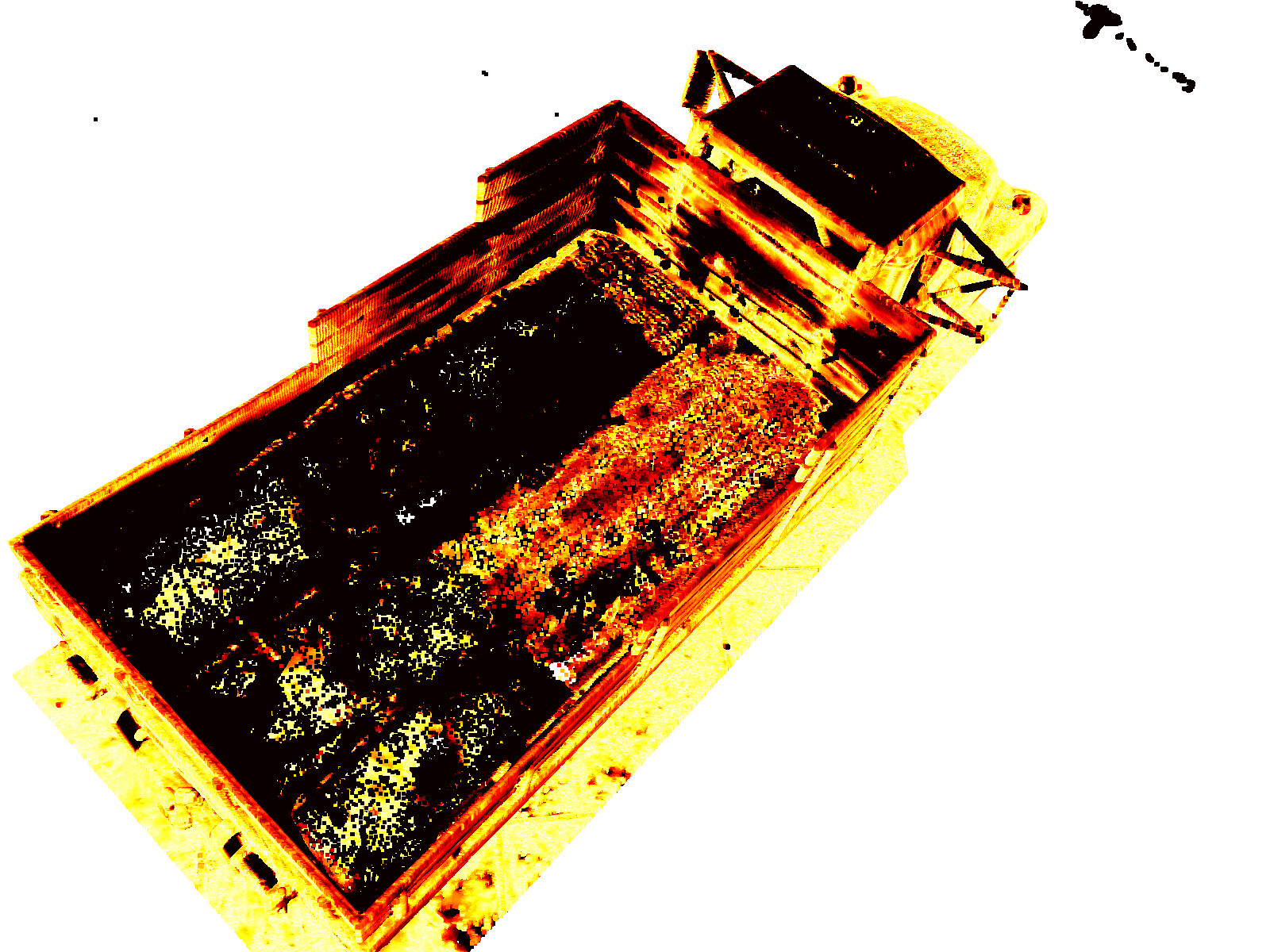}}{F1-Score: 0.634}
    \boxedtntanno{\includegraphics[width=0.32\linewidth]{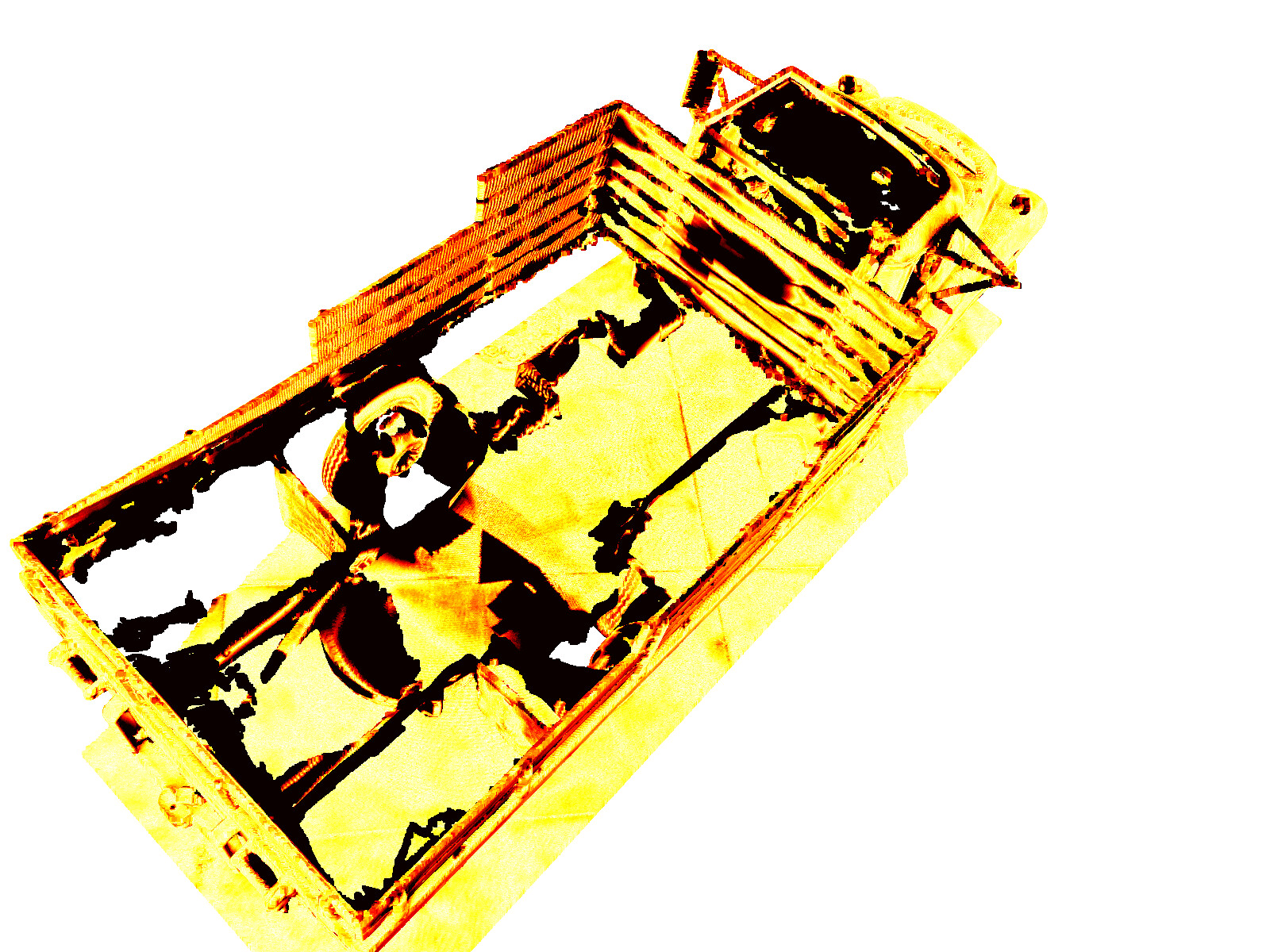}}{F1-Score: 0.658}
    
    \caption{
\textbf{Completeness Comparison:}
Due to gaps in the ground-truth point clouds, our more complete meshes are penalized in precision, compared to the bounded meshes from PGSR~\cite{chen2024pgsr} (black points denote low precision). 
This highlights the fundamental disadvantage unbounded methods face for this benchmark.
}
\label{fig:completeness}
\end{figure}

\begin{figure*}[p] 
    \centering
    \scriptsize\sffamily 
    
    \newcommand{\cropimg}[1]{\includegraphics[width=0.28\linewidth]{#1}}
    \newcommand{\cropimgs}[1]{\includegraphics[width=0.28\linewidth]{#1}}
    
    \begin{minipage}{0.28\linewidth}\centering \textbf{Render (3DGS)}\end{minipage}
    \begin{minipage}{0.28\linewidth}\centering \textbf{PGSR}\end{minipage}
    \begin{minipage}{0.28\linewidth}\centering \textbf{Ours}\end{minipage}
    \vspace{1pt}

    {{Scene 1: Caterpillar}\par}\vspace{2pt}
    \cropimg{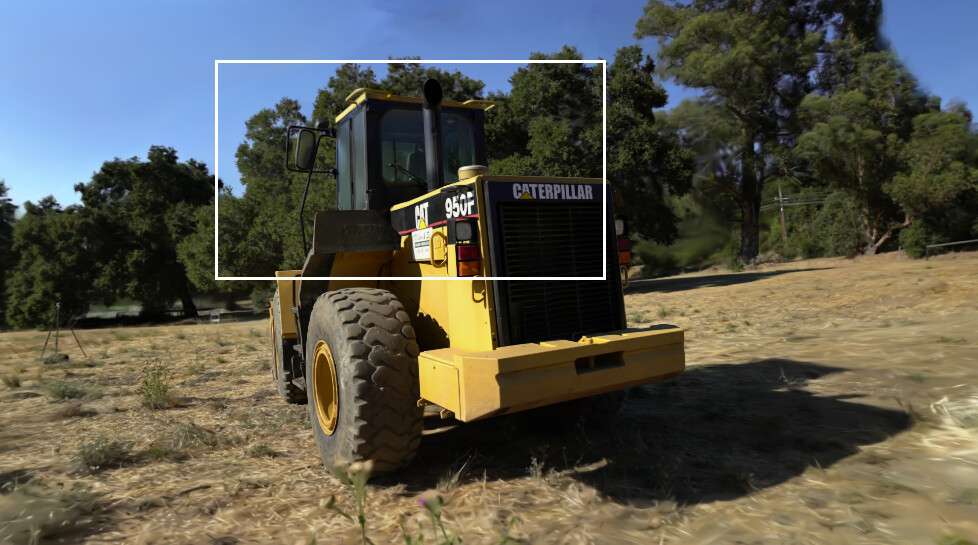}
    \cropimg{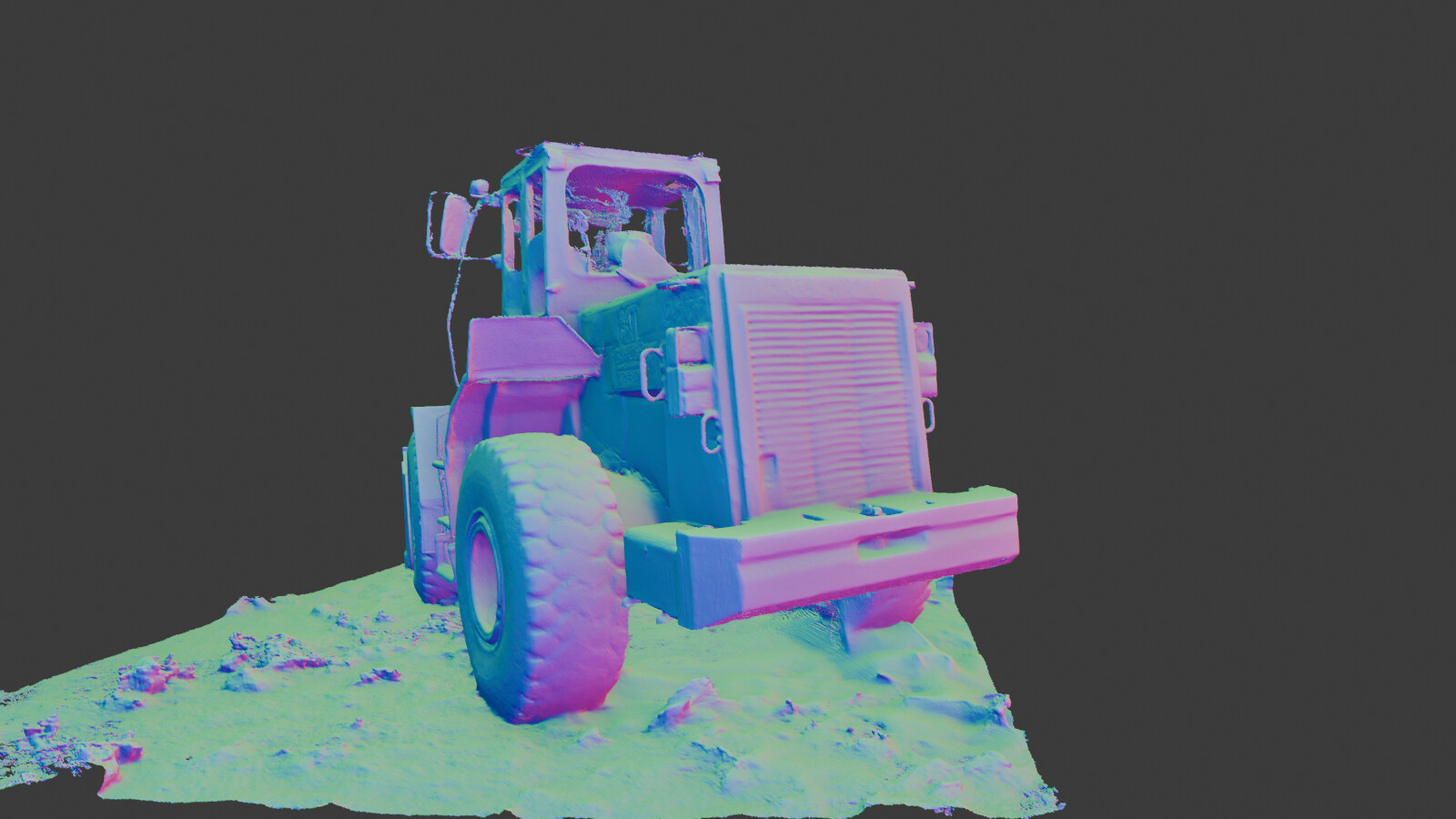}
    \cropimg{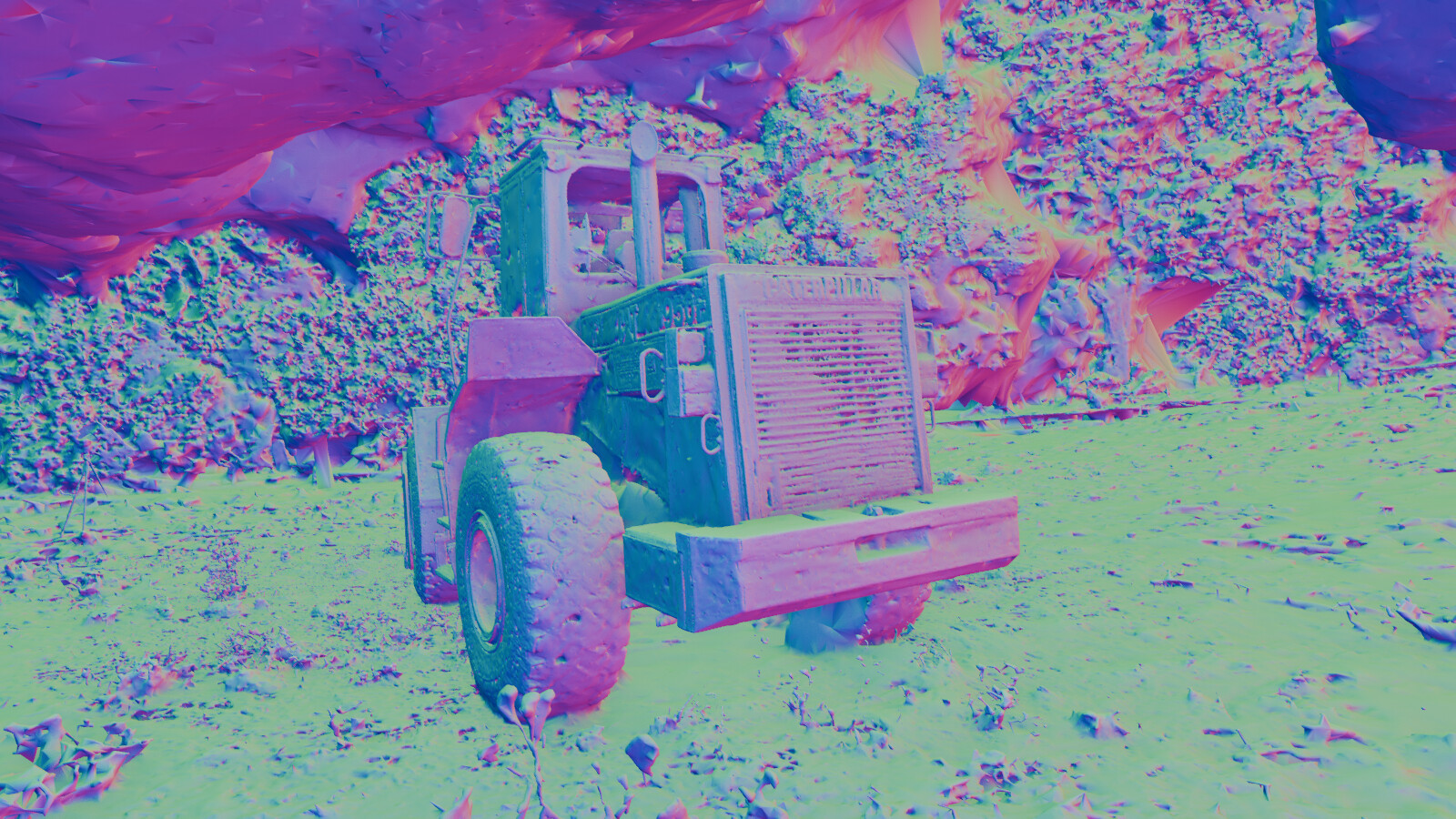}
    \\
    \cropimgs{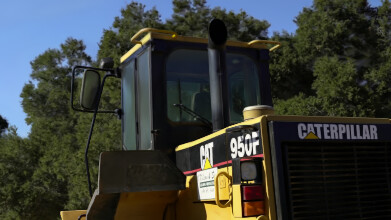}
    \cropimgs{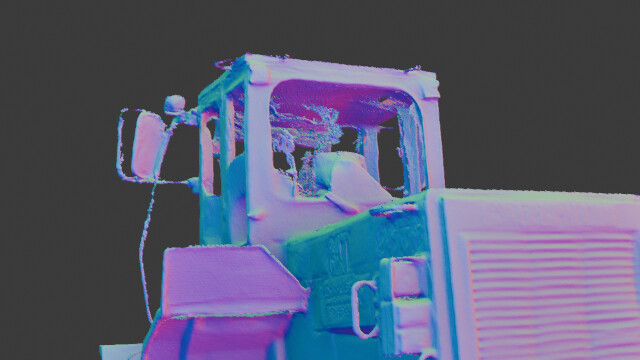}
    \cropimgs{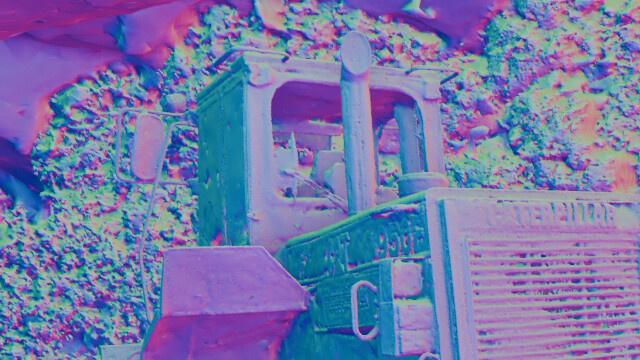}
    
    \vspace{1pt}

    {\centering{Scene 2: Meetingroom}\par}\vspace{2pt}
    \cropimg{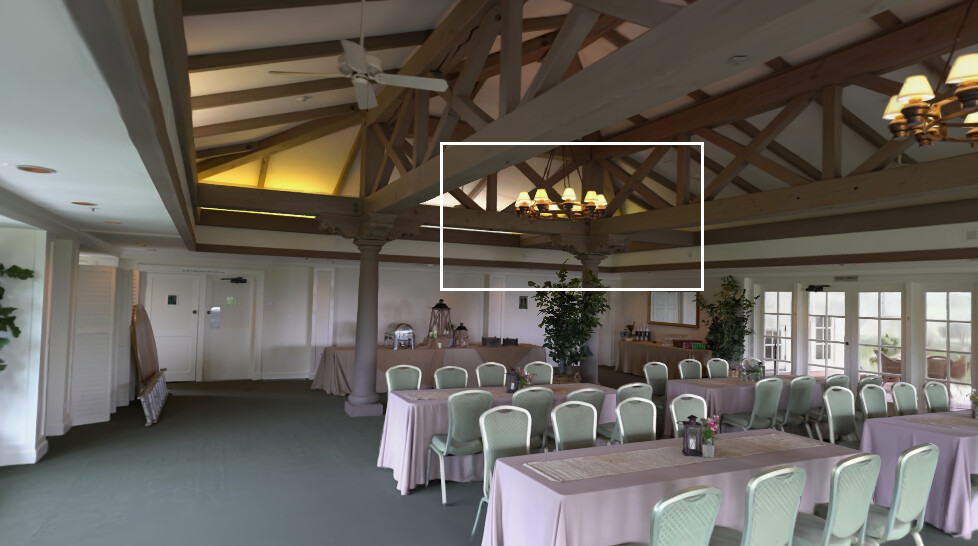}
    \cropimg{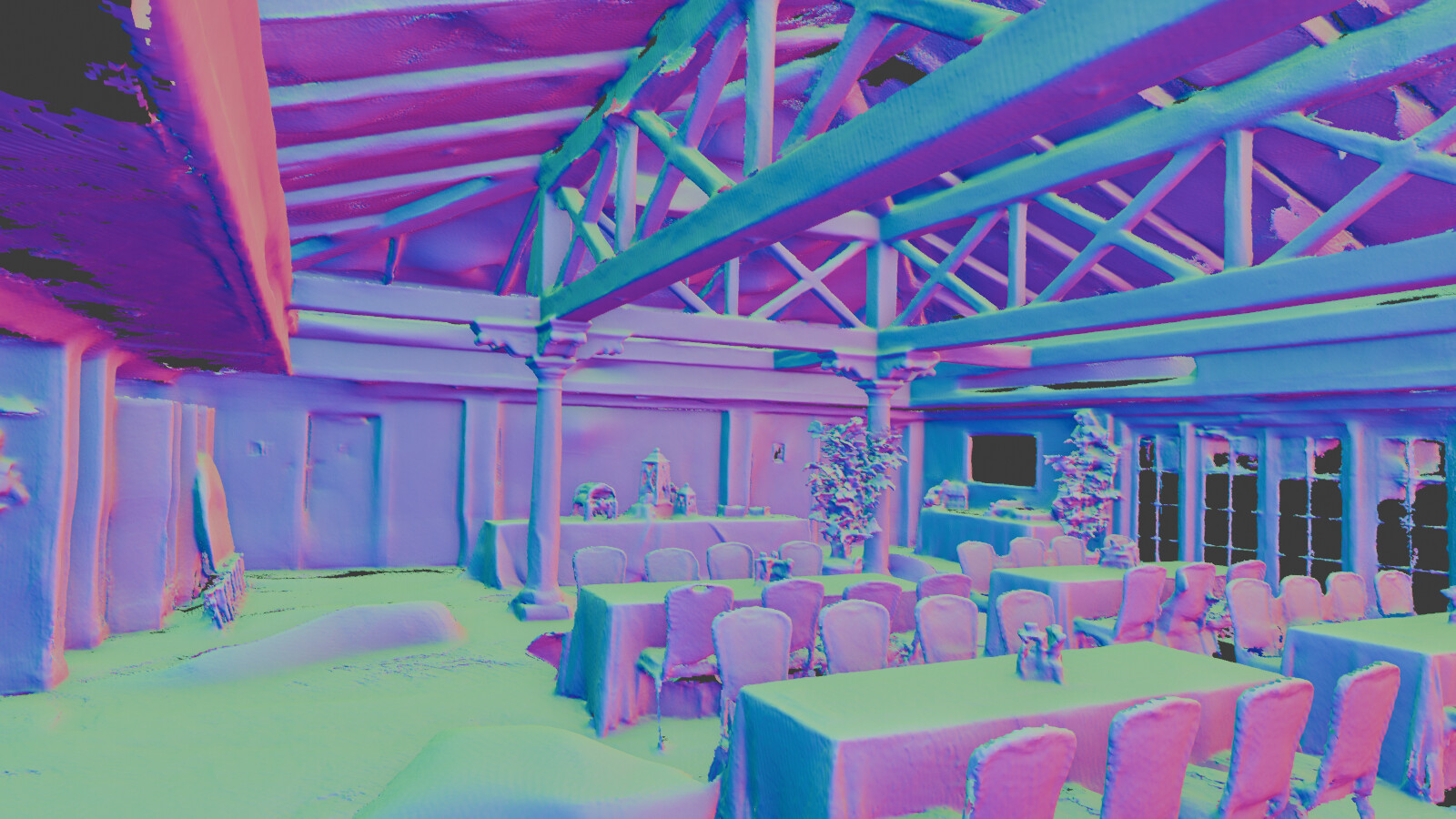}
    \cropimg{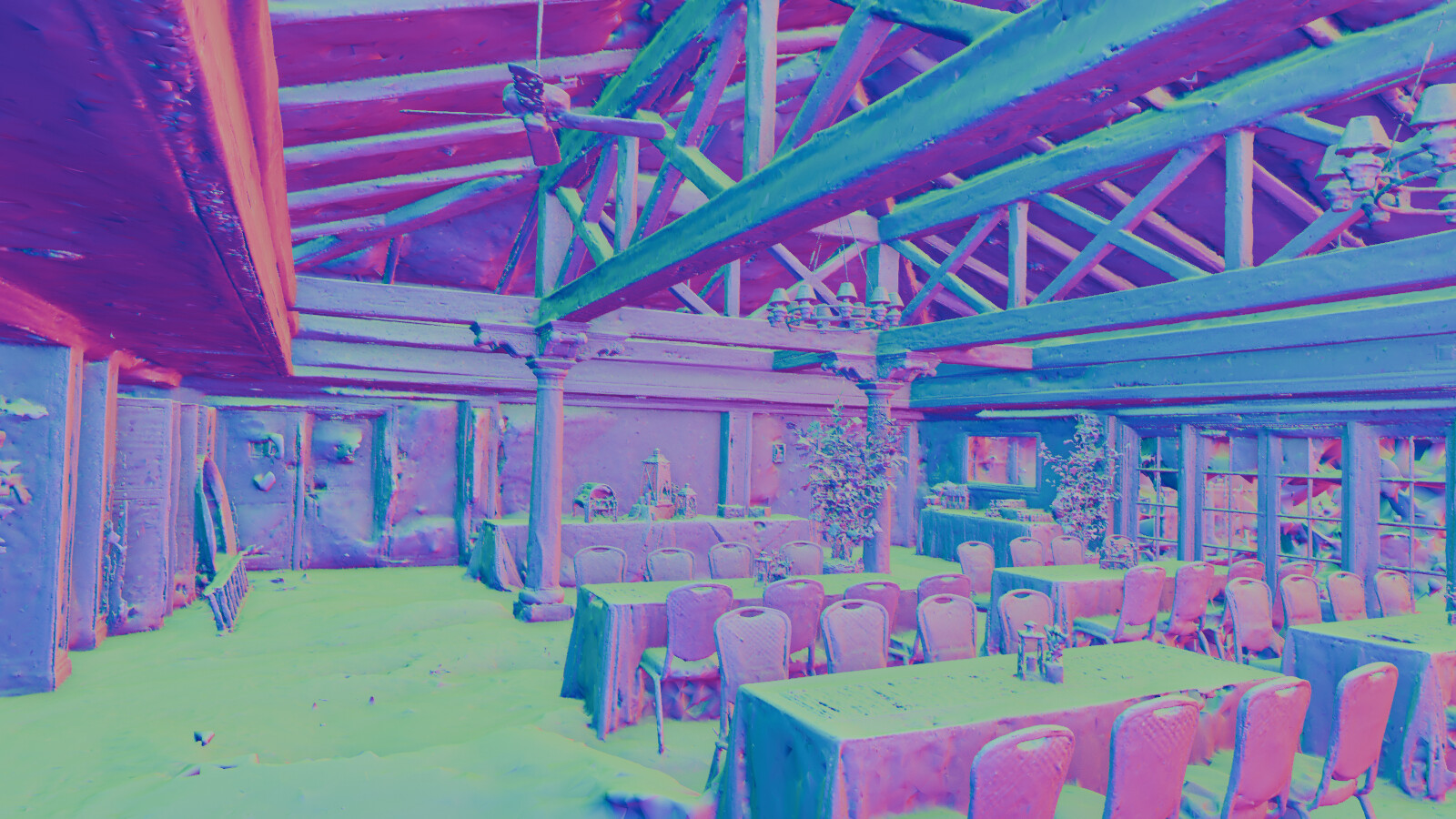}
    \\
    \cropimgs{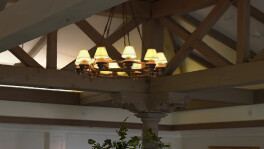}
    \cropimgs{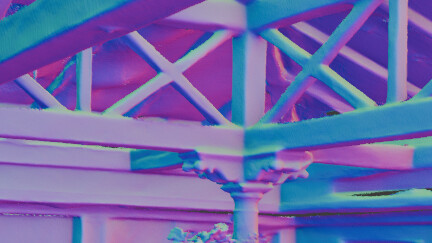}
    \cropimgs{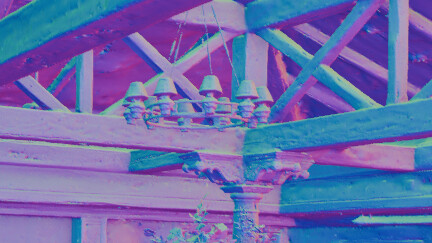}
    
    \vspace{1pt}

    {\centering{Scene 3: Courthouse}\par}\vspace{2pt}
    \cropimg{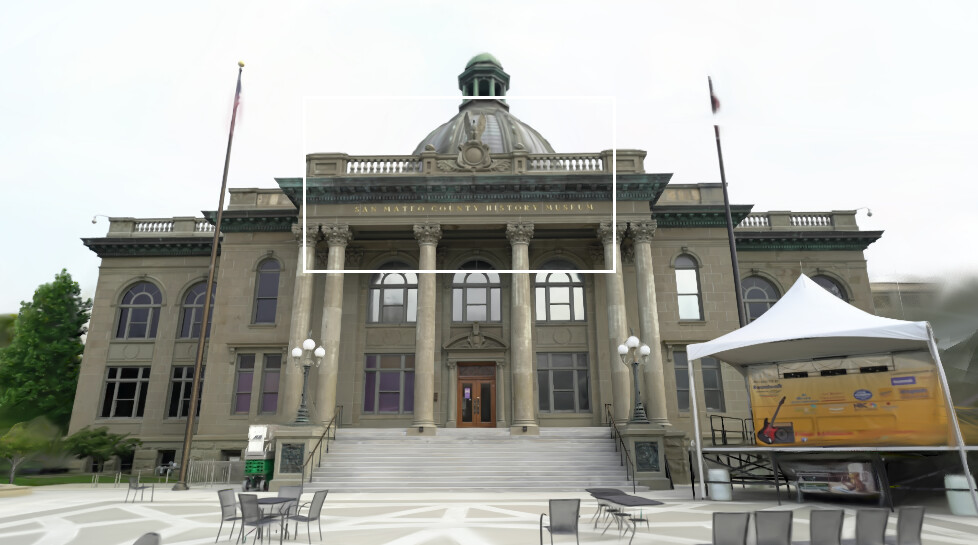}
    \cropimg{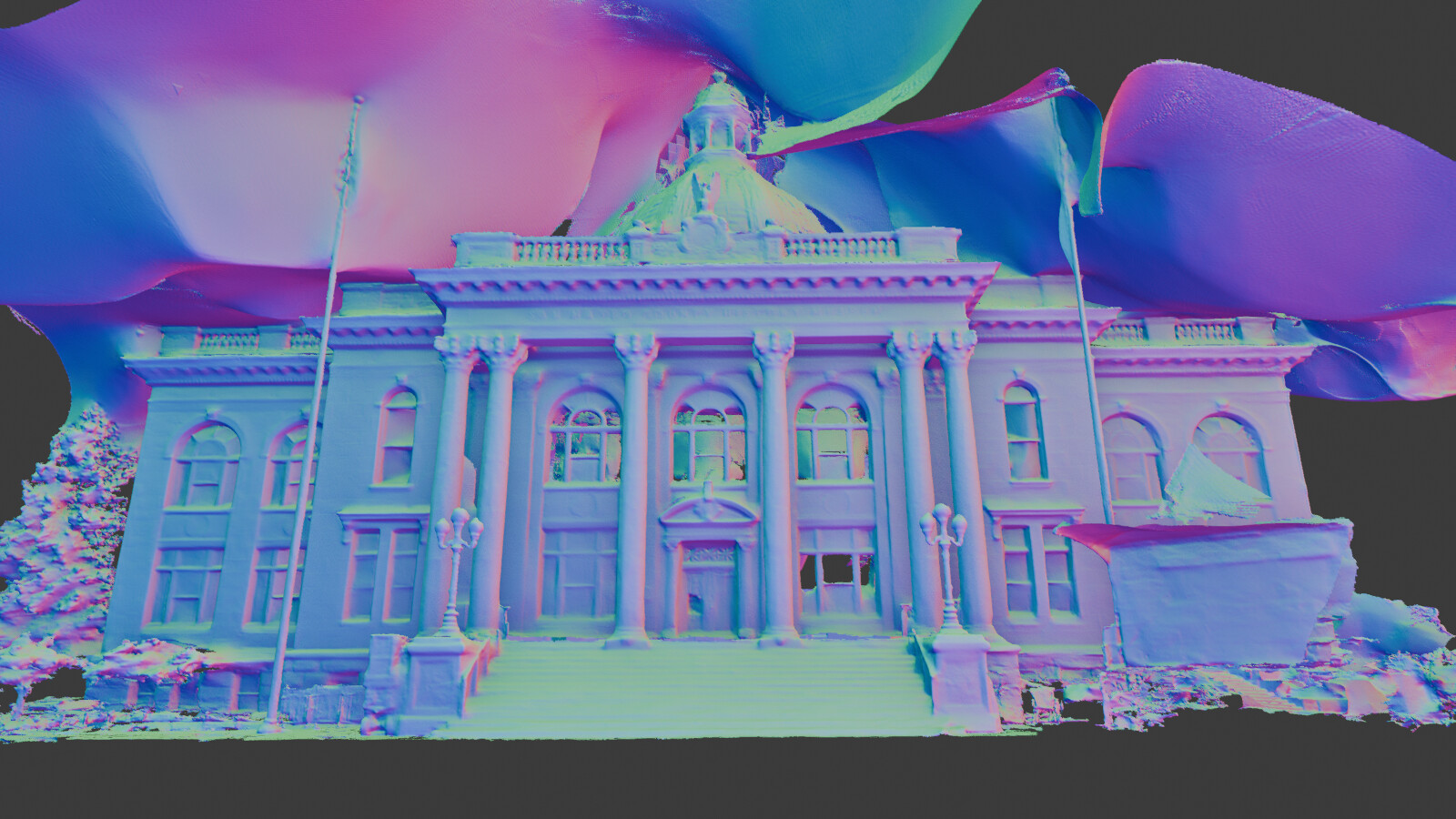}
    \cropimg{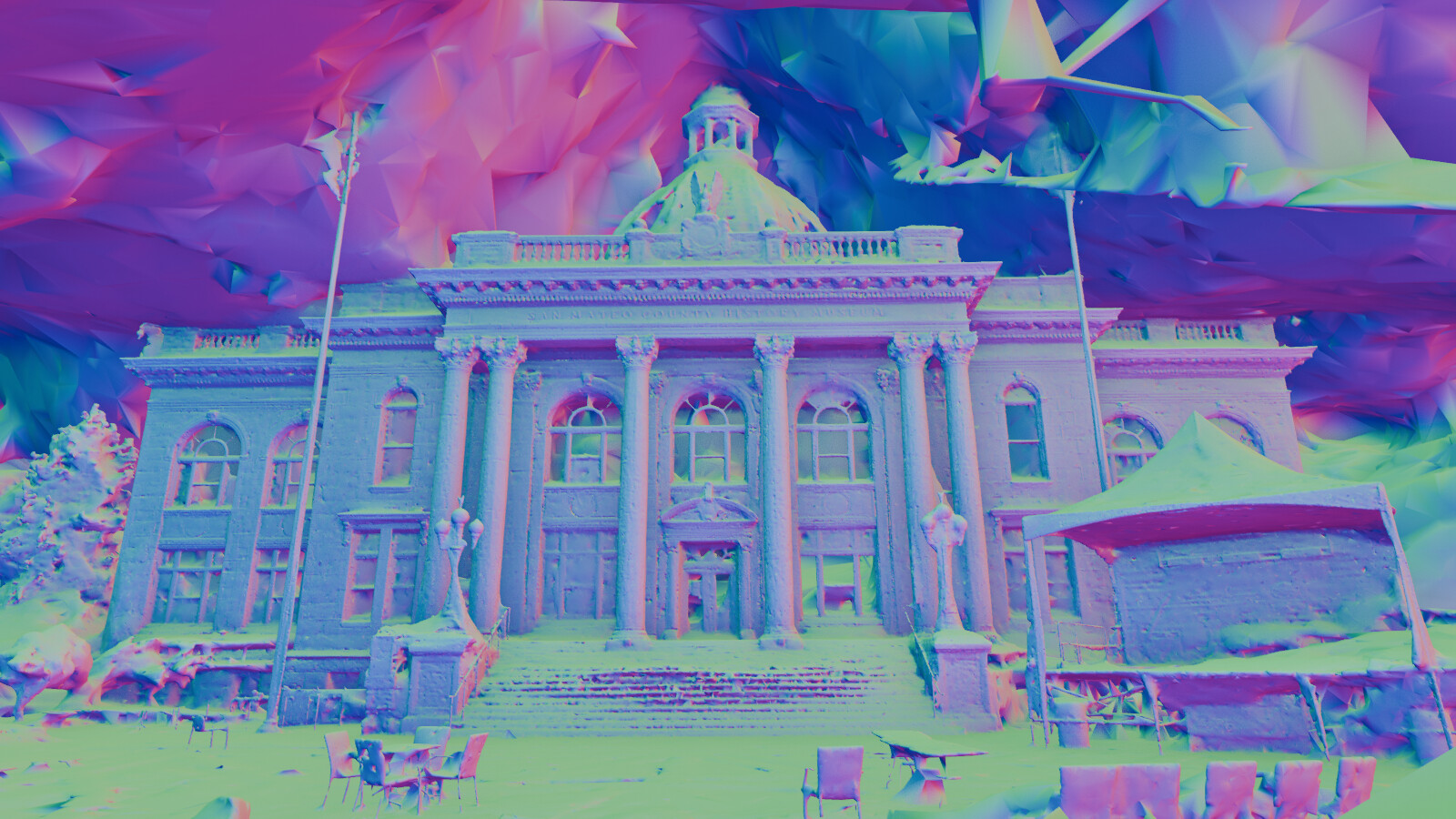}
    \\
    \cropimgs{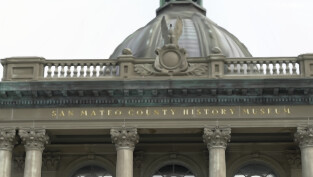}
    \cropimgs{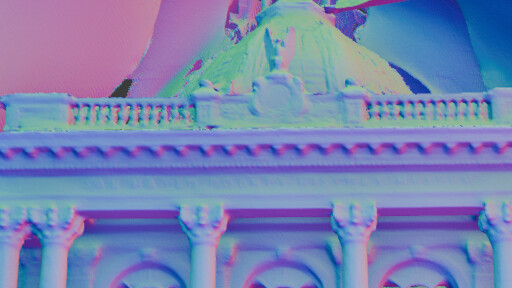}
    \cropimgs{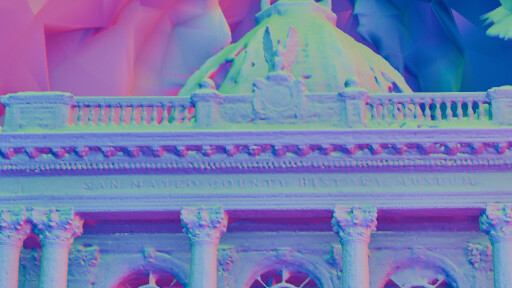}
    
    \vspace{1pt}
    
    {\centering{Scene 4: Barn}\par}\vspace{2pt}
    \cropimg{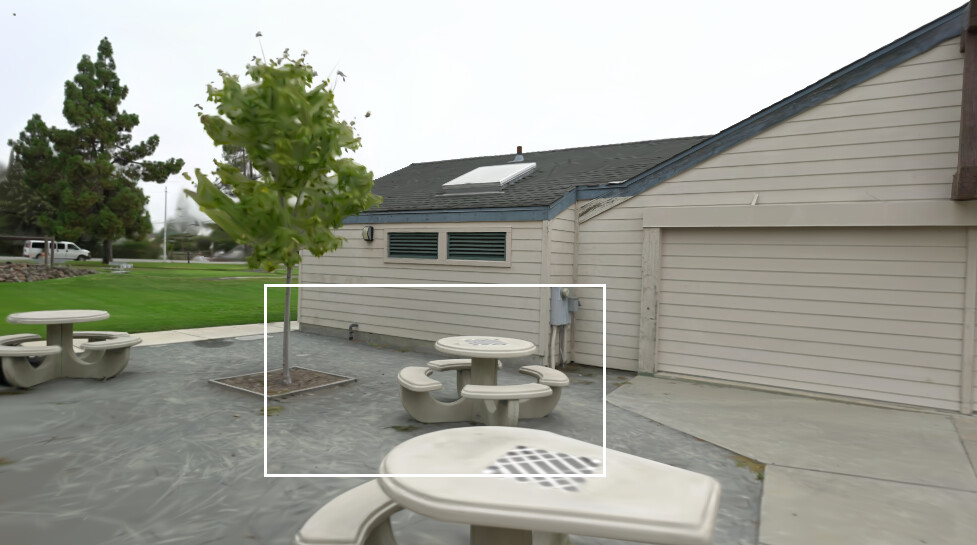}
    \cropimg{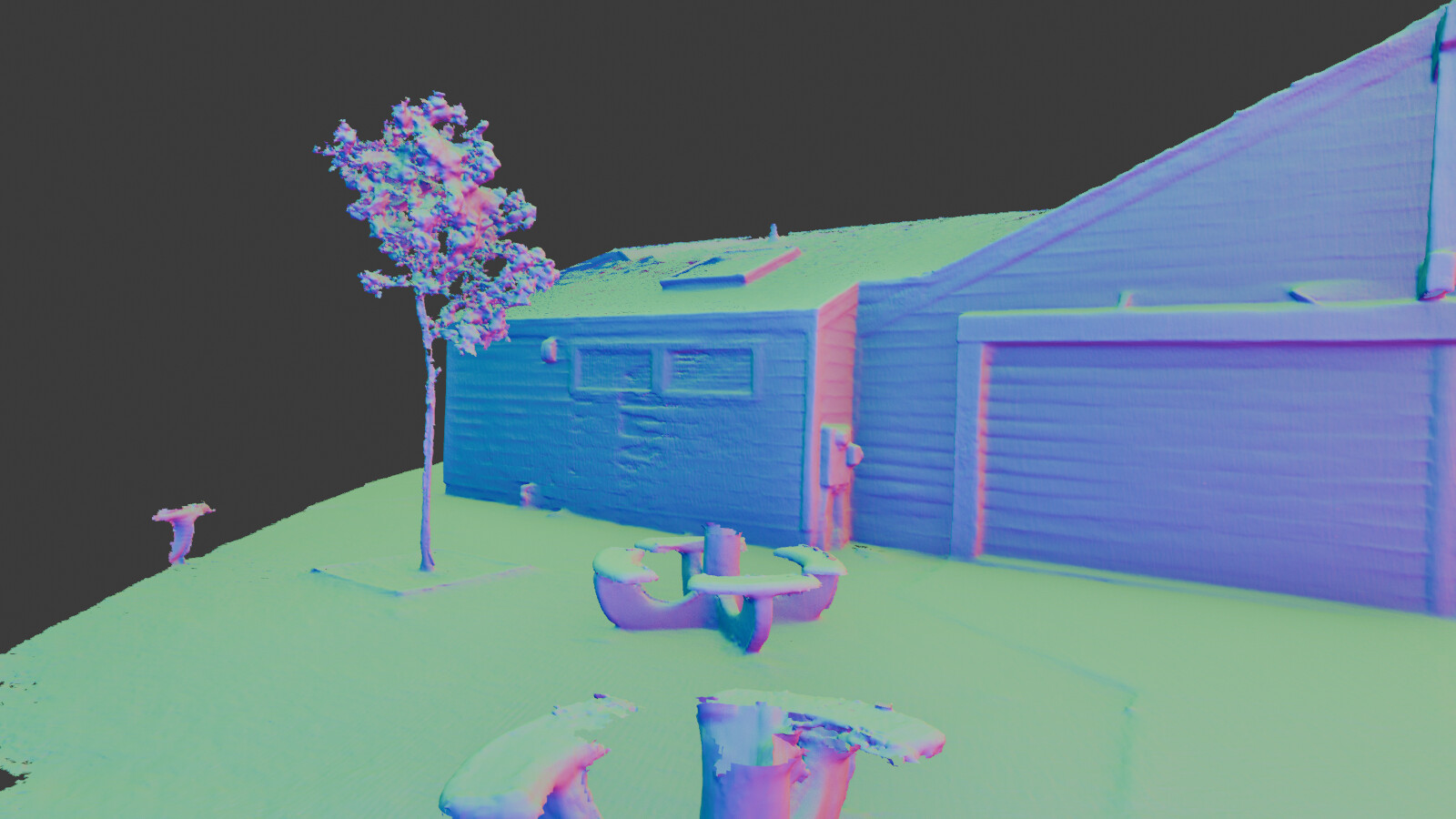}
    \cropimg{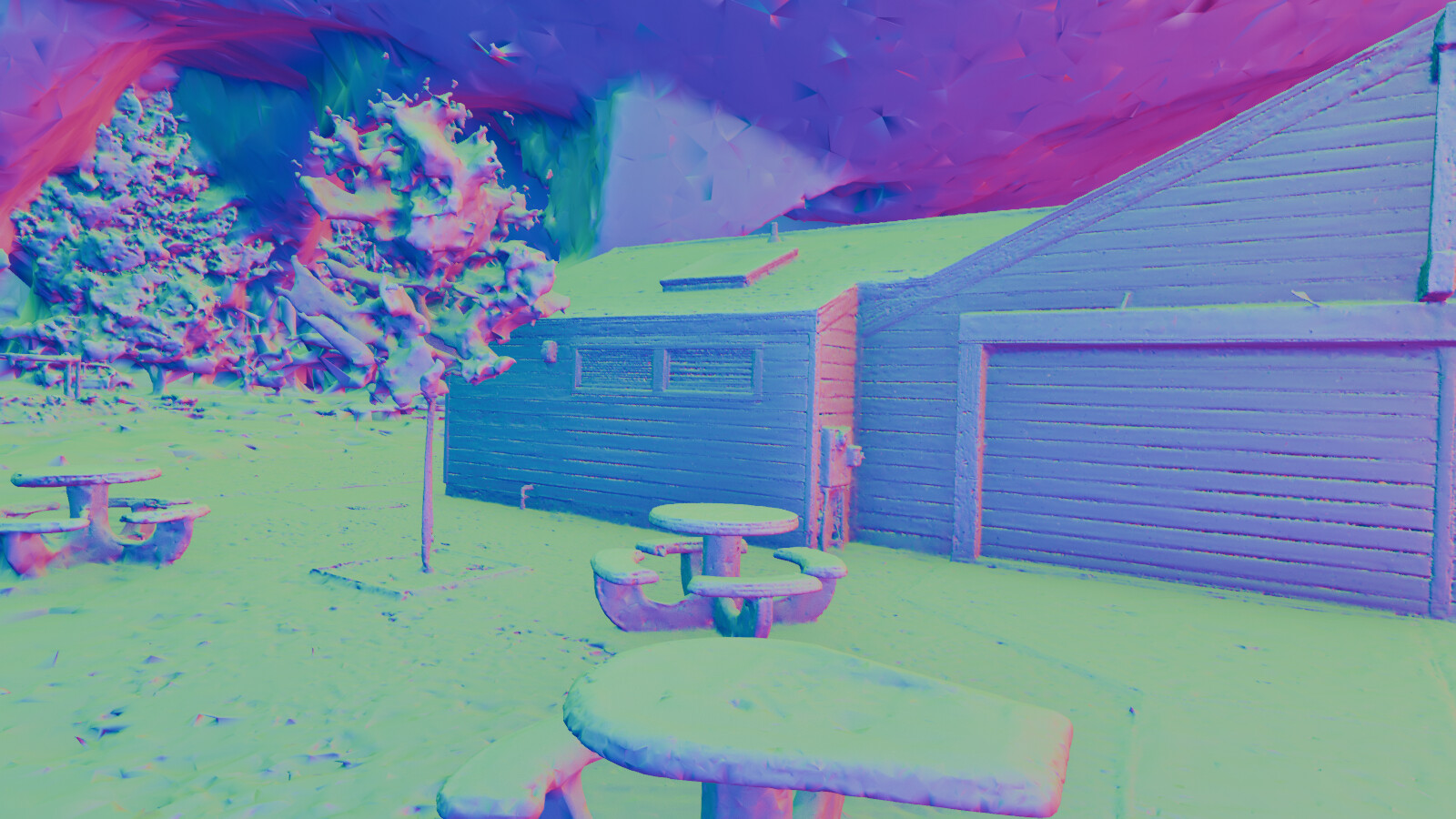}
    \\
    \cropimgs{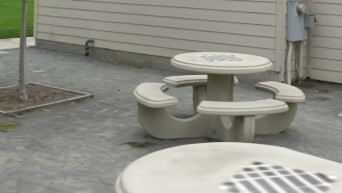}
    \cropimgs{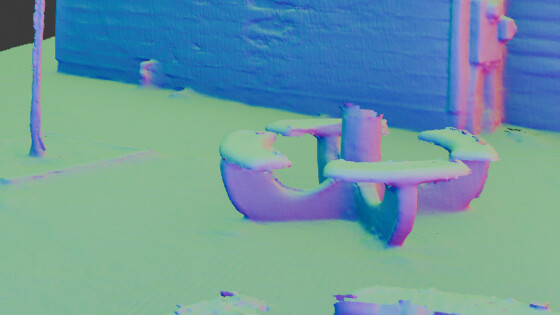}
    \cropimgs{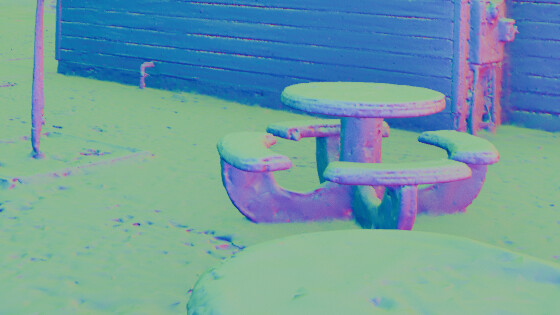}

    \caption{
\textbf{Novel View Mesh Rendering Comparison with Bounded Methods.} 
Our meshes are significantly more complete and detailed compared to bounded methods, such as PGSR~\cite{chen2024pgsr}.
}
    \label{fig:pgsr}
\end{figure*}

We provide additional comparisons with PGSR in \cref{fig:pgsr};
as we can see, our meshes are far more detailed and more complete.
Importantly, our method is also faster than PGSR, while outperforming it quantitatively across multiple datasets.

\subsection{Extended Comparison with Unbounded Methods}
Here, we draw a closer comparison to the current state-of-the-art method \milo~\cite{guedon2025milo}, which also uses a variant of Marching Tetrahedra for mesh extraction. 
Therefore, mesh extraction times are similar with 8.6 minutes for \milo and 6.6 minutes for our method, while optimization is $3\times$ slower for \milo.

Although \milo utilizes fewer primitives than our method (343K on average), our approach still achieves faster optimization.
Note that we also tested the \emph{dense} variant of \milo;
we did not see an increase in either detail or F1-score, while optimization time increased further.
Our light variant ($\beta = 0.05$) also achieves a superior F1-score of $\mathbf{0.504}$ (an improvement of $\mathbf{0.019}$ compared to \milo) with 705K primitives on average.
With an optimization time of 16 minutes, this is the fastest of all our tested methods.

\begin{figure*}[p] 
    \centering
    \scriptsize\sffamily 
    
    \newcommand{\cropimg}[1]{\includegraphics[width=0.28\linewidth]{#1}}
    \newcommand{\cropimgs}[1]{\includegraphics[width=0.28\linewidth]{#1}}
    
    \begin{minipage}{0.28\linewidth}\centering \textbf{Render (3DGS)}\end{minipage}
    \begin{minipage}{0.28\linewidth}\centering \textbf{\milo}\end{minipage}
    \begin{minipage}{0.28\linewidth}\centering \textbf{Ours}\end{minipage}
    \vspace{1pt}

    {{Scene 1: Ignatius}\par}\vspace{2pt}
    \cropimg{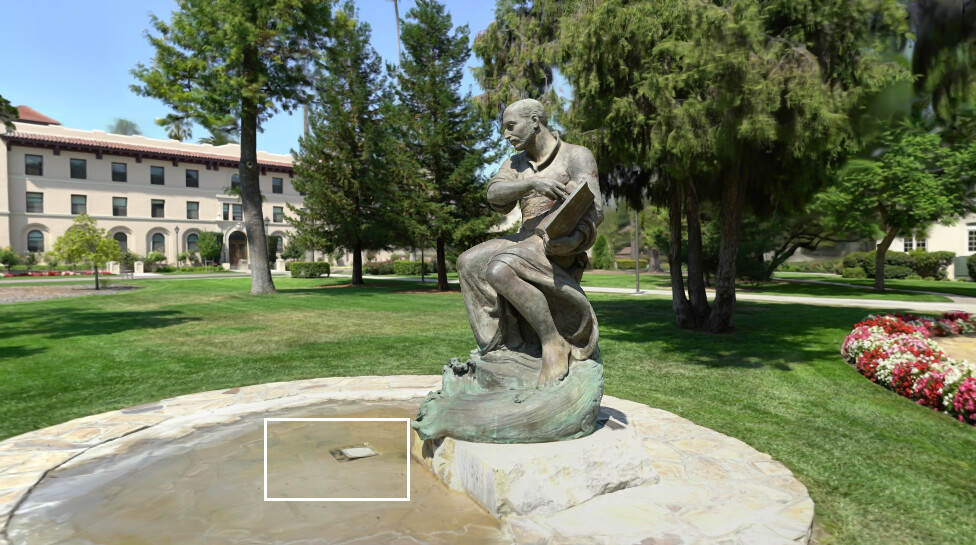}
    \cropimg{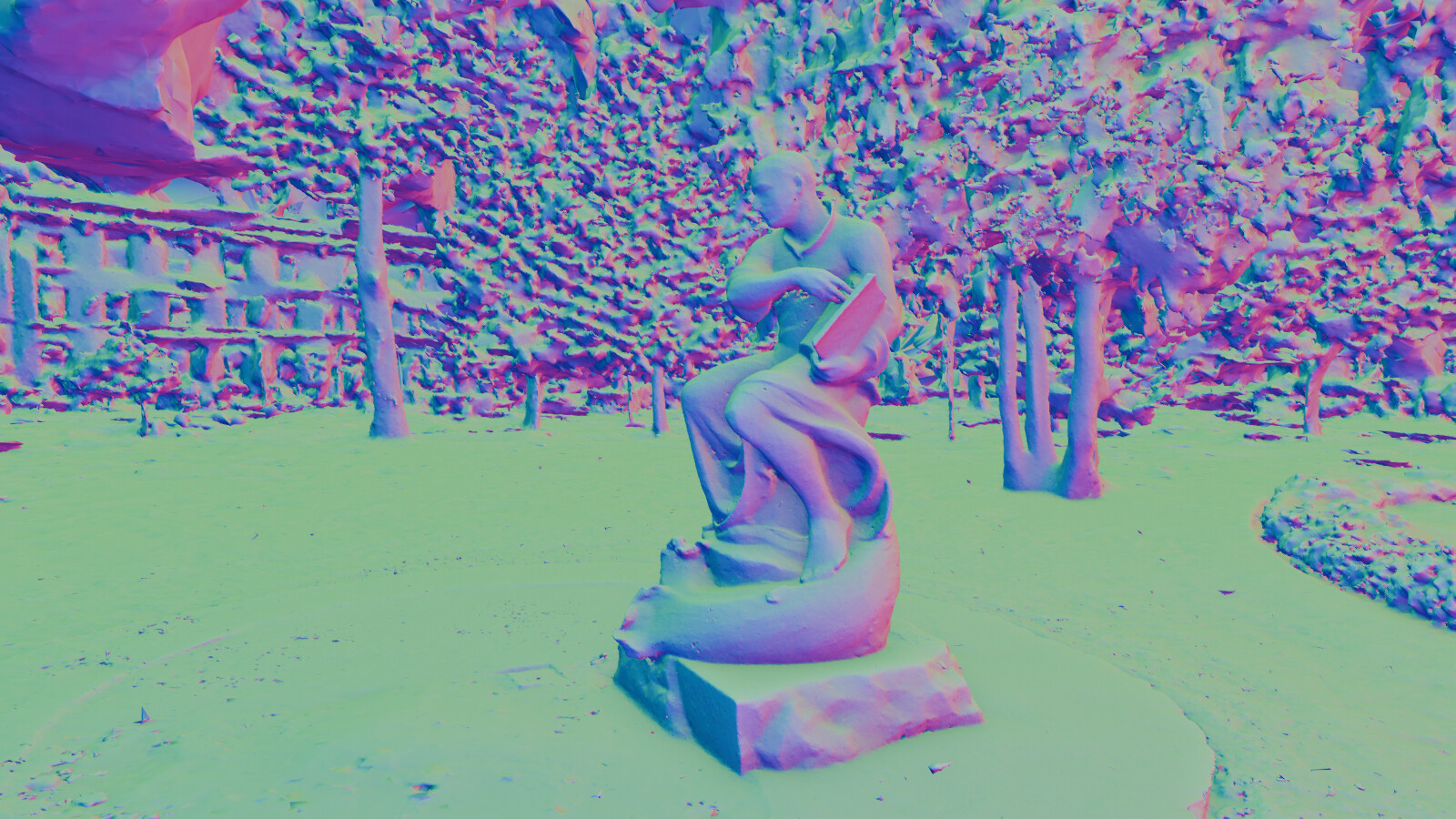}
    \cropimg{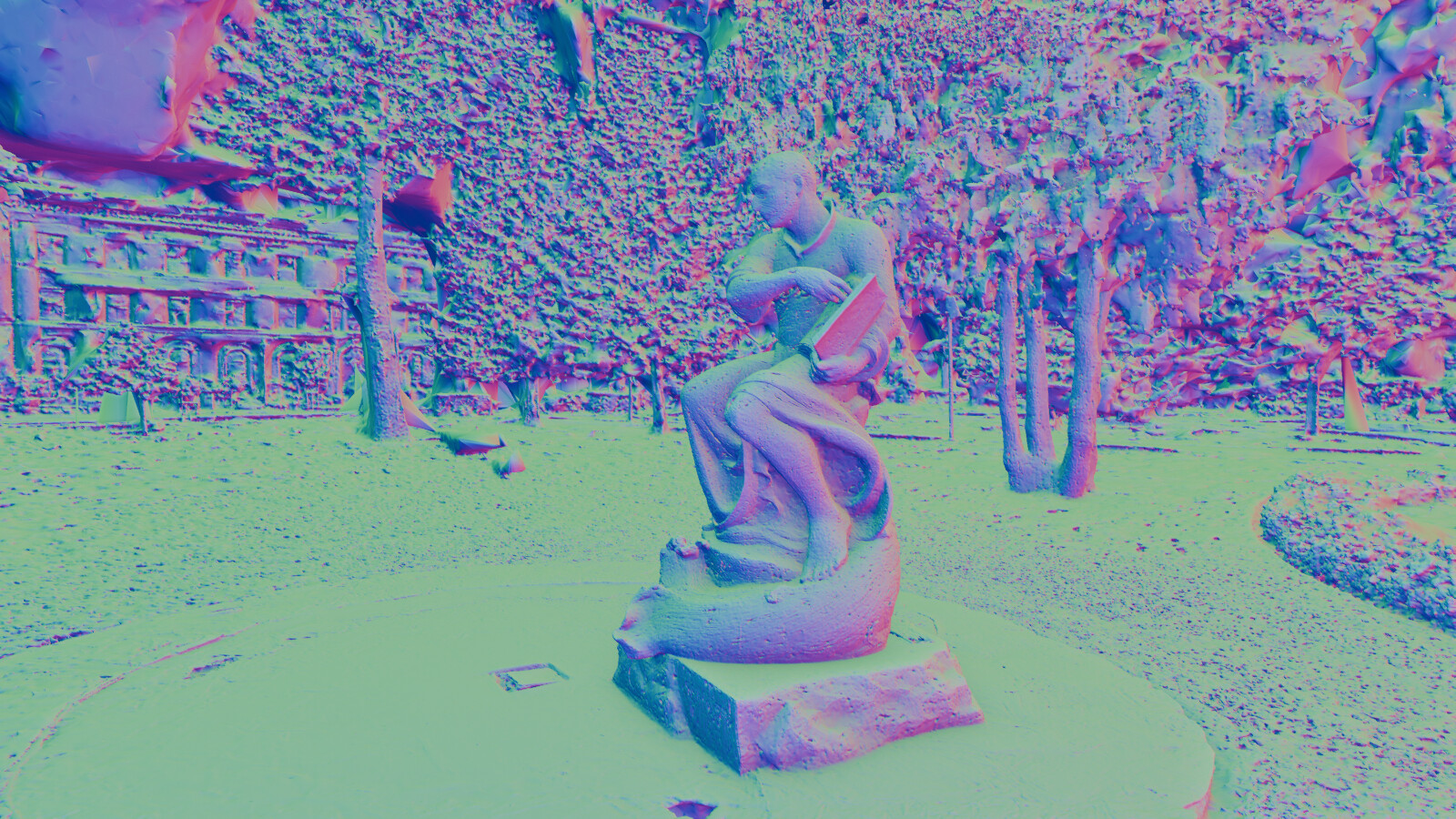}
    \\
    \cropimgs{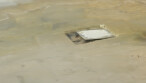}
    \cropimgs{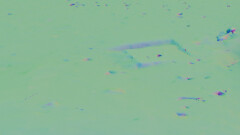}
    \cropimgs{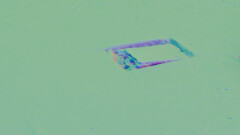}
    
    \vspace{1pt}

    {\centering{Scene 2: Courthouse}\par}\vspace{2pt}
    \cropimg{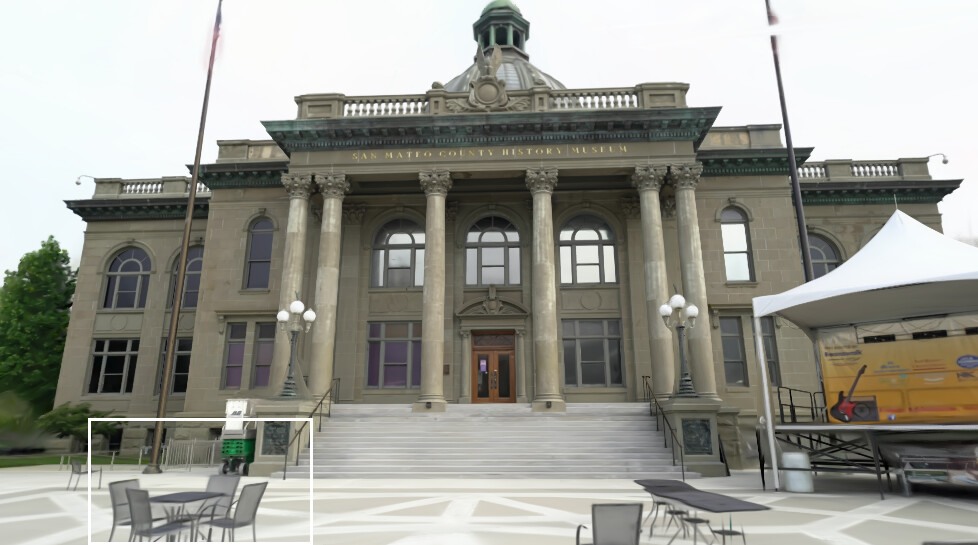}
    \cropimg{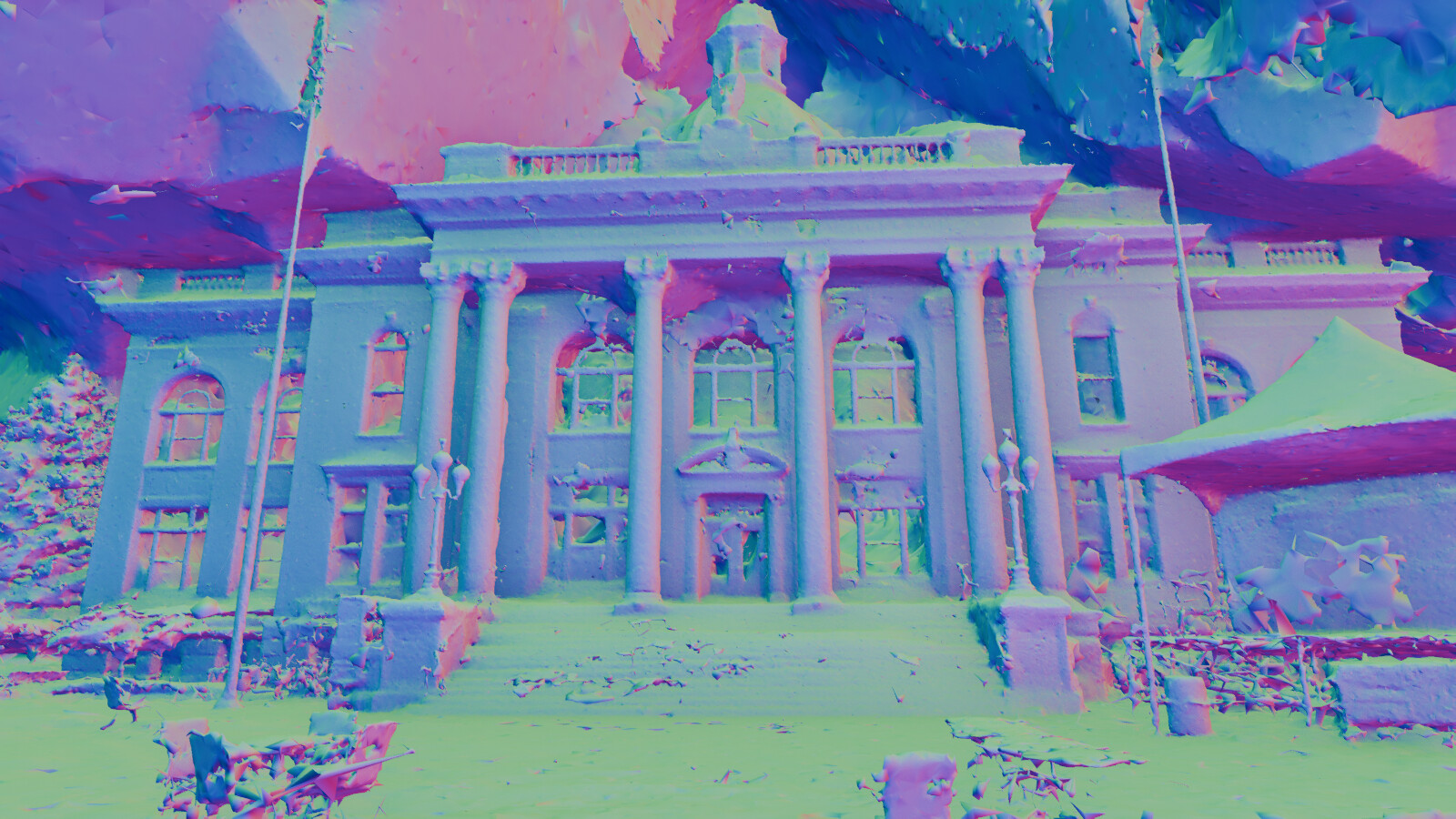}
    \cropimg{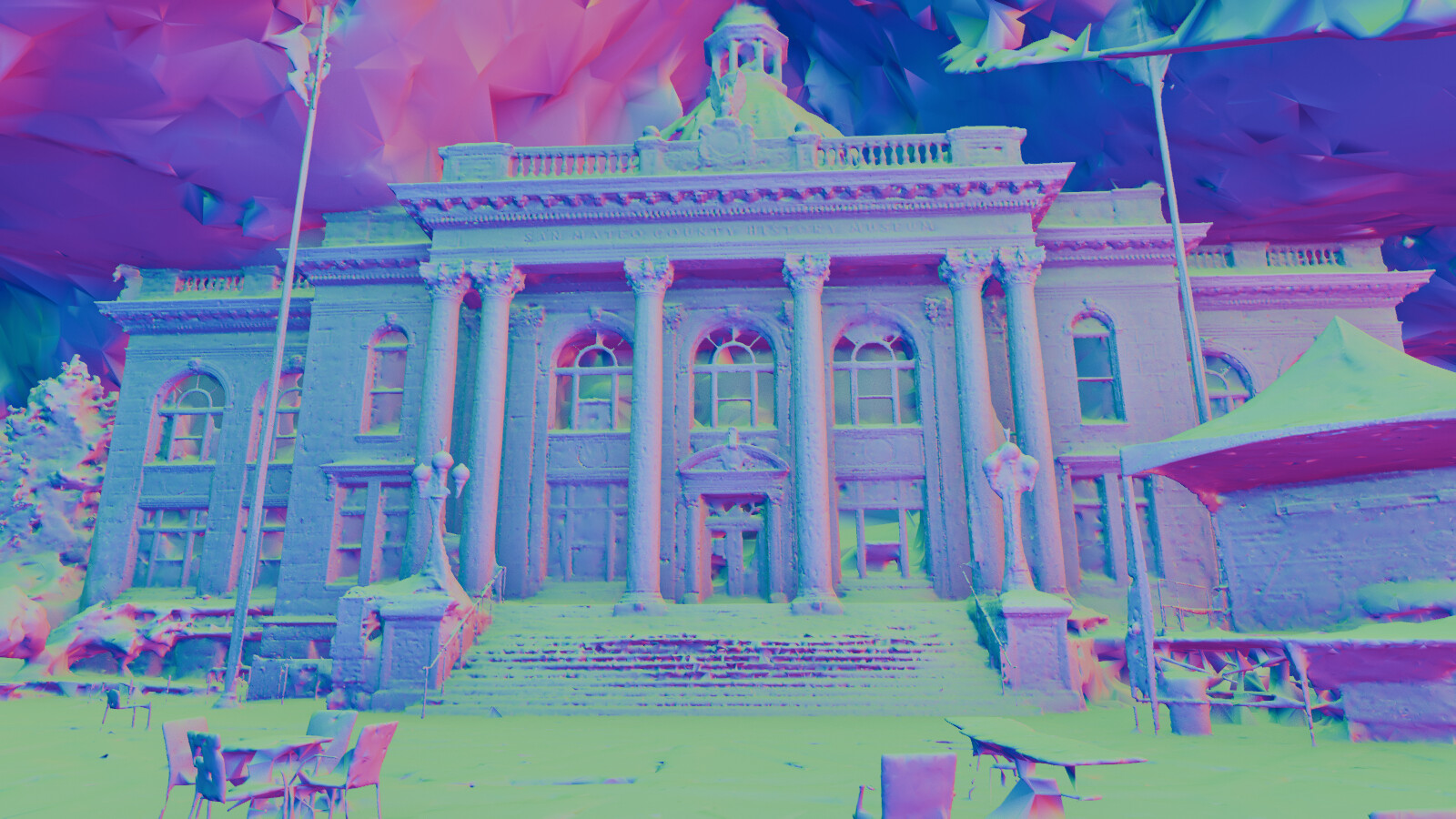}
    \\
    \cropimgs{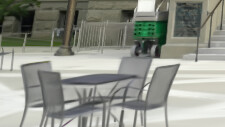}
    \cropimgs{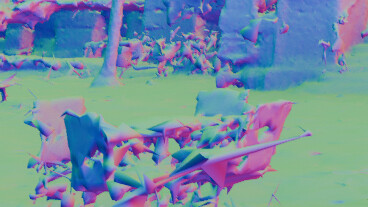}
    \cropimgs{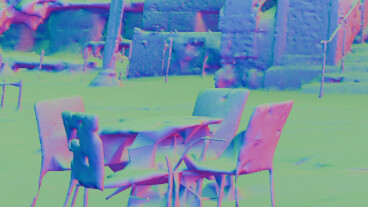}
    
    \vspace{1pt}

    {\centering{Scene 3: Truck}\par}\vspace{2pt}
    \cropimg{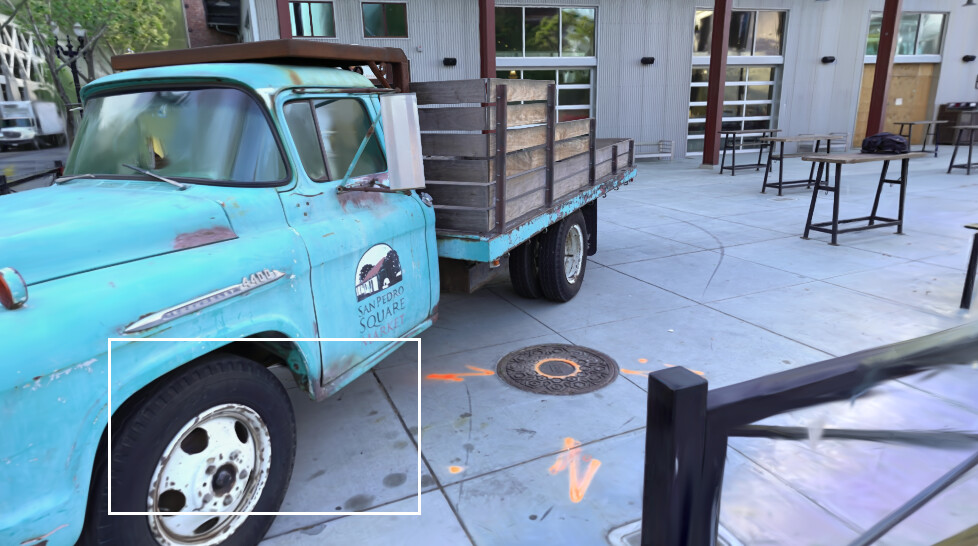}
    \cropimg{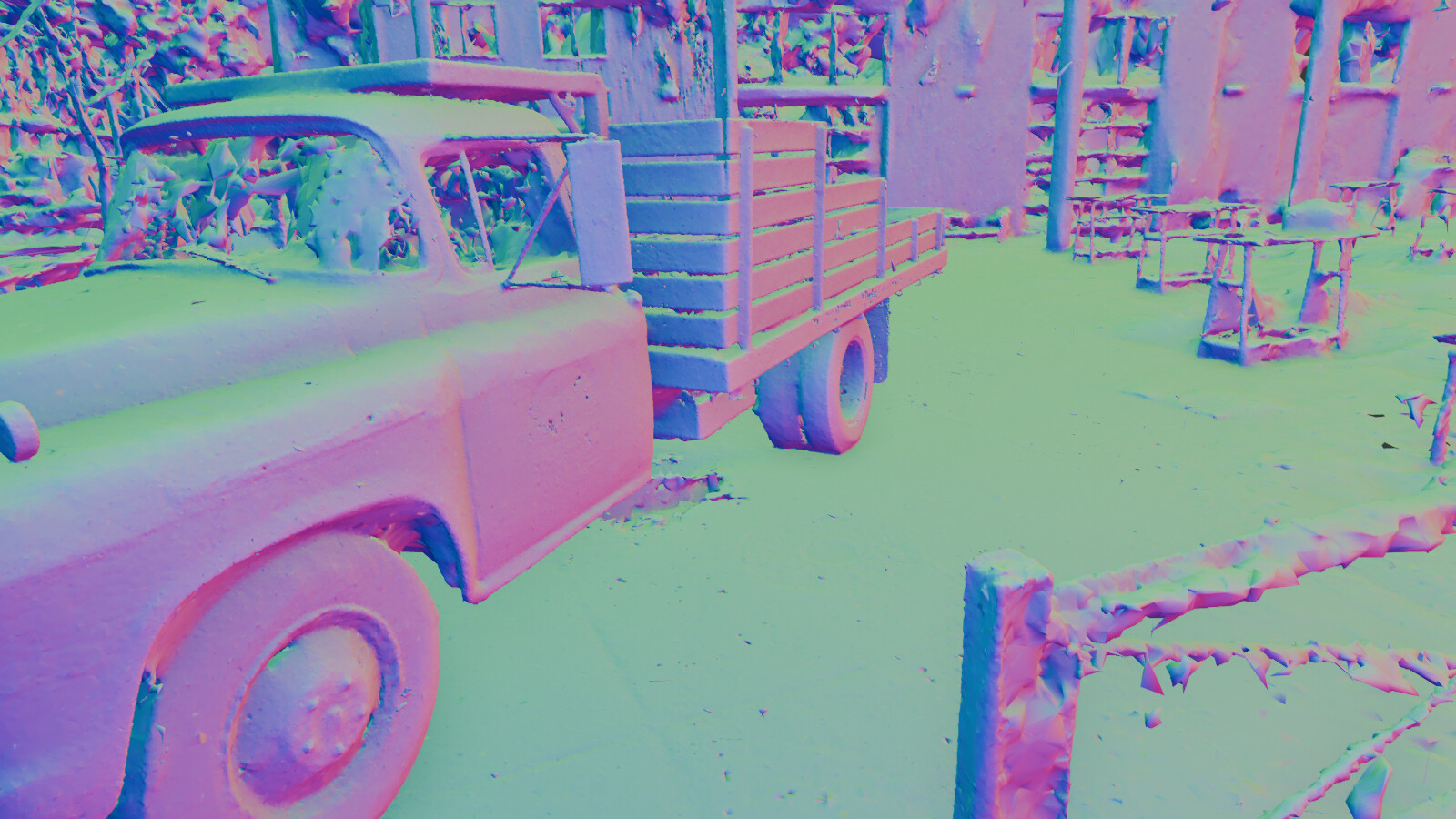}
    \cropimg{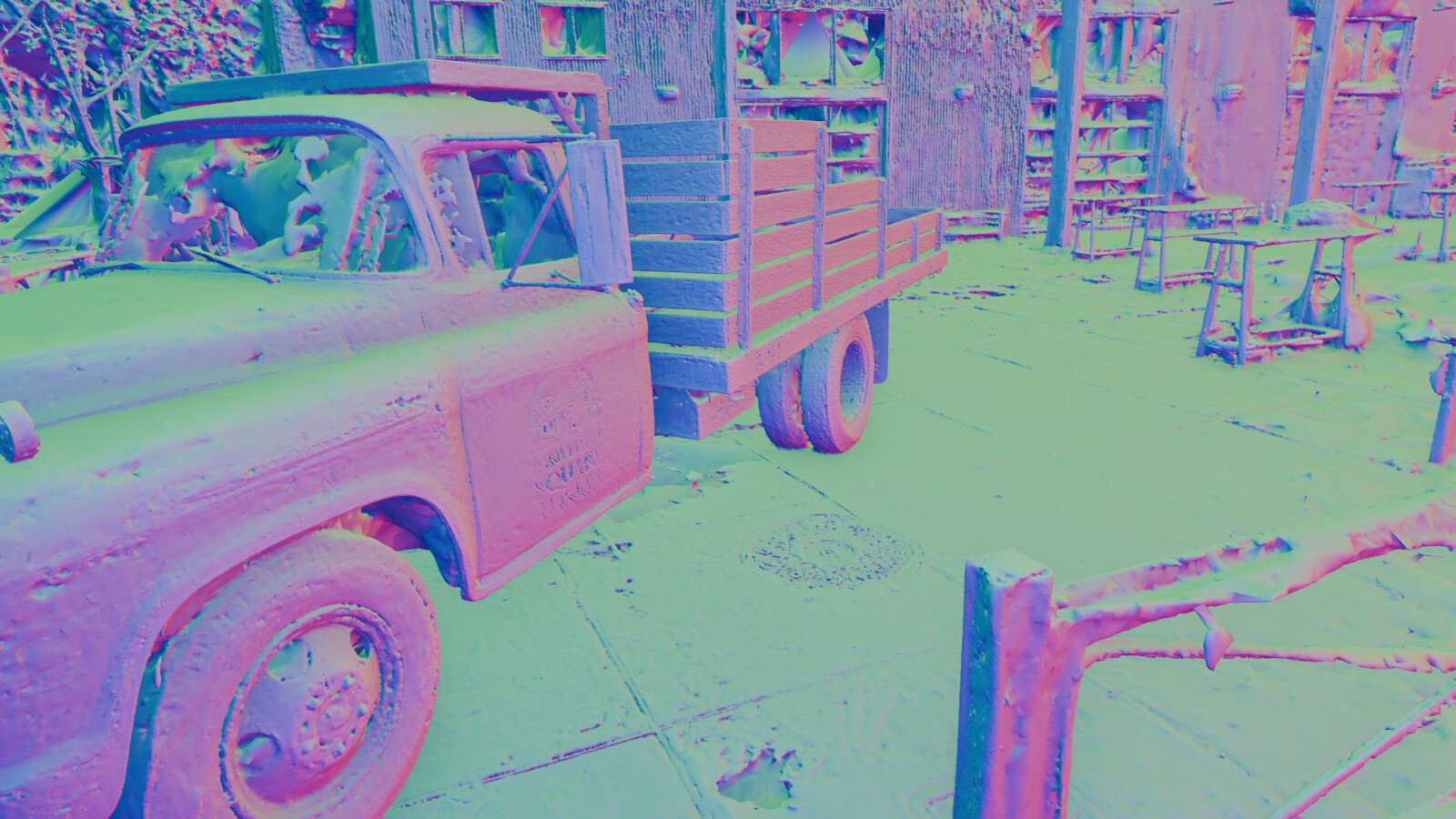}
    \\
    \cropimgs{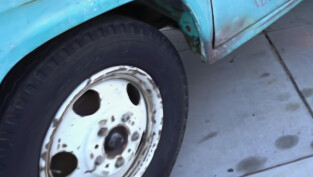}
    \cropimgs{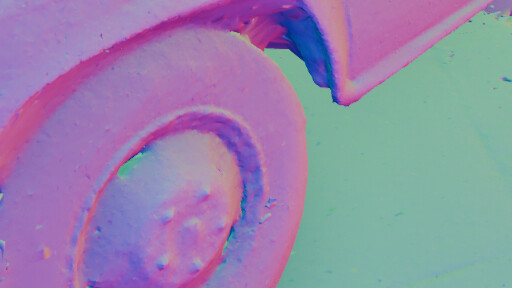}
    \cropimgs{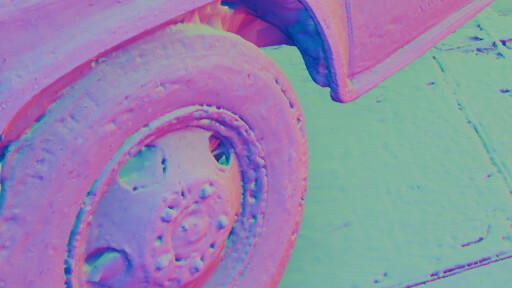}
    
    \vspace{1pt}
    
    {\centering{Scene 4: Meetingroom}\par}\vspace{2pt}
    \cropimg{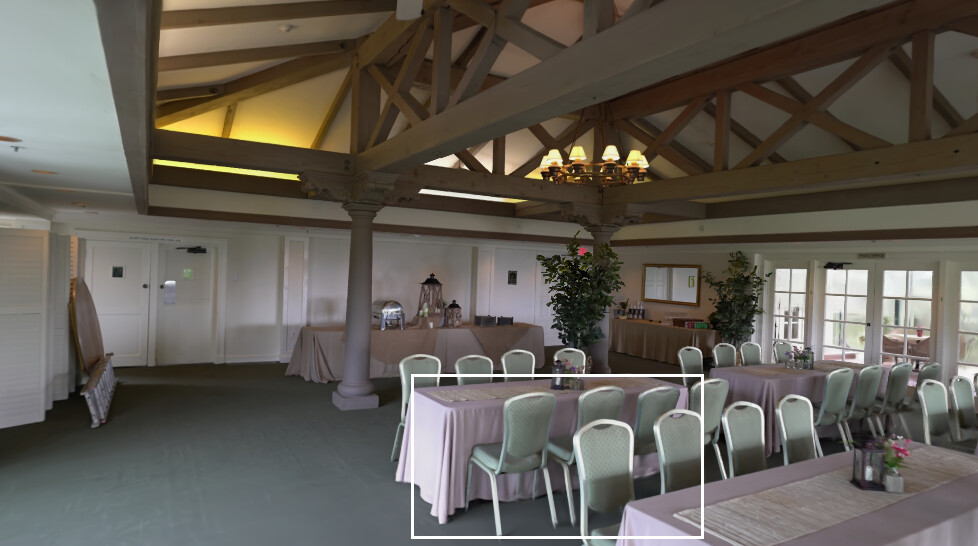}
    \cropimg{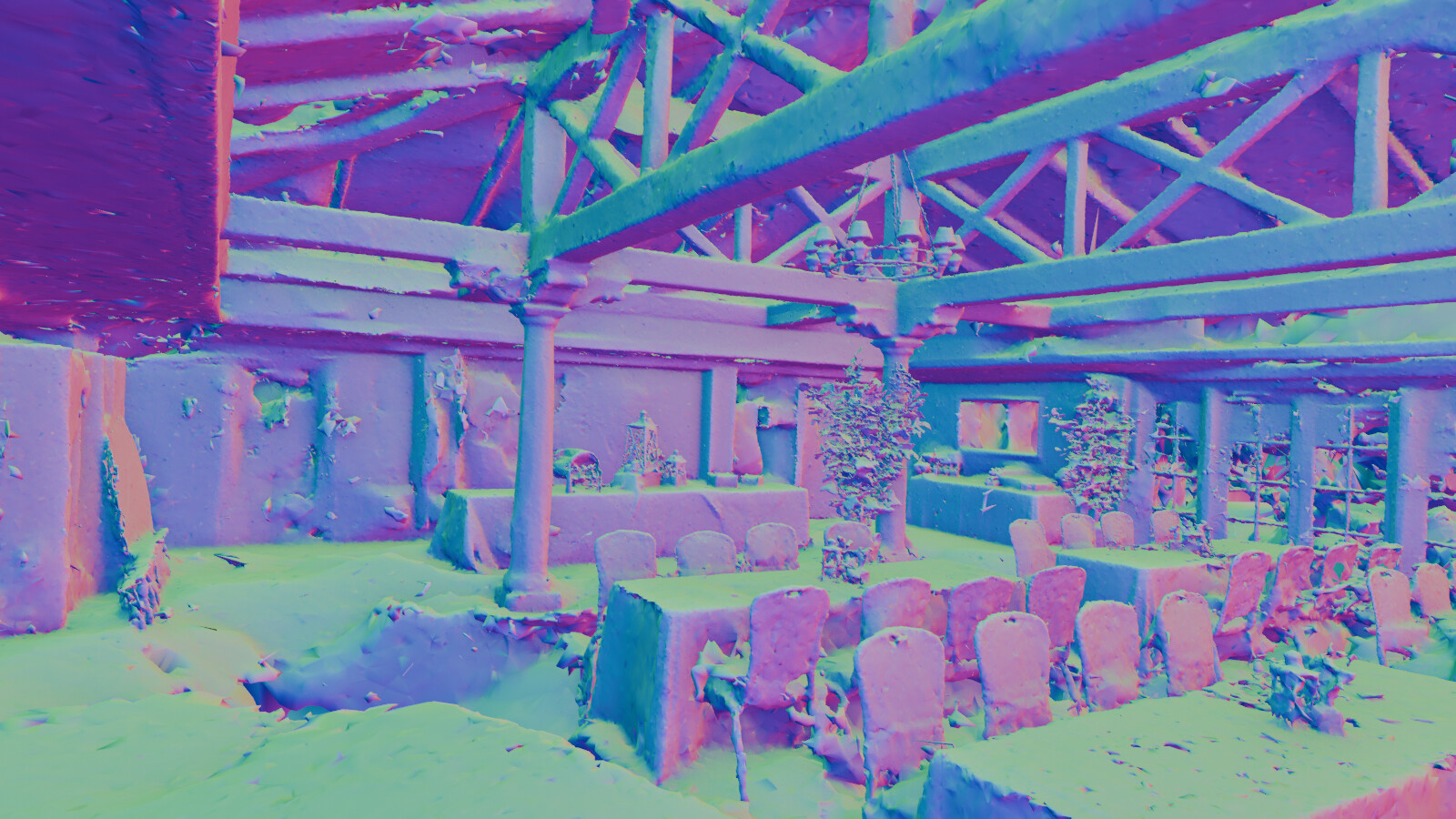}
    \cropimg{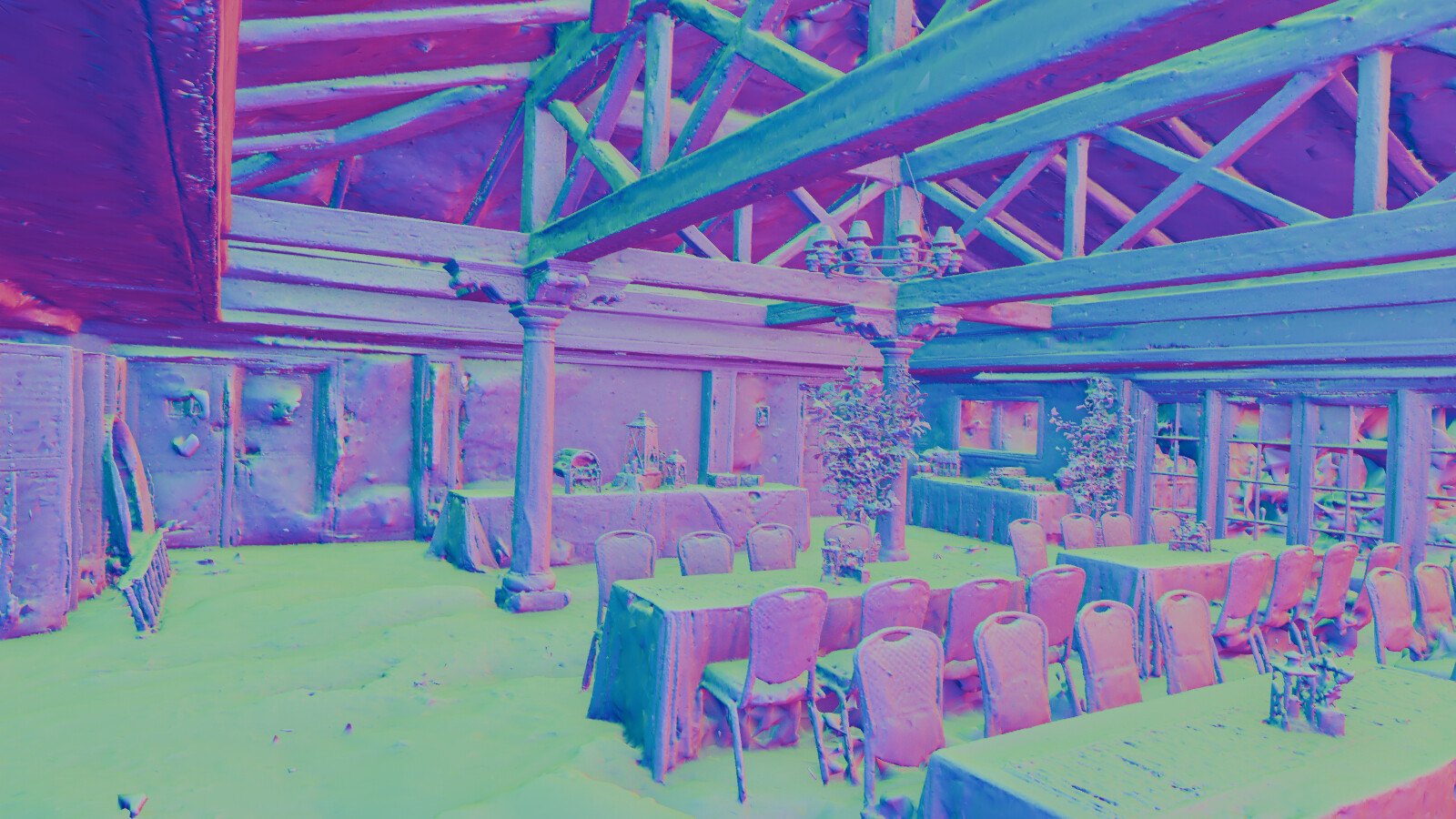}
    \\
    \cropimgs{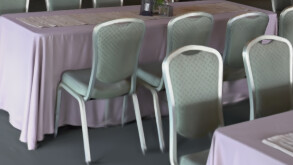}
    \cropimgs{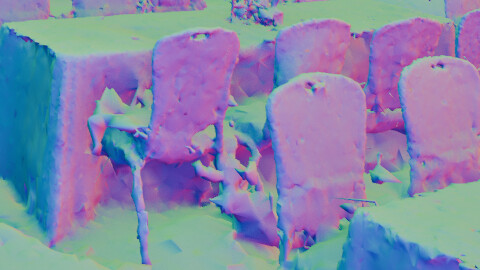}
    \cropimgs{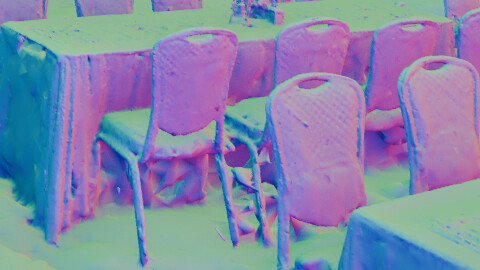}

    \caption{
\textbf{Novel View Mesh Rendering Comparison with Unbounded Methods.} 
Our method consistently preserves geometry and intricate topological details that \milo~\cite{guedon2025milo} smooths over or misrepresents.
}
    \label{fig:milo}
\end{figure*}

Further qualitative comparisons in \cref{fig:milo} show that our method is able to extract more detailed meshes, while suffering from fewer artifacts.

\subsection{DTU Results}
For completeness, we report results on the DTU dataset in \cref{tab:dtu}, reproducing all baseline results under identical conditions to guarantee a fair comparison.
Following the protocol established in recent literature~\cite{huang20242dgs, yu2024gof}, we evaluated the same 15 scenes and applied TSDF fusion to all methods using a consistent voxel size of $0.002$. 

Our method obtains the lowest Chamfer distance of all methods targeting unbounded reconstruction. 
However, in this specific context, characterized by small, bounded scenes containing a single object, methods incorporating multi-view priors~\cite{chen2024pgsr, zhang2025qgs} maintain a competitive edge.

\begin{table}[ht!]
    \centering
    \setlength{\tabcolsep}{4pt}
    \caption{
\textbf{Full Geometry Reconstruction results} for the DTU dataset~\cite{jensen2014large}, where we report the Chamfer distance (lower is better).
Our method attains the lowest average Chamfer distance for all methods designed to extract unbounded meshes.
}
\resizebox{.98\linewidth}{!}{
\begin{tabular}{lrrrrrrrrrrrrrrrr}
\toprule
Scan ID & 24 & 37 & 40 & 55 & 63 & 65 & 69 & 83 & 97 & 105 & 106 & 110 & 114 & 118 & 122 & Average \\
\midrule
PGSR & \cellcolor{tab_color!49} 0.38 & \cellcolor{tab_color!49} 0.67 & 0.42 & \cellcolor{tab_color!49} 0.35 & \cellcolor{tab_color!49} 0.76 & \cellcolor{tab_color!49} 0.58 & \cellcolor{tab_color!49} 0.53 & \cellcolor{tab_color!32} 1.08 & \cellcolor{tab_color!49} 0.70 & \cellcolor{tab_color!15} 0.61 & \cellcolor{tab_color!49} 0.45 & \cellcolor{tab_color!49} 0.55 & \cellcolor{tab_color!49} 0.34 & \cellcolor{tab_color!32} 0.42 & \cellcolor{tab_color!49} 0.40 & \cellcolor{tab_color!49} 0.55 \\
QGS & \cellcolor{tab_color!15} 0.45 & 0.77 & \cellcolor{tab_color!15} 0.37 & \cellcolor{tab_color!32} 0.37 & \cellcolor{tab_color!15} 0.84 & 0.91 & \cellcolor{tab_color!32} 0.55 & 1.15 & \cellcolor{tab_color!32} 1.00 & 0.62 & \cellcolor{tab_color!32} 0.46 & \cellcolor{tab_color!32} 0.70 & \cellcolor{tab_color!32} 0.36 & \cellcolor{tab_color!15} 0.45 & \cellcolor{tab_color!15} 0.41 & \cellcolor{tab_color!32} 0.63 \\
\midrule
GOF & 0.53 & 0.89 & 0.43 & \cellcolor{tab_color!15} 0.38 & 1.33 & 0.87 & 0.77 & 1.28 & 1.29 & 0.79 & 0.77 & 1.15 & 0.46 & 0.70 & 0.54 & 0.81 \\
SOF & 0.55 & \cellcolor{tab_color!32} 0.70 & 0.41 & 0.39 & 1.12 & \cellcolor{tab_color!15} 0.73 & \cellcolor{tab_color!15} 0.66 & \cellcolor{tab_color!15} 1.11 & 1.47 & \cellcolor{tab_color!32} 0.60 & 0.63 & 1.05 & 0.58 & 0.62 & 0.48 & 0.74 \\
\milo & 0.46 & 0.77 & \cellcolor{tab_color!32} 0.35 & 0.39 & \cellcolor{tab_color!32} 0.80 & 0.86 & 0.72 & 1.20 & 1.22 & 0.66 & 0.63 & \cellcolor{tab_color!15} 0.93 & \cellcolor{tab_color!15} 0.38 & 0.79 & 0.49 & 0.71 \\
Ours & \cellcolor{tab_color!32} 0.43 & \cellcolor{tab_color!32} 0.70 & \cellcolor{tab_color!49} 0.34 & 0.39 & 0.92 & \cellcolor{tab_color!32} 0.70 & 0.67 & \cellcolor{tab_color!49} 1.03 & \cellcolor{tab_color!15} 1.19 & \cellcolor{tab_color!49} 0.50 & \cellcolor{tab_color!15} 0.49 & 1.05 & 0.47 & \cellcolor{tab_color!49} 0.41 & \cellcolor{tab_color!49} 0.40 & \cellcolor{tab_color!15} 0.65 \\
\midrule
\multicolumn{4}{l}{Ablation Study} \\\midrule
SOF~\cite{Radl2025SOF} (baseline) & 0.55 & \cellcolor{tab_color!15} 0.70 & 0.41 & \cellcolor{tab_color!32} 0.39 & 1.12 & \cellcolor{tab_color!32} 0.73 & \cellcolor{tab_color!32} 0.66 & 1.11 & 1.47 & 0.60 & 0.63 & \cellcolor{tab_color!15} 1.05 & 0.58 & 0.62 & 0.48 & 0.74 \\
+ Improved Appearance & 0.50 & \cellcolor{tab_color!49} 0.64 & 0.43 & 0.40 & 1.00 & \cellcolor{tab_color!15} 0.74 & 0.72 & 1.13 & \cellcolor{tab_color!32} 1.21 & 0.59 & 0.66 & 1.06 & 0.48 & 0.57 & 0.49 & 0.71 \\
+ $\msub{\mathcal{L}}{conf}$ & \cellcolor{tab_color!15} 0.47 & \cellcolor{tab_color!32} 0.65 & \cellcolor{tab_color!15} 0.36 & 0.41 & \cellcolor{tab_color!15} 0.96 & 0.80 & \cellcolor{tab_color!15} 0.67 & \cellcolor{tab_color!15} 1.05 & 1.24 & \cellcolor{tab_color!32} 0.50 & \cellcolor{tab_color!32} 0.50 & \cellcolor{tab_color!49} 0.74 & \cellcolor{tab_color!49} 0.42 & \cellcolor{tab_color!15} 0.43 & \cellcolor{tab_color!15} 0.42 & \cellcolor{tab_color!49} 0.64 \\
+ $\msub{\mathcal{L}}{color-var}$ & \cellcolor{tab_color!32} 0.44 & 0.72 & \cellcolor{tab_color!32} 0.35 & \cellcolor{tab_color!49} 0.38 & \cellcolor{tab_color!32} 0.95 & 0.76 & \cellcolor{tab_color!49} 0.65 & \cellcolor{tab_color!49} 0.99 & \cellcolor{tab_color!15} 1.23 & \cellcolor{tab_color!49} 0.49 & \cellcolor{tab_color!15} 0.53 & \cellcolor{tab_color!32} 1.03 & \cellcolor{tab_color!49} 0.42 & \cellcolor{tab_color!49} 0.40 & \cellcolor{tab_color!49} 0.40 & \cellcolor{tab_color!32} 0.65 \\
+ $\msub{\mathcal{L}}{normal-var}$ & \cellcolor{tab_color!49} 0.43 & \cellcolor{tab_color!15} 0.70 & \cellcolor{tab_color!49} 0.34 & \cellcolor{tab_color!32} 0.39 & \cellcolor{tab_color!49} 0.92 & \cellcolor{tab_color!49} 0.70 & \cellcolor{tab_color!15} 0.67 & \cellcolor{tab_color!32} 1.03 & \cellcolor{tab_color!49} 1.19 & \cellcolor{tab_color!32} 0.50 & \cellcolor{tab_color!49} 0.49 & \cellcolor{tab_color!15} 1.05 & \cellcolor{tab_color!15} 0.47 & \cellcolor{tab_color!32} 0.41 & \cellcolor{tab_color!49} 0.40 & \cellcolor{tab_color!32} 0.65 \\
\bottomrule
\end{tabular}
}
    \label{tab:dtu}
\end{table}
We additionally provide an ablation study of our core components in \cref{tab:dtu}.
Our improved appearance module already provides slight improvement in Chamfer distance. 
Results improve further when enabling $\msub{\mathcal{L}}{conf}$.
This configuration (w/o our variance losses) attains the lowest average Chamfer distance for the DTU dataset~\cite{jensen2014large}.
However, when comparing the results to configurations which use $\msub{\mathcal{L}}{color-var}$ and $\msub{\mathcal{L}}{normal-var}$, we see that this quantitative improvement is mainly caused by vastly improved results for scan 110.
These results once again underscore the robustness of our method.
Overall, our techniques lead to an improvement of $\mathbf{0.09}$ in Chamfer distance compared to the baseline SOF~\cite{Radl2025SOF}.

\subsection{Novel View Synthesis Results}
\label{app:confidence}
To round out our evaluation, we provide additional novel view synthesis results on the popular Mip-NeRF 360 dataset~\cite{barron2022mipnerf360}.
While this task is not a focus of this work, we are interested in the effect of our confidence framework on image quality.
Following SOF~\cite{Radl2025SOF}, we disabled both $\msub{\mathcal{L}}{geom}$ and our appearance module for this ablation, as they inherently impact standard image quality metrics.
To further isolate the effect of our confidence-framework, we also disable our variance losses, as they inherently constrain blending.
Other than that, we use the same set of hyperparameters as in previous experiments. 

We report the quantitative results in \cref{tab:nvs}, where we also report numbers for all other tested methods; these numbers were taken from~\cite{zhang2025qgs, Radl2025SOF, guedon2025milo}, respectively.
Notably, $\msub{\mathcal{L}}{conf}$ does not negatively impact image quality metrics, while significantly decreasing primitive count: 
We observe an overall primitive reduction of $\mathbf{26\%}$, while the averaged PSNR over all scenes drops by merely $\mathbf{0.02}$ dB.
\begin{table}[ht!]
    \centering
    \setlength{\tabcolsep}{4pt}
    \caption{
\textbf{Novel View Synthesis Ablation} for the Mip-NeRF 360 dataset~\cite{barron2022mipnerf360}, where we report PSNR, SSIM and LPIPS.
Notably, adding $\msub{\mathcal{L}}{conf}$ does not negatively impact image quality metrics overall.
Results with \textsuperscript{$\dagger$} were taken from~\cite{zhang2025qgs, Radl2025SOF, guedon2025milo}.
}
\resizebox{.98\linewidth}{!}{
\begin{tabular}{lrrrrrrr}
\toprule
& \multicolumn{3}{c}{Mip-NeRF 360 Indoor} & \multicolumn{3}{c}{Mip-NeRF 360 Outdoor} \\
\cmidrule(lr){2-4}\cmidrule(lr){5-7}
 & PSNR\textsuperscript{$\uparrow$} & SSIM\textsuperscript{$\uparrow$} & LPIPS\textsuperscript{$\downarrow$} & PSNR\textsuperscript{$\uparrow$} & SSIM\textsuperscript{$\uparrow$} & LPIPS\textsuperscript{$\downarrow$} & \# Gaussians \\
\midrule
PGSR\textsuperscript{$\dagger$} & \cellcolor{tab_color!15} 30.35 & \cellcolor{tab_color!49} 0.924 & \cellcolor{tab_color!49} 0.176 & 24.29 & 0.718 & 0.236 & - \\
QGS\textsuperscript{$\dagger$} & \cellcolor{tab_color!32} 30.45 & \cellcolor{tab_color!15} 0.919 & \cellcolor{tab_color!15} 0.184 & 24.32 & 0.706 & 0.242 & -\\
\midrule
GOF\textsuperscript{$\dagger$} & \cellcolor{tab_color!49} 30.47 & 0.918 & 0.189 & \cellcolor{tab_color!49} 24.79 & \cellcolor{tab_color!49} 0.745 & \cellcolor{tab_color!49} 0.208 & -\\
SOF\textsuperscript{$\dagger$} & 30.13 & 0.916 & 0.188 & \cellcolor{tab_color!32} 24.78 & \cellcolor{tab_color!32} 0.745 & \cellcolor{tab_color!32} 0.208 & -\\
\milo\textsuperscript{$\dagger$} & 29.96 & \cellcolor{tab_color!32} 0.920 & 0.191 & 24.47 & 0.718 & 0.290 & -\\
\midrule
Ours & 29.92 & 0.911 & 0.193 & \cellcolor{tab_color!15} 24.69 & \cellcolor{tab_color!15} 0.742 & \cellcolor{tab_color!15} 0.211 & 2.89M\\
Ours w/ conf & 30.24 & 0.919 & \cellcolor{tab_color!32} 0.179 & 24.41 & 0.719 & 0.258 & \cellcolor{tab_color!32}2.12M\\
\bottomrule
\end{tabular}
}
    \label{tab:nvs}
\end{table}

\para{Confidence Analysis}
When analyzing the novel view synthesis results further, we find that the learned confidences differ significantly between indoor and outdoor scenes.
Notably, for indoor scenes, we observe an average confidence of $\mathbf{3.22}$  on test views, whereas the average confidence for outdoor scenes is $\mathbf{1.66}$. 
This effect is visualized in \cref{fig:conf_ablation}.
\newcommand{\mipanno}[2]{%
    \stackinset{l}{1pt}{b}{1pt}{
        \begin{tikzpicture}
            \node[fill=black, fill opacity=0.5, text opacity=1, 
                  inner sep=1.5pt, rounded corners=0.5pt] 
            {\sffamily\fontsize{5}{4}\selectfont{\textcolor{white}{#2}}};
        \end{tikzpicture}%
    }{#1}
}

\begin{figure}
    \centering
    \scriptsize\sffamily
    \begin{minipage}{0.32\textwidth}
        \centering \small Render
    \end{minipage}
    \begin{minipage}{0.32\textwidth}
        \centering \small Confidence $\hat{C}$
    \end{minipage}
    \begin{minipage}{0.32\textwidth}
        \centering \small Mean Squared Error
    \end{minipage}
    \vspace{2pt}
    \mipanno{\includegraphics[width=0.32\linewidth]{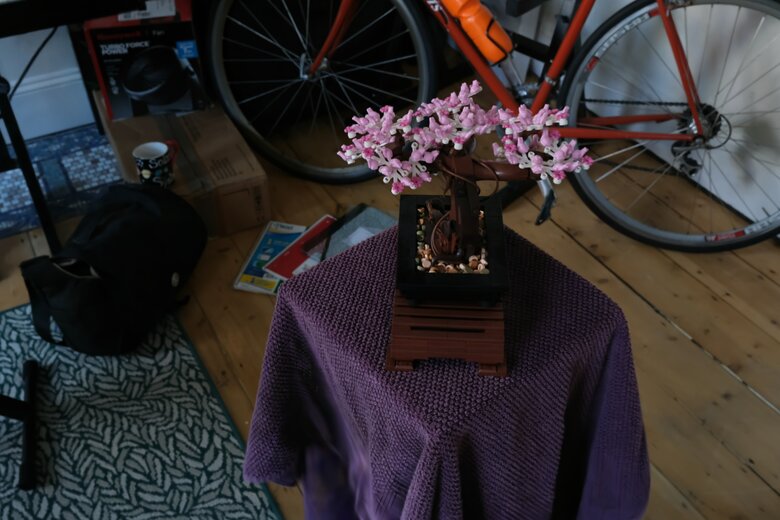}}{Bonsai}
    \mipanno{\includegraphics[width=0.32\linewidth]{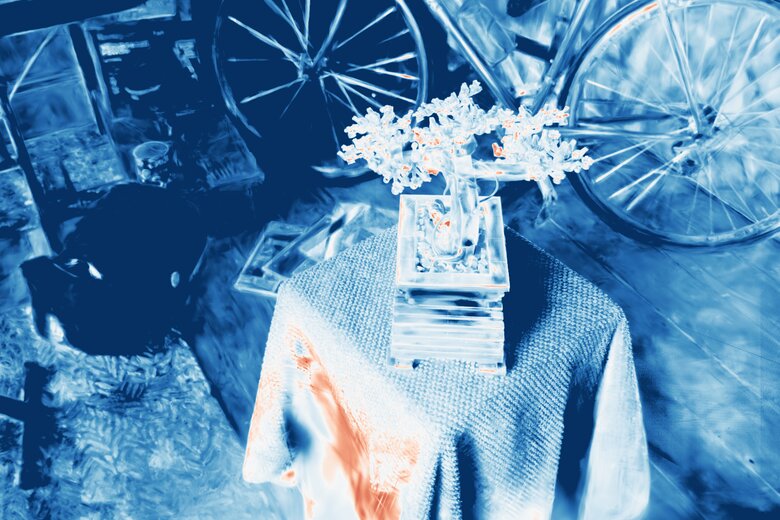}}{$\text{mean}(\hat{\mathsf{C}}) = 3.62$}
    \includegraphics[width=0.32\linewidth]{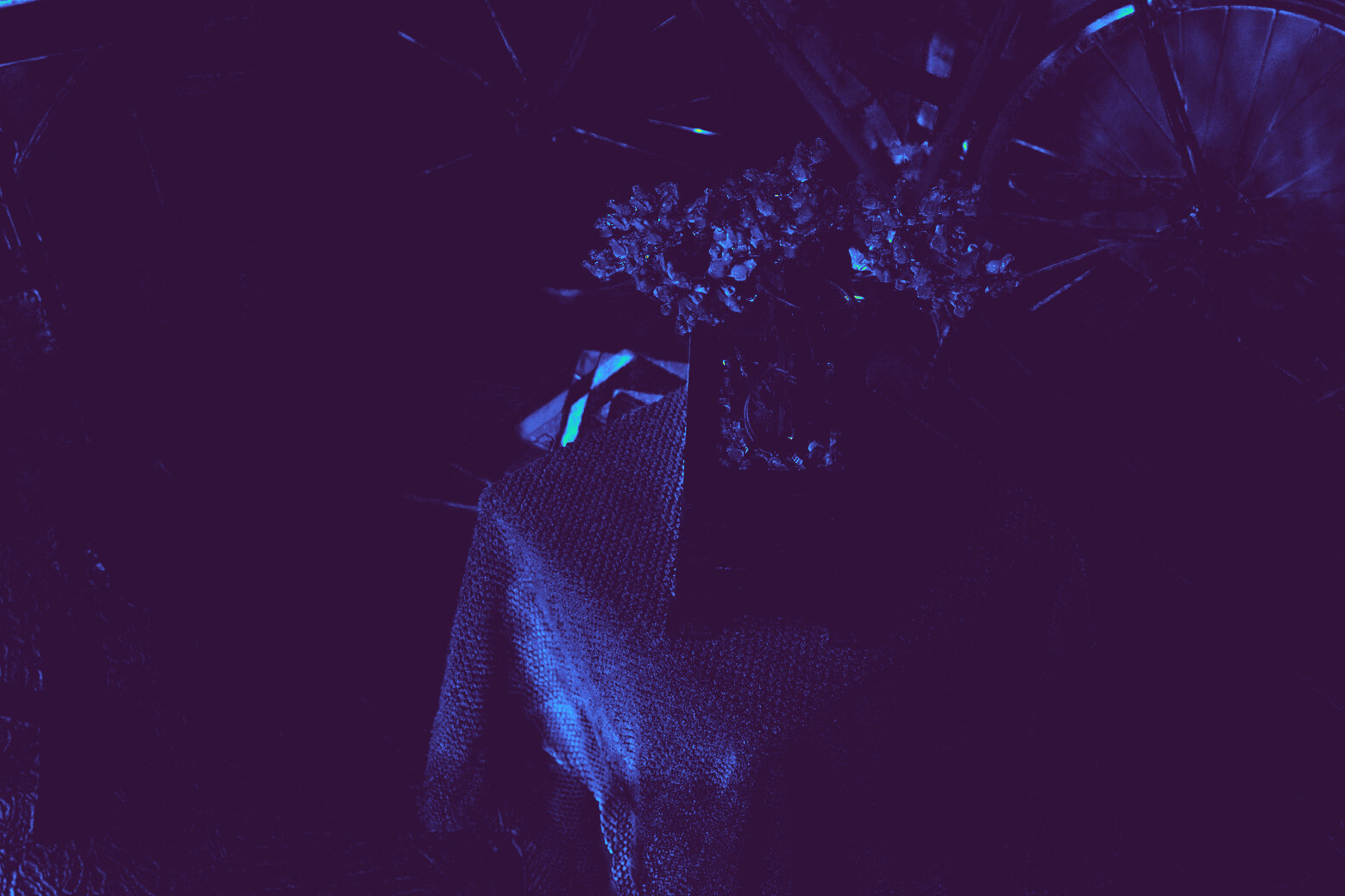} \\
    \mipanno{\includegraphics[width=0.32\linewidth]{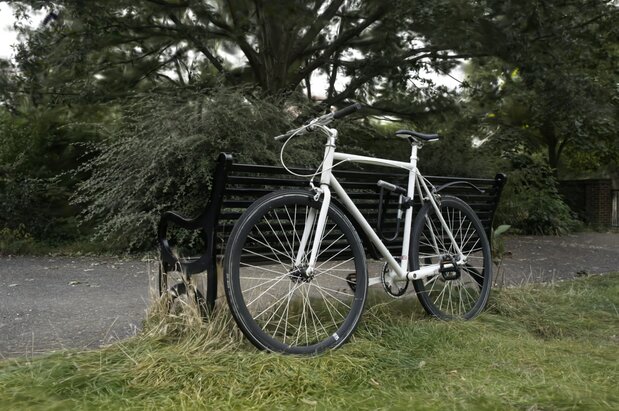}}{Bicycle}
    \mipanno{\includegraphics[width=0.32\linewidth]{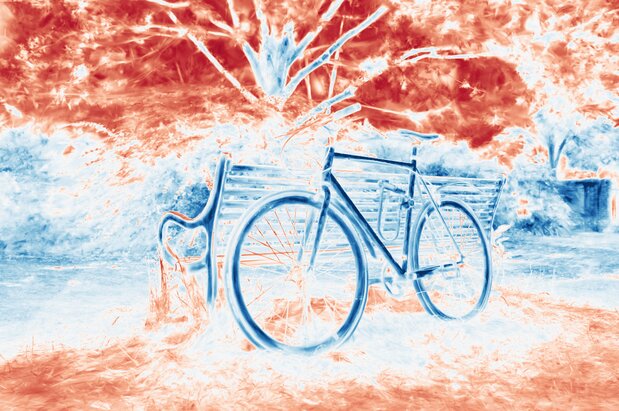}}{$\text{mean}(\hat{\mathsf{C}}) = 1.48$}
    \includegraphics[width=0.32\linewidth]{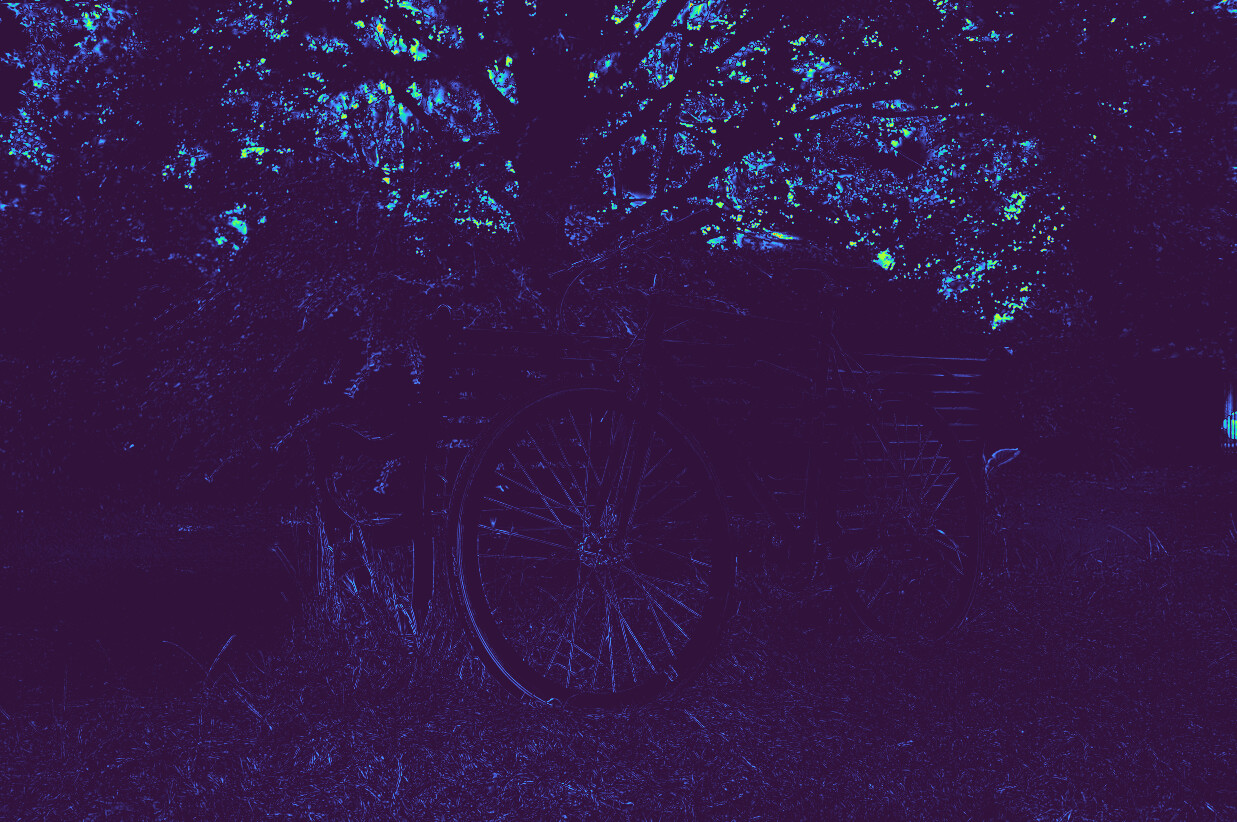}
    \caption{
\textbf{Confidence Evaluation:}
Our learned confidence directly correlates with the error maps, demonstrating that it can detect under-reconstructed regions.
    }
    \label{fig:conf_ablation}
\end{figure}

Crucially, we find that the improvement in image quality metrics is directly correlated to these average test-view confidences.
In high-confidence indoor scenes, our formulation strictly improves rendering metrics while naturally allocating more primitives to accurately capture well-defined structures (an increase from 1.04M to 1.57M). 
Conversely, in outdoor environments, the network predicts lower confidence for ambiguous, highly complex backgrounds. 
This elegantly prunes spurious geometry, leading to a massive reduction in the primitive count (from 4.38M down to 2.58M), without severely degrading photometric fidelity. 
This dynamic, scene-adaptive behavior once again underscores the potential of our method to be used as a standalone densification mechanism.

\subsection{Variance Losses}
\noindent \newcommand{\annot}[2]{%
    \stackinset{l}{1pt}{t}{1pt}{
        \begin{tikzpicture}
            \node[fill=black, fill opacity=0.5, text opacity=1, 
                  inner sep=1.5pt, rounded corners=0.5pt] 
            {\fontsize{2}{3}\selectfont{\textcolor{white}{#2}}};
        \end{tikzpicture}%
    }{#1}
}
\begin{figure}[t]
    \centering
    \scriptsize\sffamily 
    
    \begin{minipage}{0.32\linewidth}
        \centering First Hit
    \end{minipage}
    \begin{minipage}{0.32\linewidth}
        \centering Last Hit
    \end{minipage}
    \begin{minipage}{0.32\linewidth}
        \centering Full alpha-blending
    \end{minipage}
    \vspace{2pt}

    \annot{\includegraphics[width=0.32\linewidth]{images/render_variance/first_hit_color_var_00006_19.00685691833496.jpg}}{w/ $\msub{\mathcal{L}}{color-var}$}
    \includegraphics[width=0.32\linewidth]{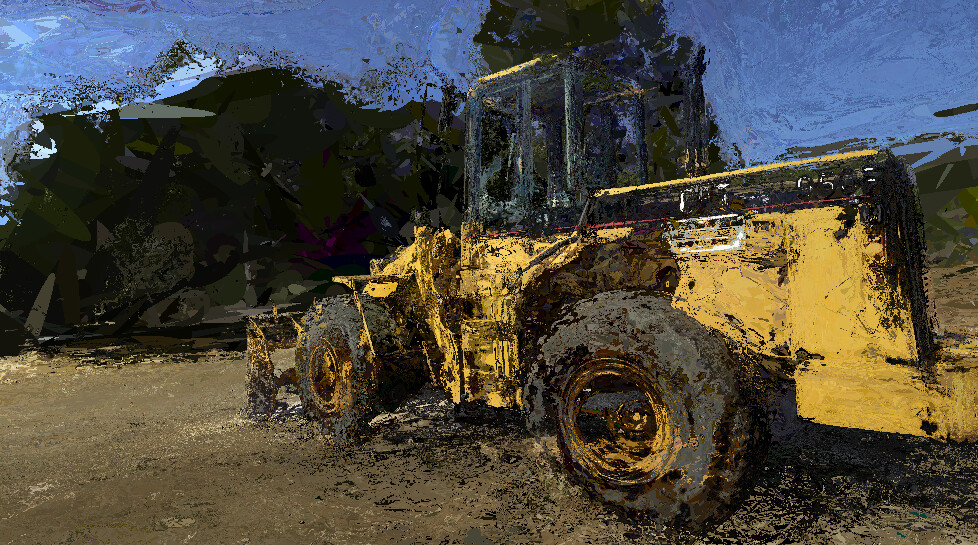}
    \includegraphics[width=0.32\linewidth]{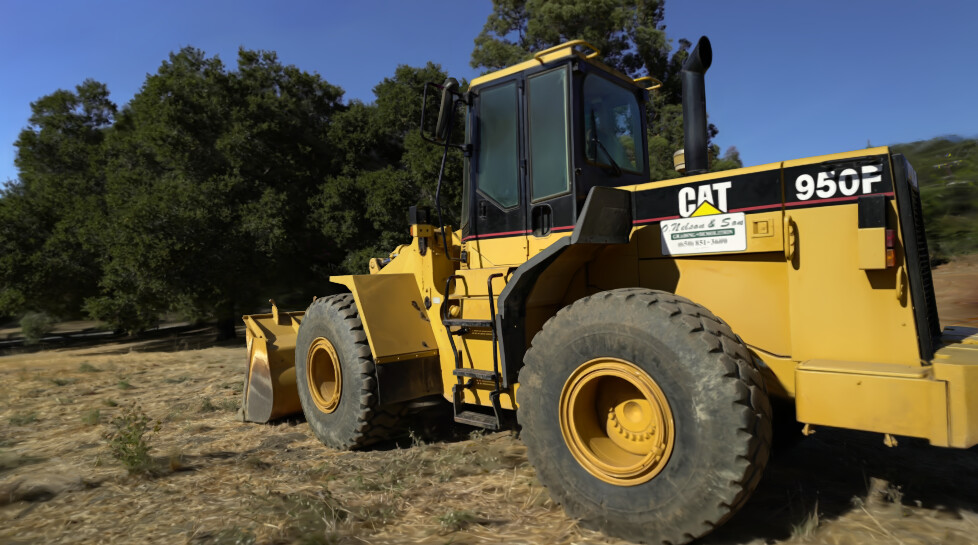}
    \\ \vspace{2pt} 
    
    \annot{\includegraphics[width=0.32\linewidth]{images/render_variance/first_hit_no_var_00006_14.12417984008789.jpg}}{w/o $\msub{\mathcal{L}}{color-var}$}
    \includegraphics[width=0.32\linewidth]{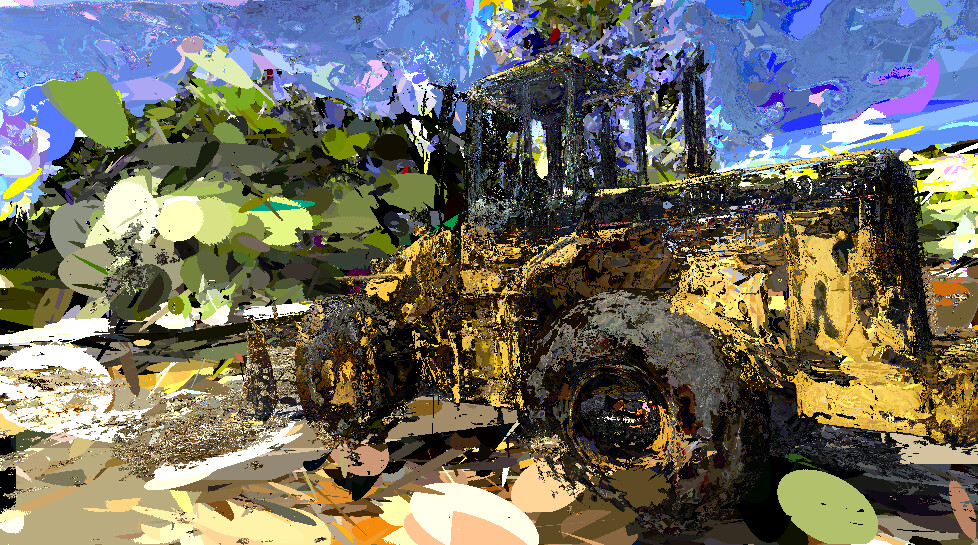}
    \includegraphics[width=0.32\linewidth]{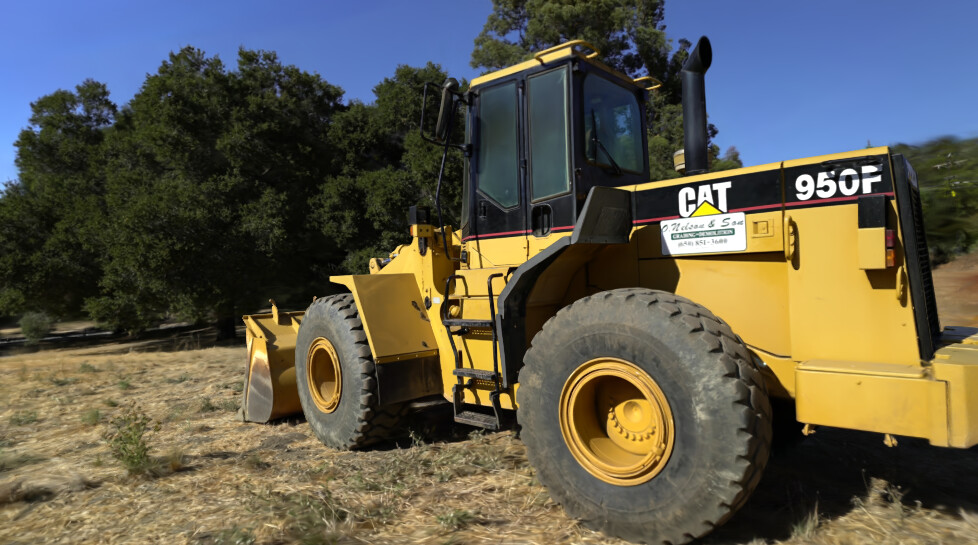} 
    \\ \vspace{4pt} 
    
    \annot{\includegraphics[width=0.32\linewidth]{images/render_variance/first_hit_normal_normal_var_00006_18.88519287109375.jpg}}{w/ $\msub{\mathcal{L}}{normal-var}$}
    \includegraphics[width=0.32\linewidth]{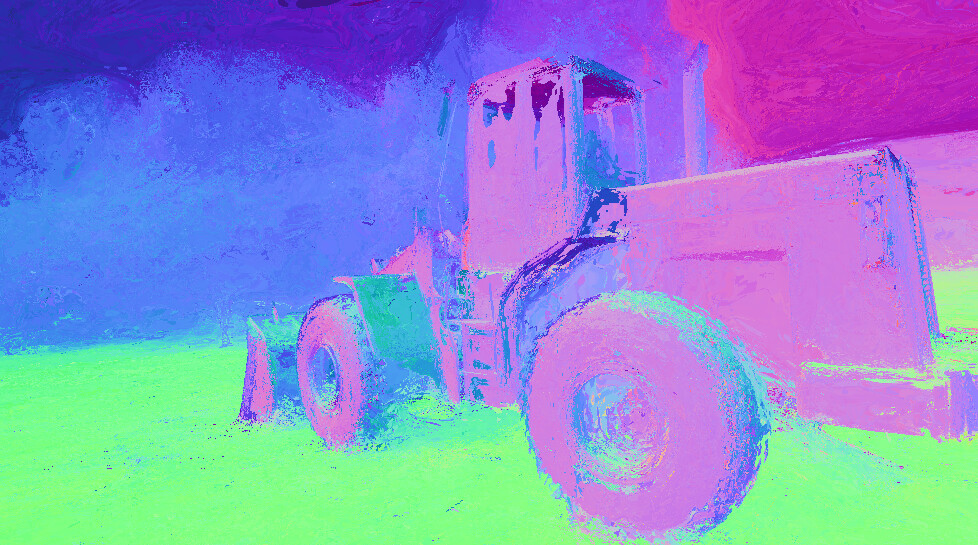}
    \includegraphics[width=0.32\linewidth]{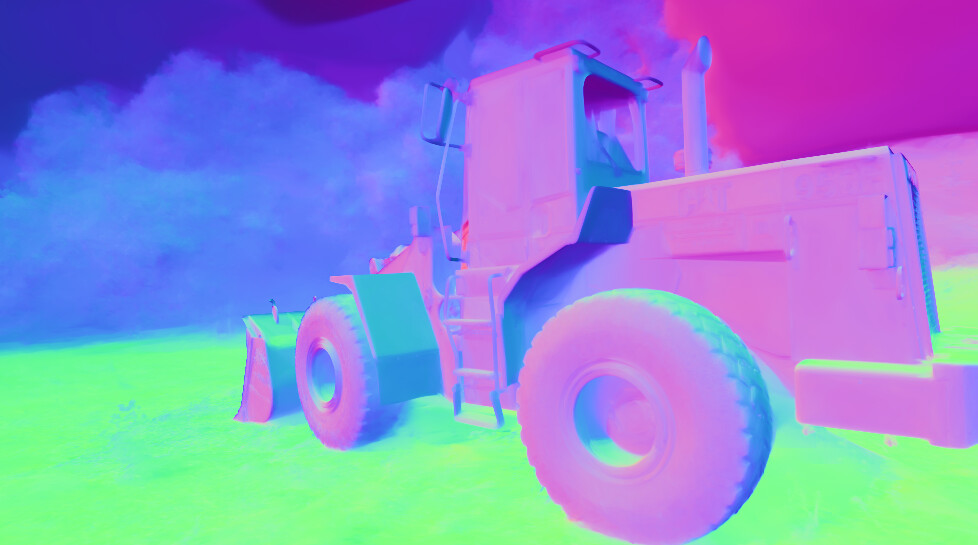} 
    \\ \vspace{2pt} 
    
    \annot{\includegraphics[width=0.32\linewidth]{images/render_variance/first_hit_normal_no_var_00006_14.12417984008789.jpg}}{w/o $\msub{\mathcal{L}}{normal-var}$}
    \includegraphics[width=0.32\linewidth]{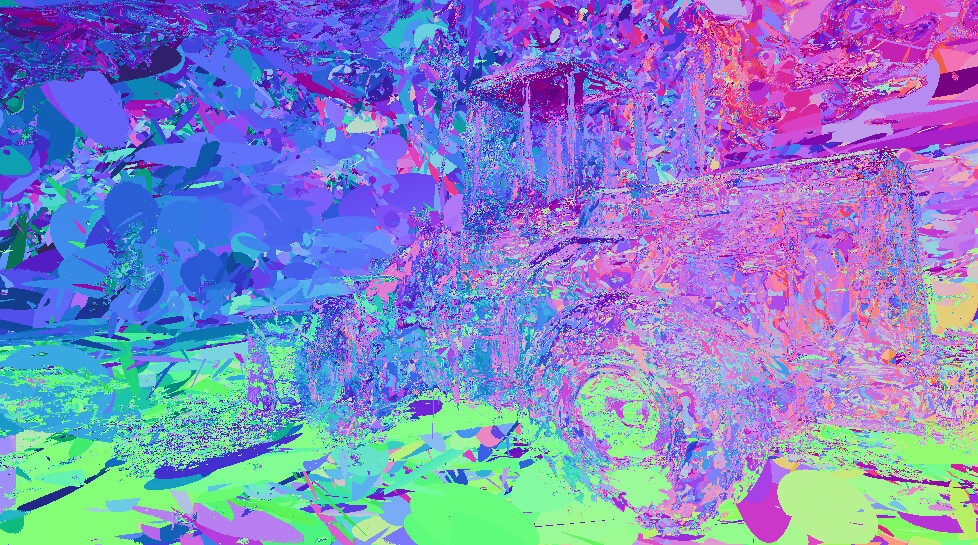}
    \includegraphics[width=0.32\linewidth]{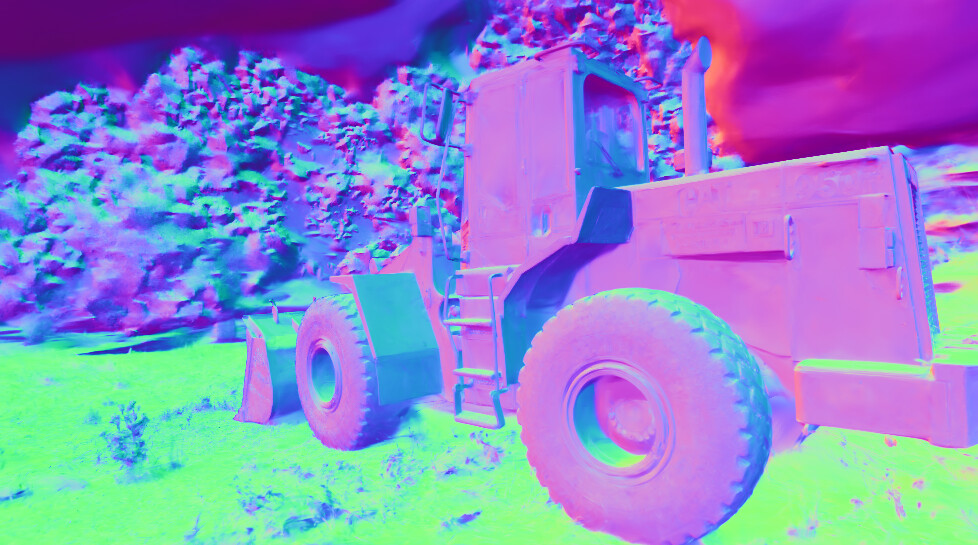}
    \caption{
\textbf{Effect of our proposed Variance Losses:}
We visualize the \emph{first-hit}, \emph{last-hit} and fully alpha-blended colors and normals.
As can be seen, our variance losses constrain spurious geometry and lead to smoother normal maps.
}
    \label{fig:color_variance}
\end{figure}

To motivate our variance losses, we provide additional visualizations in \cref{fig:color_variance}, which depict rendered color and normals using \emph{first-hit} and \emph{last-hit} rendering, \ie the quantities are only shown for the first and last blended Gaussians.

We observe that our variance losses result in smoother blended normals. 
Crucially, this does not result in a loss of detail for the final reconstruction (\cf \cref{fig:milo}).
Our variance losses provide a strong optimization signal for every primitive to be aligned with object surfaces, removing spurious geometry.

\section{Appearance Embeddings}
\begin{figure}[t]
    \centering
    
    \begin{minipage}{0.49\textwidth}
        \centering \small VastGaussian
    \end{minipage}\hfill
    \begin{minipage}{0.49\textwidth}
        \centering \small Ours
    \end{minipage}
    \vspace{2pt} 
    
    \includegraphics[width=0.49\textwidth]{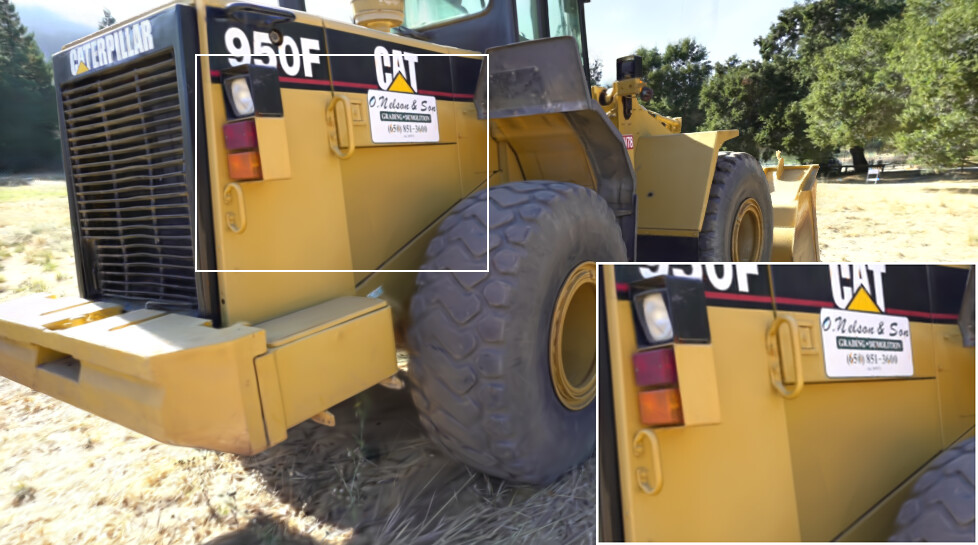}\hfill
    \includegraphics[width=0.49\textwidth]{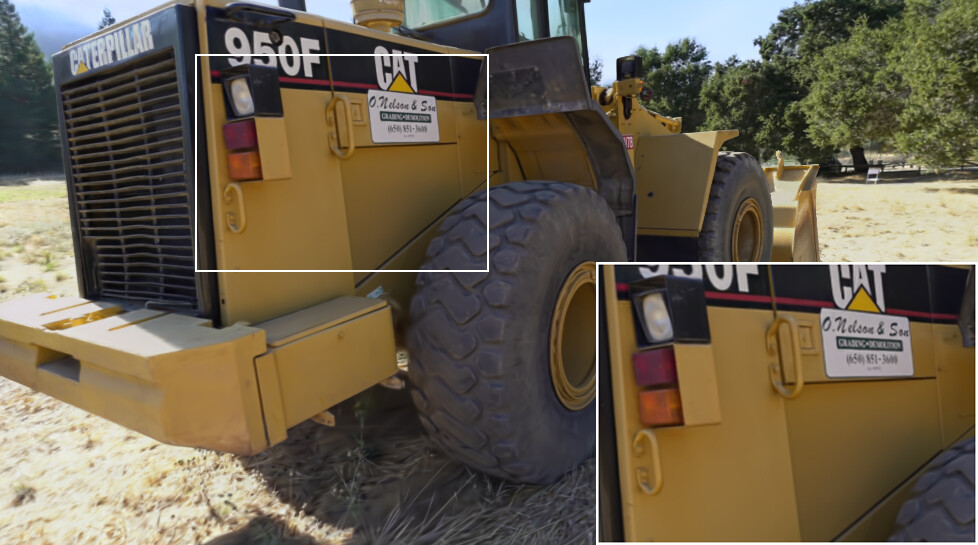}
    \\ \vspace{2pt}
    
    \includegraphics[width=0.49\textwidth]{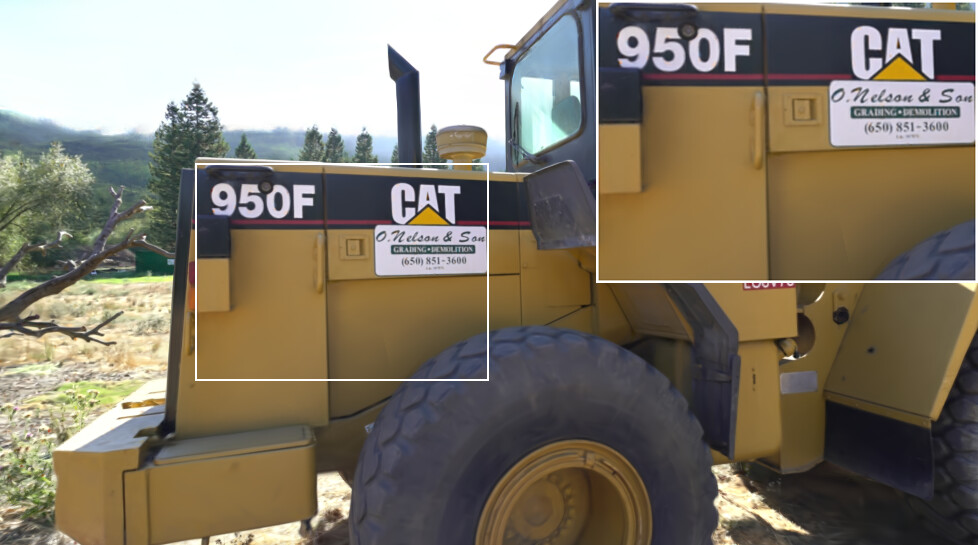}\hfill
    \includegraphics[width=0.49\textwidth]{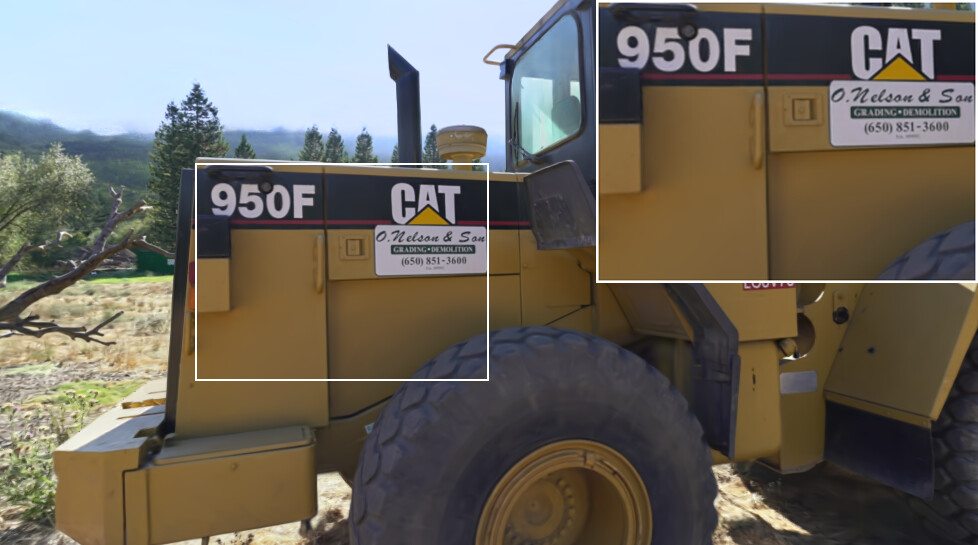}
    
    \caption{
\textbf{Effect of our SSIM-decoupled Appearance Module}:
We show rendered images from two spatially close camera poses.
As can be seen, the illumination for Ours is more consistent; this is caused by our improved appearance module.
    }
    \label{fig:illumination}
\end{figure}
We provide an additional qualitative evaluation of our improved appearance embedding in \cref{fig:illumination}.
Here, we compare rendered images using SOF~\cite{Radl2025SOF} as a baseline, with either the VastGaussian appearance embedding, or our proposed SSIM-decoupled appearance embedding.

Compared to VastGaussian, our illumination is visibly more consistent.
This shows how varying illumination is baked into the 3DGS point cloud, which degrades overall mesh quality (as demonstrated in the quantitative evaluation).

\section{Mesh-Based Novel View Synthesis}
Following \milo~\cite{guedon2025milo}, we additionally provide numbers for mesh-based novel view synthesis, presented in \cref{tab:mesh_nvs}.
We utilize the same script as open-sourced by \milo\footnote{\url{https://github.com/Anttwo/MILo/blob/mesh_eval/milo/eval/mesh_nvs/render_mesh_nvs.py}}, ensuring a fair and reproducible comparison.
For this purpose, we again utilize the Tanks \& Temples dataset, as well as the Mip-NeRF 360 dataset~\cite{barron2022mipnerf360}; 
we report results for all methods which extract unbounded meshes.
Note that differently from NVS-focused methods such as 3DGS~\cite{kerbl20233dgs}, we use all 6 scenes from the Tanks \& Temples dataset.

As we can see, our method outperforms all other method for the Tanks \& Temples dataset, whilst achieving the second-best scores across Mip-NeRF 360.
Particularly for the Tanks \& Temples dataset, our confidence-based losses and our improved appearance module directly translate to better geometry.
For the Mip-NeRF 360 dataset, our method achieves the second-best scores for all metrics, although outperformed by \milo~\cite{guedon2025milo}.
Note that as the coloring is achieved via a TensoRF backbone~\cite{Chen2022ECCV} (see \milo for details) conditioned on the rasterized depth, this evaluation setting does not favor methods with a higher primitive count; 
which would negatively impact \milo.
Conversely, due to the inherent smoothness of the color field, this comparison benefits \milo, as it exhibits smoother geometry overall.

\begin{table}[]
    \centering
        \setlength{\tabcolsep}{2pt}
    \caption{
\textbf{Mesh-Based Novel View Synthesis:}
We report novel view synthesis metrics as a proxy for evaluating underlying geometric accuracy, using the Mip-NeRF 360 and Tanks \& Temples datasets.
We attain comparable quality to \milo~\cite{guedon2025milo} across both tested datasets.
    }
    \label{tab:mesh_nvs}
\resizebox{.98\linewidth}{!}{
\begin{tabular}{lrrrrrrrr}
\toprule
& \multicolumn{4}{c}{Tanks \& Temples} & \multicolumn{4}{c}{Mip-NeRF 360} \\
\cmidrule(lr){2-5}\cmidrule(lr){6-9}
 & PSNR\textsuperscript{$\uparrow$} & SSIM\textsuperscript{$\uparrow$} & LPIPS\textsuperscript{$\downarrow$}  & \# Verts\textsuperscript{$\downarrow$} & PSNR\textsuperscript{$\uparrow$} & SSIM\textsuperscript{$\uparrow$} & LPIPS\textsuperscript{$\downarrow$} & \# Verts\textsuperscript{$\downarrow$}\\
\midrule
Ours & \cellcolor{tab_color!49} 21.186 & \cellcolor{tab_color!49} 0.683 & \cellcolor{tab_color!49} 0.345  & \cellcolor{tab_color!32} 13.06M & \cellcolor{tab_color!32} 23.639 & \cellcolor{tab_color!32} 0.661 & \cellcolor{tab_color!32} 0.332 & \cellcolor{tab_color!32} 18.43M \\
\milo & \cellcolor{tab_color!32} 21.169 & \cellcolor{tab_color!15} 0.673 & \cellcolor{tab_color!32} 0.354 & \cellcolor{tab_color!49} 5.34M & \cellcolor{tab_color!49} 23.869 & \cellcolor{tab_color!49} 0.675 & \cellcolor{tab_color!49} 0.322 & \cellcolor{tab_color!49} 6.20M \\
SOF & 20.936 & 0.672 & \cellcolor{tab_color!32} 0.354
& \cellcolor{tab_color!15} 14.07M & \cellcolor{tab_color!15} 23.304 & \cellcolor{tab_color!15} 0.634 & \cellcolor{tab_color!15} 0.351 & 23.28M \\
GOF & \cellcolor{tab_color!15} 20.959 & \cellcolor{tab_color!32} 0.675 & \cellcolor{tab_color!32} 0.354  & 14.40M & 22.998 & 0.619 & 0.365 & \cellcolor{tab_color!15} 22.37M \\
\bottomrule
\end{tabular}
}
\end{table}

\end{document}